\documentclass[letterpaper,twocolumn,10pt]{article}

\usepackage{zhanggroup}
\usepackage{tikz}
\usepackage{amsfonts}
\usepackage{xspace} 
\usepackage{mathrsfs}
\usepackage{amssymb}
\usepackage{amsmath}
\usepackage{amsthm}
\usepackage{subcaption}
\usepackage{booktabs}
\usepackage{multirow}
\usepackage{graphicx}
\usepackage{caption,subcaption}
\usepackage[absolute]{textpos}
\usepackage[linesnumbered,ruled]{algorithm2e}
\hypersetup{
  colorlinks,
  linkcolor={blue!70!green},
  citecolor={green!70!blue},
  urlcolor={orange!70!red}
}

\usepackage{balance}

\newcommand{\mypara}[1]{\smallskip\noindent{\bf {#1}.} \xspace}

\newcommand{\DatasetTrain}{\mathcal{D}^{\textit{train}}}

\newcommand{\ShadowDataset}{\mathcal{D}_{\textit{shadow}}}
\newcommand{\AuxDataset}{\mathcal{D}_{\textit{aux}}}
\newcommand{\TargetTrain}{\mathcal{D}_{\textit{target}}^{\textit{train}}}
\newcommand{\TargetTest}{\mathcal{D}_{\textit{target}}^{\textit{test}}}
\newcommand{\ShadowTrain}{\mathcal{D}_{\textit{shadow}}^{\textit{train}}}
\newcommand{\ShadowTest}{\mathcal{D}_{\textit{shadow}}^{\textit{test}}}
\newcommand{\TargetModel}{\mathcal{T}}
\newcommand{\ShadowModel}{\mathcal{S}}

\newcommand{\MIAAttackModel}{\mathcal{A}_{\textit{MemInf}}}
\newcommand{\AIAttackModel}{\mathcal{A}_{\textit{AttInf}}}
\newcommand{\MLModel}{\mathcal{M}}
\newcommand{\FeatureSet}{X}
\newcommand{\LabelSet}{Y}
\newcommand{\Encoder}{f}
\newcommand{\ProjectionHead}{g}

\newcommand{\SensitiveAttribute}{s}
\newcommand{\AdvClassifier}{C\xspace}
\newcommand{\FinalSimCLRLoss}{\mathcal{L}_{\textit{Contrastive}}}
\newcommand{\similarity}{\textit{sim}}
\newcommand{\DataPoint}{x}
\newcommand{\Posterior}{p}
\newcommand{\Label}{y}
\newcommand{\LossFunction}{\mathcal{L}}
\newcommand{\CELoss}{\mathcal{L}_{\textit{CE}}}

\newcommand{\RepresentationVector}{h}
\newcommand{\PrivacyPreservingLoss}{\LossFunction_{\textit{Talos}}}
\newcommand{\Talos}{\textit{Talos}\xspace}
\newcommand{\MemGuard}{\textit{MemGuard}\xspace}
\newcommand{\AttriGuard}{\textit{AttriGuard}\xspace}
\newcommand{\Olympus}{\textit{Olympus}\xspace}
\begin{document}
% ----------------------------------------------------

\begin{textblock}{15}(1.9,1)
To Appear in 2021 ACM SIGSAC Conference on Computer and Communications Security, November 2021
\end{textblock}

% ----------------------------------------------------
\title{Quantifying and Mitigating Privacy Risks of \\ Contrastive Learning}
% ----------------------------------------------------

\date{}

\author{
Xinlei He and Yang Zhang
\\
\textit{CISPA Helmholtz Center for Information Security}
}

\title{\bf Quantifying and Mitigating Privacy Risks of Contrastive Learning}

\maketitle

% ----------------------------------------------------
\begin{abstract}
% ----------------------------------------------------

Data is the key factor to drive the development of machine learning (ML) during the past decade.
However, high-quality data, in particular labeled data, is often hard and expensive to collect.
To leverage large-scale unlabeled data, self-supervised learning, represented by contrastive learning, is introduced.
The objective of contrastive learning is to map different views derived from a training sample (e.g., through data augmentation) closer in their representation space, while different views derived from different samples more distant.
In this way, a contrastive model learns to generate informative representations for data samples, which are then used to perform downstream ML tasks.
Recent research has shown that machine learning models are vulnerable to various privacy attacks.
However, most of the current efforts concentrate on models trained with supervised learning.
Meanwhile, data samples' informative representations learned with contrastive learning may cause severe privacy risks as well.

In this paper, we perform the first privacy analysis of contrastive learning through the lens of membership inference and attribute inference.
Our experimental results show that contrastive models trained on image datasets are less vulnerable to membership inference attacks but more vulnerable to attribute inference attacks compared to supervised models.
The former is due to the fact that contrastive models are less prone to overfitting, while the latter is caused by contrastive models' capability of representing data samples expressively.
To remedy this situation, we propose the first privacy-preserving contrastive learning mechanism, \Talos, relying on adversarial training.
Empirical results show that \Talos can successfully mitigate attribute inference risks for contrastive models while maintaining their membership privacy and model utility.\footnote{Our code is available at \url{https://github.com/xinleihe/ContrastiveLeaks}.}

% ----------------------------------------------------
\end{abstract}
% ----------------------------------------------------

% ----------------------------------------------------
\section{Introduction}
% ----------------------------------------------------

Machine learning (ML) has progressed tremendously, and data is the key factor to drive such development.
However, high-quality data, in particular labeled data, is often hard and expensive to collect as this relies on large-scale human annotation.
Meanwhile, unlabeled data is being generated at every moment. 
To leverage unlabeled data for machine learning tasks, \textit{self-supervised learning} has been introduced~\cite{LZHWMZT20}.
The goal of self-supervised learning is to derive labels from an unlabeled dataset and train an unsupervised task in a supervised manner.
A trained self-supervised model serves as an encoder transforming data samples into their representations which are then used to perform supervised downstream ML tasks. 
One of the most prominent self-supervised learning paradigms is \textit{contrastive learning}~\cite{GH10,OLV18,HFLGBTB19,CKNH20,YCSCWS20,HFWXG20,JXZZZZ20}, with SimCLR~\cite{CKNH20} as its most representative framework~\cite{LZHWMZT20}.

Different from supervised learning which directly optimizes an ML model on a labeled training dataset, referred to as a supervised model, contrastive learning aims to train a contrastive model, which is able to generate expressive representations for data samples, and relies on such representations to perform downstream supervised ML tasks.
The optimization objective for contrastive learning is to map different views derived from one training sample (e.g., through data augmentation) closer in the representation space while different views derived from different training samples more distant.
By doing this, a contrastive model is capable of representing each sample in an informative way.

Recently, machine learning models have been demonstrated to be vulnerable to various privacy attacks against their training dataset~\cite{SSSS17,SS19,MSCS19,SZHBFB19,HMDC19,SS20,CYZF20,CTWJHLRBSEOR20,HJBGZ21}.
The two most representative attacks in this domain are membership inference attack~\cite{SSSS17,SZHBFB19} and attribute/property inference attack~\cite{MSCS19,SS20}.
The former aims to infer whether a data sample is part of a target ML model's training dataset.
The latter leverages the overlearning property of a machine learning model to infer the sensitive attribute of a data sample.
So far, most of the research on the privacy of machine learning concentrates on supervised models.
Meanwhile, informative representations for data samples learned by contrastive models may cause severe privacy risks as well.
To the best of our knowledge, this has been left largely unexplored.

\mypara{Our Contributions}
In this paper, we perform the first privacy quantification of contrastive learning, the most representative self-supervised learning paradigm.
More specifically, we study the privacy risks of data samples in the contrastive learning setting,  with a focus on SimCLR, through the lens of membership inference and attribute inference, and we concentrate on contrastive models trained on image datasets.

We adapt the existing attack methodologies for membership inference (neural network-based, metric-based, and label-only) and attribute inference against supervised models to contrastive models.
Our empirical results show that contrastive models are less vulnerable to membership inference attacks than supervised models.
For instance, considering the neural network-based attacks, we achieve 0.620 membership inference accuracy on a contrastive model trained on STL10~\cite{CNL11} with ResNet-50~\cite{HZRS16}, while the result is 0.810 on the corresponding supervised model.
The reason behind this is contrastive models are less prone to overfitting.

On the other hand, we observe that contrastive models are more vulnerable to attribute inference attacks than supervised models.
For instance, on the UTKFace~\cite{ZSQ17} dataset with ResNet-18, we can achieve 0.701 attribute inference attack accuracy on the contrastive model while only 0.422 on the supervised model.
This is due to the fact that the representations generated by a contrastive model contain rich and expressive information about their original data samples, which can be exploited for effective attribute inference.

To mitigate the attribute inference risks stemming from contrastive models, we propose the first privacy-preserving contrastive learning mechanism, namely \Talos, relying on adversarial training.
Concretely, \Talos introduces an adversarial classifier into the original contrastive learning framework to censor the sensitive attributes learned by a contrastive model.
Our evaluation reveals that \Talos can successfully mitigate attribute inference risks for contrastive models while maintaining their membership privacy and model utility.
Our code and models will be made publicly available.

In summary, we make the following contributions:
\begin{itemize}
\item We take the first step towards quantifying the privacy risks of contrastive learning.
\item Our empirical evaluation shows that contrastive models trained on image datasets are less vulnerable to membership inference attacks but more prone to attribute inference attacks compared to supervised models.
\item We propose the first privacy-preserving contrastive learning mechanism, which is able to protect the trained contrastive models from attribute inference attacks without jeopardizing their membership privacy and model utility.
\end{itemize}

% ----------------------------------------------------
\section{Preliminary}
\label{section:Preliminary}
% ----------------------------------------------------

% ----------------------------------------------------
\subsection{Supervised Learning}
% ----------------------------------------------------

Supervised learning, represented by classification, is one of the most common and important ML applications.
We first denote a set of data samples by $\FeatureSet$ and a set of labels by $\LabelSet$.
The objective of a supervised ML model $\MLModel$ is to learn a mapping function from each data sample $\DataPoint\in \FeatureSet$ to its label/class $\Label\in \LabelSet$.
Formally, we have
\begin{equation}
\MLModel: \DataPoint\mapsto\Label
\end{equation}
Given a sample $\DataPoint$, its output from $\MLModel$, denoted by $\Posterior=\MLModel(\DataPoint)$, is a vector that represents the probability distribution of the sample belonging to a certain class.
In this paper, we refer to $\Posterior$ as the prediction posteriors.
To train an ML model, we need to define a loss function $\LossFunction(\Label, \MLModel(\DataPoint))$ which measures the distance between a sample's prediction posteriors and its label.
The training process is then performed by minimizing the expectation of the loss function over a training dataset $\DatasetTrain$, i.e., the empirical loss.
We formulate this as follow:
\begin{equation}
\arg \min_{\MLModel} \frac{1}{\vert\DatasetTrain \vert} \sum_{(\DataPoint, \Label) \in \DatasetTrain} \LossFunction(\Label, \MLModel(\DataPoint))
\end{equation}
Cross-entropy loss is one of the most common loss functions used for classification tasks, it is defined as the following.
\begin{equation}
\label{equation:CELoss}
\LossFunction_{\textit{CE}}(\Label, \Posterior) = -\sum_{i=1}^{k} \Label^i\log{\Posterior^{i}}
\end{equation}
Here, $k$ is the total number of classes, $\Label^{i}$ equals to 1 if the sample belongs to class $i$ (otherwise 0), and $\Posterior^{i}$ is the $i$-th element of the posteriors $\Posterior$.
In this paper, we use cross-entropy as the loss function to train all the supervised models.

% ----------------------------------------------------
\subsection{Contrastive Learning}
% ----------------------------------------------------

Supervised learning is powerful, but its success heavily depends on the labeled training dataset.
In the real world, high-quality labeled dataset is hard and expensive to obtain as it often relies on human annotation.
For instance, the ILSVRC2011 dataset~\cite{RDSKSMHKKBBF15} contains more than 12 million labeled images that are all annotated by Amazon Mechanical Turk workers.
Meanwhile, unlabeled data is being generated at every moment.
To leverage large-scale unlabeled data, self-supervised learning is introduced.

The goal of self-supervised learning is to get labels from an unlabeled dataset for free so that one can train an unsupervised task on this unlabeled dataset in a supervised manner.
Contrastive learning/loss~\cite{GH10,OLV18,HFLGBTB19,CKNH20,YCSCWS20,HFWXG20,JXZZZZ20} is one of the most successful and representative self-supervised learning paradigms in recent years and has received a lot of attention from both academia and industry.
In general, contrastive learning aims to map a sample closer to its correlated views and more distant to other samples' correlated views.
In this way, contrastive learning is able to learn an informative representation for each sample, which can then be leveraged to perform different downstream tasks.
Contrastive learning relies on Noise Contrastive Estimation (NCE)~\cite{GH10} as its objective function, which can be formulated as:
\begin{equation}
\label{equation:NCE}
\mathcal{L}=-\log (\frac{ e^{\similarity(\Encoder(\DataPoint), \Encoder(\DataPoint^{+}))} }{ e^{\similarity(\Encoder(\DataPoint), \Encoder(\DataPoint^{+}))} + e^{\similarity(\Encoder(\DataPoint), \Encoder(\DataPoint^{-}))} })
\end{equation}
where $f$ is an encoder that maps a sample into its representation, $ \DataPoint^{+}$ is similar to $\DataPoint$ (referred to as a positive pair), $\DataPoint^{-}$ is dissimilar to $\DataPoint$ (referred to as a negative pair), and $\similarity$ is a similarity function.
The structure of the encoder and the similarity function can vary from different tasks.
In this paper, we focus on one of the most popular contrastive learning frameworks~\cite{LZHWMZT20}, namely SimCLR~\cite{CKNH20}.
This framework is assembled with the following components.

\mypara{Data Augmentation} 
SimCLR first uses a data augmentation module to transform a given data sample $\DataPoint$ to its two augmented views, denoted by $\tilde{\DataPoint}_{i}$ and $\tilde{\DataPoint}_{j}$, which can be considered as a positive pair for $\DataPoint$.
In our work, we follow the same data augmentation process used by SimCLR~\cite{CKNH20}, i.e., first random cropping and flipping with resizing, second random color distortions, and third random Gaussian blur. 

\mypara{Base Encoder $\Encoder$}  
Base encoder $\Encoder$ is used to extract representations from the augmented data samples. 
The base encoder can follow various neural network (NN) architectures.
In this paper, we apply the widely used ResNet~\cite{HZRS16} (ResNet-18 and ResNet-50) and MobileNetV2~\cite{SHZZC18} to obtain the representation $h_i = \Encoder(\tilde{\DataPoint}_{i})$ for $\tilde{\DataPoint}_{i}$. 

\mypara{Projection Head $\ProjectionHead$}
Projection head $\ProjectionHead$ is a simple neural network that maps the representations from the base encoder to another latent space to apply the contrastive loss.
The goal of the projection head is to enhance the encoder's performance.
Following Chen et al.~\cite{CKNH20}, we implement it with a 2-layer MLP (multilayer perceptron) to obtain the output $z_i = \ProjectionHead(h_i)$ for $h_i$.

\mypara{Contrastive Loss Function} 
The contrastive loss function is defined to guide the model to learn the general representation from the data itself.
Given a set of augmented samples $\{\tilde{\DataPoint}_{k}\}$ including a positive pair $\tilde{\DataPoint}_{i}$ and $\tilde{\DataPoint}_{j}$, the contrastive loss maximizes the similarity between $\tilde{\DataPoint}_{i}$ and $\tilde{\DataPoint}_{j}$ and minimizes the similarity between $\tilde{\DataPoint}_{i}$ ($\tilde{\DataPoint}_{j}$) and other samples.
For each mini-batch of $N$ samples, we have $2N$ augmented samples.
The loss function for a positive pair $\tilde{\DataPoint}_{i}$ and $\tilde{\DataPoint}_{j}$ can be formulated as:
\begin{equation}
\label{equation:simclr_loss}
\ell(i, j)=-\log \frac{e^{\operatorname{\similarity}\left(\boldsymbol{z}_{i}, \boldsymbol{z}_{j}\right) / \tau}}{\sum_{k=1, k \neq i}^{2 N} e^{\operatorname{\similarity}\left(\boldsymbol{z}_{i}, \boldsymbol{z}_{k}\right) / \tau}}
\end{equation}
where $\similarity(z_{i}, z_{j})= {z_{i}}^{\top} z_{j} /\|z_{i}\|\|z_{j}\|$ represents the cosine similarity between $z_{i}$ and $z_{j}$ and $\tau$ is a temperature parameter.
The final loss is calculated over all positive pairs in a mini-batch, which can be defined as the following.
\begin{equation}
\label{equation:contrastive_loss}
\FinalSimCLRLoss=\frac{1}{2N} \sum_{k=1}^{N}[\ell(2 k-1,2 k)+\ell(2 k, 2 k-1)]
\end{equation}
Here, $2k-1$ and $2k$ are the indices for each positive pair.

Training classifiers with SimCLR can be partitioned into two phases.
In the first phase, we train a base encoder as well as a projection head by the contrastive loss using an unlabeled dataset.
After training, we discard the projection head and keep the base encoder only.
In the second phase, to perform classification tasks, we freeze the parameters of the encoder, add a trainable linear layer at the end of the encoder, and fine-tune the linear layer with the cross-entropy loss (see \autoref{equation:CELoss}) on a labeled dataset.
The linear layer serves as a classifier, with its input being the representations generated by the encoder.
We refer to this linear layer as the \textit{classification layer}.
In the rest of the paper, we call a model trained with supervised learning as a \textit{supervised model} and a model trained with contrastive learning as a \textit{contrastive model}.
Also, we consider contrastive models trained on image datasets, as most of the current development of contrastive learning focus on images.

Compared to supervised learning, contrastive learning can learn more informative representations for data samples.
Previous work shows that supervised models are vulnerable to various privacy attacks~\cite{SSSS17,YGFJ18,SZHBFB19,MSCS19,CYZF20,SS20,CTWJHLRBSEOR20}.
However, to the best of our knowledge, privacy risks stemming from contrastive models have been left largely unexplored.
In this work, we aim to fill this gap.

% ----------------------------------------------------
\section{Membership Inference Attack}
\label{section:MIA}
% ----------------------------------------------------

We first quantify the privacy risks of contrastive models through the lens of membership inference.
Note that our goal here is not to propose a novel membership inference attack, instead, we aim to quantify the membership privacy of contrastive models.
Therefore, we follow existing attacks and their threat models~\cite{SSSS17,SZHBFB19,SM21,LZ21,CTCP20}.

% ----------------------------------------------------
\subsection{Attack Definition and Threat Model}
\label{subsection:MIAThreatModel}
% ----------------------------------------------------

Membership inference attack is one of the most popular privacy attacks against ML models~\cite{SSSS17,SZHBFB19,JSBZG19,HMDC19,CYZF20,LF20,LZ21,SM21,CTCP20,CZWBHZ20}.
The goal of membership inference is to determine whether a data sample $\DataPoint$ is part of the training dataset of a target model $\TargetModel$.
We formally define a membership inference attack model $\MIAAttackModel: \DataPoint, \TargetModel \mapsto \{\textit{member}, \textit{non-member}\}$.
Here, the target model is the contrastive model introduced in \autoref{section:Preliminary}.
A successful membership inference attack can cause severe privacy risks.
For instance, if a model is trained on data samples collected from people with certain sensitive information, then successfully inferring a sample from a person being a member of the model can directly reveal the person's sensitive information.

Following previous work~\cite{SSSS17,SZHBFB19,SM21,LZ21,CTCP20}, we assume that an adversary only has black-box access to the target model $\TargetModel$, i.e., they can only query $\TargetModel$ with their data samples and obtain the outputs.
In addition, the adversary also has a shadow dataset $\ShadowDataset$, which comes from the same distribution as the target model's training dataset.
The shadow dataset $\ShadowDataset$ is used to train a shadow model $\ShadowModel$, the goal of which is to obtain the necessary information to perform the attack.
We further assume that the shadow model shares the same architecture as the target model~\cite{SSSS17}.
This is realistic as the adversary can use the same machine learning service as the target model owner to train their shadow model. 
Alternatively, the adversary can also learn the target model's architecture first by applying model extraction attacks~\cite{TZJRR16,WG18,OASF18,OSF19}.

% ----------------------------------------------------
\subsection{Methodology}
\label{subsection:MIAMethodology}
% ----------------------------------------------------

We adapt the previous membership inference attacks, which are designed for supervised models, to contrastive models~\cite{SSSS17,SZHBFB19,CTCP20,SM21}.
Concretely, we consider three types of membership inference attacks, i.e., NN-based attacks~\cite{SSSS17, SZHBFB19}, metric-based attacks~\cite{SM21}, and label-only attacks~\cite{CTCP20}.

\mypara{NN-based Attacks (Neural Network-based Attacks)}
In NN-based attacks, the adversary aims to train an attack model to differentiate members and non-members using the posteriors generated from the target model and their predicted labels. 
Given a shadow dataset $\ShadowDataset$, the adversary first splits it into two disjoint sets, namely shadow training dataset $\ShadowTrain$ and shadow testing dataset $\ShadowTest$.
$\ShadowTrain$ is used to train the shadow model $\ShadowModel$, which mimics the behavior of the target model.
This means the shadow model is trained to perform the same task as the target model.
Then, the adversary uses $\ShadowDataset$ (including both $\ShadowTrain$ and $\ShadowTest$) to query the shadow model $\ShadowModel$ and obtains the corresponding posteriors and prediction labels.
For each data sample in $\ShadowDataset$, the adversary ranks its posteriors in descending order and takes the largest two posteriors (classification tasks considered in this paper have at least two classes) as part of the input to the attack model. 
The other part is an indicator representing whether the prediction is correct or not.
Thus, the dimension of the input to $\MIAAttackModel$ is 3.
If a sample belongs to $\ShadowTrain$, the adversary labels its corresponding input to the attack model as a member, otherwise as a non-member.
Then, this obtained dataset is used to train the attack model, which is a binary machine learning classifier.
To determine whether a target data sample $\DataPoint$ is used to train the target model, the adversary first queries the target model $\TargetModel$ with $\DataPoint$ and obtains the input to the attack model for this sample.
Then, the adversary queries this input to the attack model and gets its membership prediction.

\mypara{Metric-based Attacks} 
Song and Mittal~\cite{SM21} propose several metric-based attacks.
Similar to NN-based attacks, metric-based attacks need to train shadow models.
However, instead of training an attack model, metric-based attacks leverage a certain metric and a predefined threshold on that metric (calculated over the shadow model) to determine a sample's membership status.
Song and Mittal~\cite{SM21} propose four metrics, i.e., prediction correctness (metric-corr), prediction confidence (metric-conf), prediction entropy (metric-ent), and modified prediction entropy (metric-ment).

\mypara{Label-only Attacks} 
Label-only attacks~\cite{CTCP20} consider a more restrict scenario where the target model only exposes the predicted label instead of posteriors.
Similar to previous attacks, this attack requires the adversary to train a shadow model.
Label-only attacks focus more on the input samples instead of the model's outputs, relying on the adversarial example techniques.
The key intuition is that the magnitude of perturbation to change the predicted label of member samples is larger than that of non-member samples.
The adversary can exploit the magnitude of the perturbation to distinguish members and non-members.

% ----------------------------------------------------
\subsection{Experimental Settings}
\label{subsection:MIAExperimentalSetting}
% ----------------------------------------------------

\begin{figure*}[!t]
\centering
\begin{subfigure}{\columnwidth}
\includegraphics[width=\columnwidth]{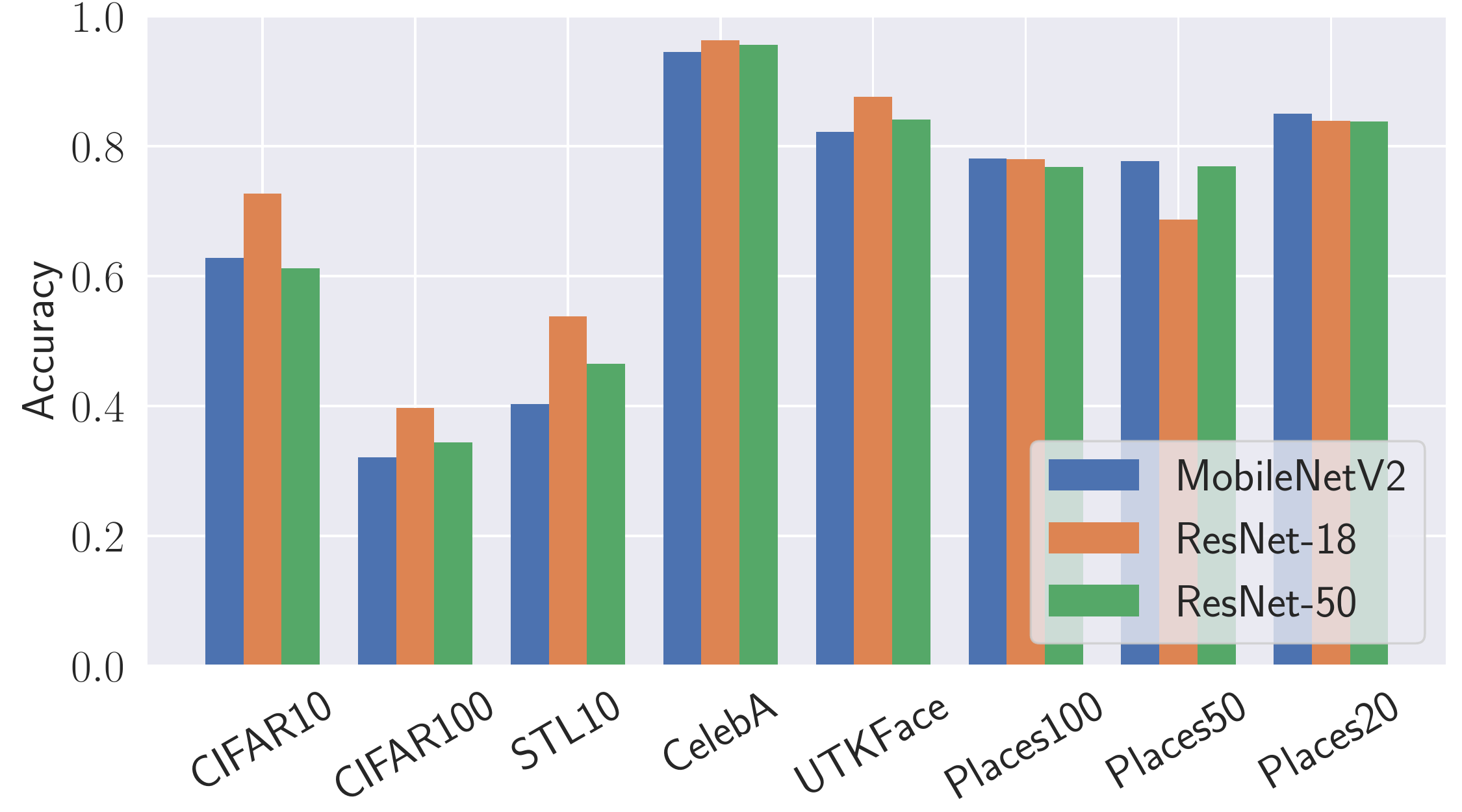}
\caption{Supervised Model}
\label{figure:mia_target_train}
\end{subfigure}
\begin{subfigure}{\columnwidth}
\includegraphics[width=\columnwidth]{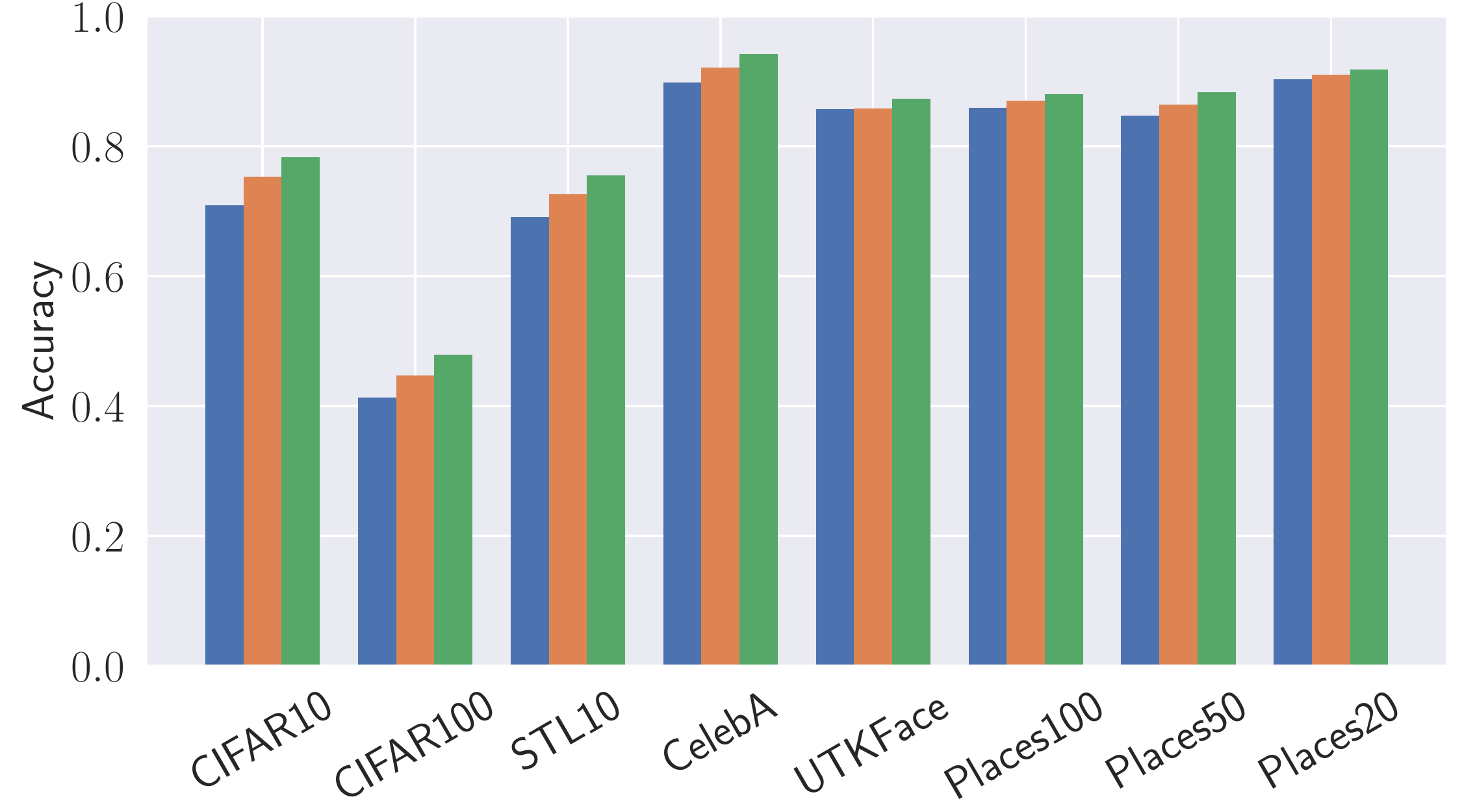}
\caption{Contrastive Model}
\label{figure:mia_target_test}
\end{subfigure}
\caption{The performance of original classification tasks for both supervised models and contrastive models with MobileNetV2, ResNet-18, and ResNet-50 on 8 different datasets.
The x-axis represents different datasets.
The y-axis represents original classification tasks' accuracy.}
\label{figure:target_performance}
\end{figure*}

\mypara{Datasets}
We utilize 8 different image datasets to conduct our experiments for membership inference.

\begin{itemize}
\item \textbf{CIFAR10~\cite{CIFAR}.} 
This dataset contains 60,000 images in 10 classes. 
Each class represents one object and has 6,000 images.
The size of each image is $32 \times 32$.
\item \textbf{CIFAR100~\cite{CIFAR}.}
This dataset is similar to CIFAR10, except it has 100 classes, with each class containing 600 images.
The size of each image is also $32 \times 32$.
\item \textbf{STL10~\cite{CNL11}.}
This dataset is composed of 10 classes of images.
Each class has 1,300 samples.
The size of each image is $96 \times 96$.
Besides the labeled image, STL10 also contains 100,000 unlabeled images, which we use for pretraining the encoder for the contrastive model (detailed later).
These images are extracted from a broader distribution compared to those with labeled classes.
\item \textbf{CelebA~\cite{LLWT15}.} 
This dataset is composed of more than 200,000 celebrities' facial images. 
Note that in CelebA, we randomly select 60,000 images for our experiments.
We set its target model's classification task as gender classification.
\item \textbf{UTKFace~\cite{ZSQ17}.} 
This dataset consists of over 23,000 facial images labeled with gender, age, and race.
We set its target model's classification task as gender classification as well.
\item \textbf{Places365~\cite{ZLKOT18}.} 
This dataset is composed of more than 1.8 million images with 365 scene categories.
We randomly select 100 scene categories and randomly select 400 images per category to form the \textbf{Places100} dataset.
Besides, we randomly select 50 (20) scene categories and randomly select 800 (2,000) images per category to form the \textbf{Places50} (\textbf{Places20}) dataset.
Each dataset contains 40,000 images in total.
We follow Song and Shmatikov~\cite{SS20} and set its target model's classification task as predicting whether the scene is indoor or outdoor.
\end{itemize}

All the datasets are used to evaluate membership inference attacks, while UTKFace, Places100, Places50, and Places20 are also used to evaluate attribute inference attacks since they have extra labels that can be used as sensitive attributes (see \autoref{subsection:AIExperimentalSetting}).
For all the datasets, we rescale their images to the size of $96 \times 96$. 
Note that we concentrate on image datasets as it is the most prominent domain for applying contrastive learning at the moment~\cite{HFWXG20,GH10,OLV18,CKNH20,YCSCWS20,JXZZZZ20}.
We leave our investigation in other data domains as future work.

\begin{figure*}[!t]
\centering
\begin{subfigure}{\columnwidth}
\includegraphics[width=\columnwidth]{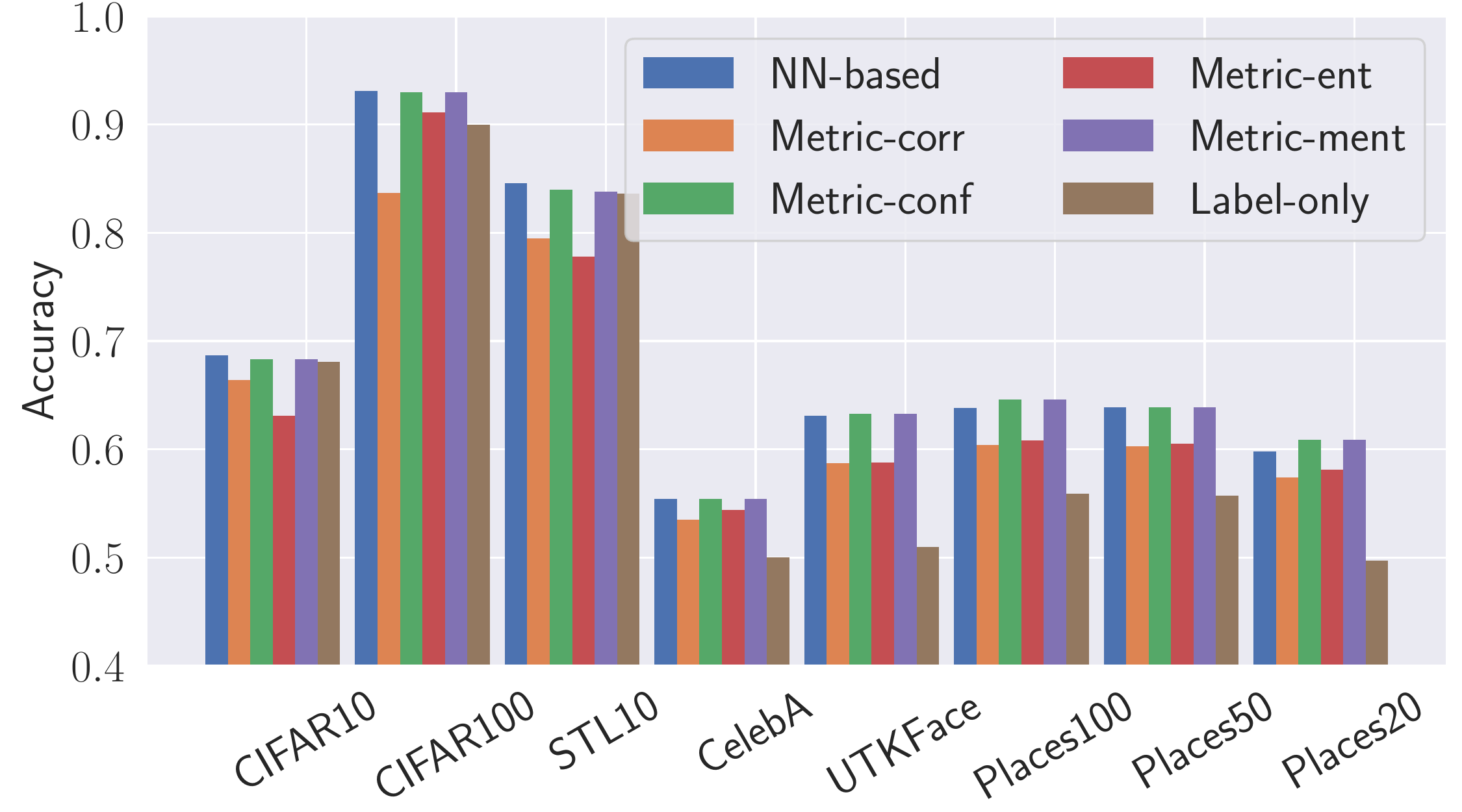}
\caption{Supervised Model}
\label{figure:mia_attack_supervised_mobilenet}
\end{subfigure}
\begin{subfigure}{\columnwidth}
\includegraphics[width=\columnwidth]{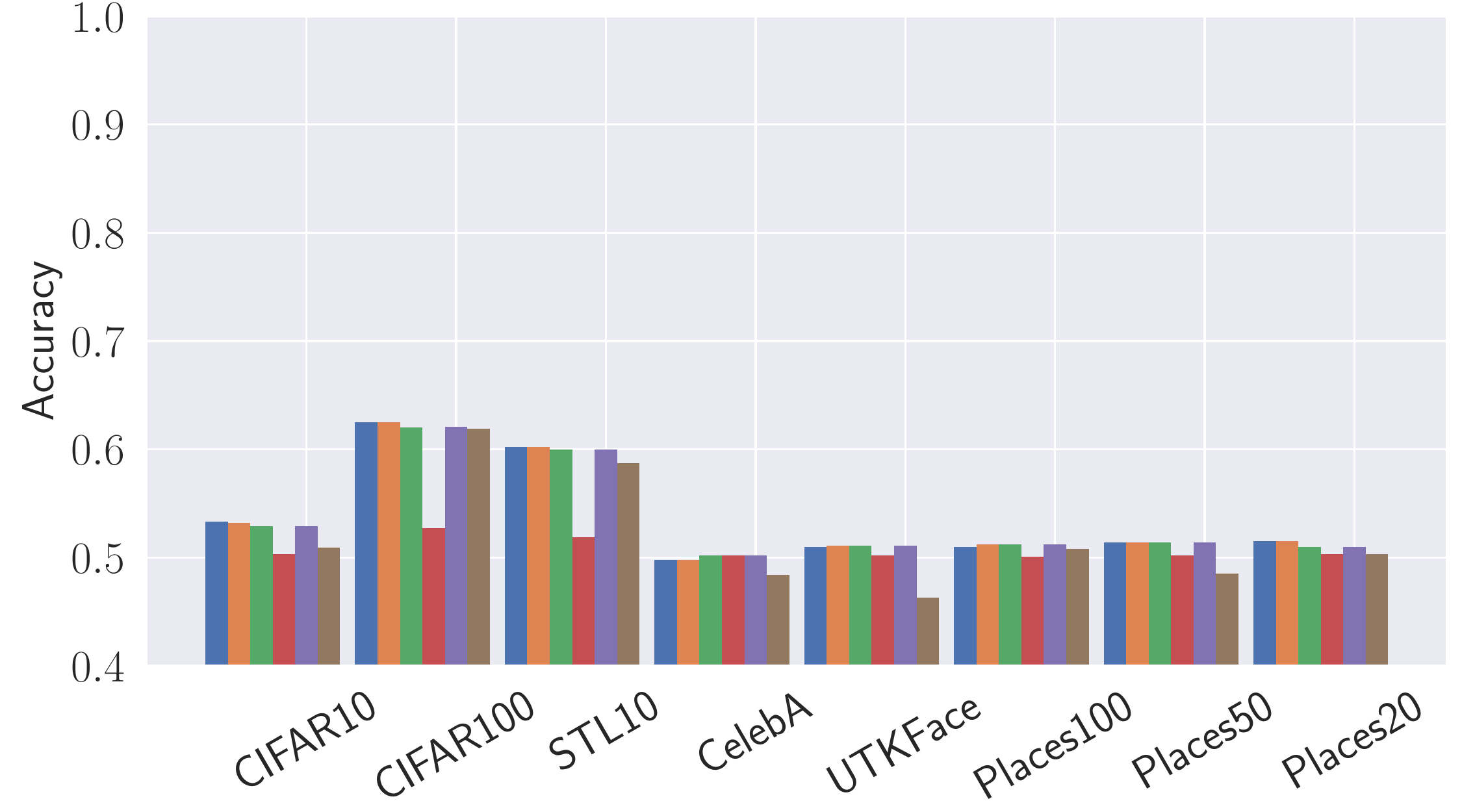}
\caption{Contrastive Model}
\label{figure:mia_attack_contrastive_mobilenet}
\end{subfigure}
\caption{The performance of different membership inference attacks against both supervised models and contrastive models with MobileNetV2 on 8 different datasets.
The x-axis represents different datasets.
The y-axis represents membership inference attacks' accuracy.}
\label{figure:mia_attack_performance_mobilenet}
\end{figure*}

\mypara{Datasets Configuration}
For each dataset, we first split it into four equal parts, i.e., $\TargetTrain$, $\TargetTest$, $\ShadowTrain$, and $\ShadowTest$.
$\TargetTrain$ is used to train the target model $\TargetModel$, the samples of which are thus considered as members of the target model.
We treat $\TargetTest$ as non-members of the target model $\TargetModel$.
$\ShadowTrain$ is used to train the shadow model $\ShadowModel$, and $\ShadowTrain$ and $\ShadowTest$ are used to create the attack model $\MIAAttackModel$.

\mypara{Metric}
Since the attack model's training and testing datasets are both balanced with respect to membership distribution, we adopt accuracy as our evaluation metric following previous work~\cite{SSSS17,SZHBFB19}.

\mypara{Attack Model}
For NN-based attacks, the attack model is a 3-layer MLP, and the number of neurons for each hidden layer is set to 32.
We use cross-entropy as the loss function and Adam as the optimizer with a learning rate of 0.05.
The attack model is trained for 100 epochs.
For metric-based attacks, we follow the implementation of Song et al.~\cite{SM21}.
For label-only attacks, we leverage the implementation of ART~\cite{ART}.

\mypara{Target Model (Contrastive Model)}
We adopt three popular neural network architectures as the contrastive model's base encoder $\Encoder$ in our experiments, including MobileNetV2~\cite{SHZZC18}, ResNet-18~\cite{HZRS16}, and ResNet-50~\cite{HZRS16}.
Specifically, we discard the last classification layer of MobileNetV2, ResNet-18, and ResNet-50 and use the remaining parts as $\Encoder$.
Then, a 2-layer MLP is added after $\Encoder$ as the projection head $\ProjectionHead$.
For ResNet-18, the dimensions for the output of $\Encoder$, the first-layer of $\ProjectionHead$, and the second-layer of $\ProjectionHead$ are set to 512, 512, and 256, respectively.
For ResNet-50, the corresponding dimensions are 2,048, 256, and 256.
For MobileNetV2, the corresponding dimensions are 1,280, 256, and 256.

After training the base encoder with the contrastive loss, we ignore the projection head $\ProjectionHead$ and add a new linear layer to the base encoder $\Encoder$ as its classification layer.
For all datasets, we first use the unlabeled dataset of STL10 to pretrain the base encoder $\Encoder$ for 100 epochs.
Then, we fine-tune the base encoder $\Encoder$ with the corresponding training dataset (without label) for 100 epochs.
In the end, we freeze the parameters of $\Encoder$ and use the corresponding training dataset to only fine-tune the classification layer for 100 epochs to establish the contrastive model.
In all cases, Adam is utilized as the optimizer.

\mypara{Baseline (Supervised Model)}
To fully understand the privacy leakage of contrastive models, we further use supervised models as the baseline.
We train three models including MobileNetV2, ResNet-18, and ResNet-50 from scratch for all the datasets.
The models are trained for 100 epochs.
Cross-entropy is adopted as the loss function, and we again use Adam as the optimizer.
Our code is currently implemented in Python 3.6 and PyTorch 1.6.0, and run on an NVIDIA DGX-A100 server with Ubuntu 18.04.

% ----------------------------------------------------
\subsection{Results}
\label{subsection:MIAResult}
% ----------------------------------------------------

We first show the performance of supervised models and contrastive models on their original classification tasks in \autoref{figure:target_performance}.
We observe that contrastive models perform better than supervised models on most of the datasets.
For instance, on STL10 with ResNet-18 as the base encoder, the contrastive model achieves 0.726 accuracy while the supervised model achieves 0.538 accuracy.

\begin{figure}[!t]
\centering
\begin{subfigure}{0.47\columnwidth}
\includegraphics[width=\columnwidth]{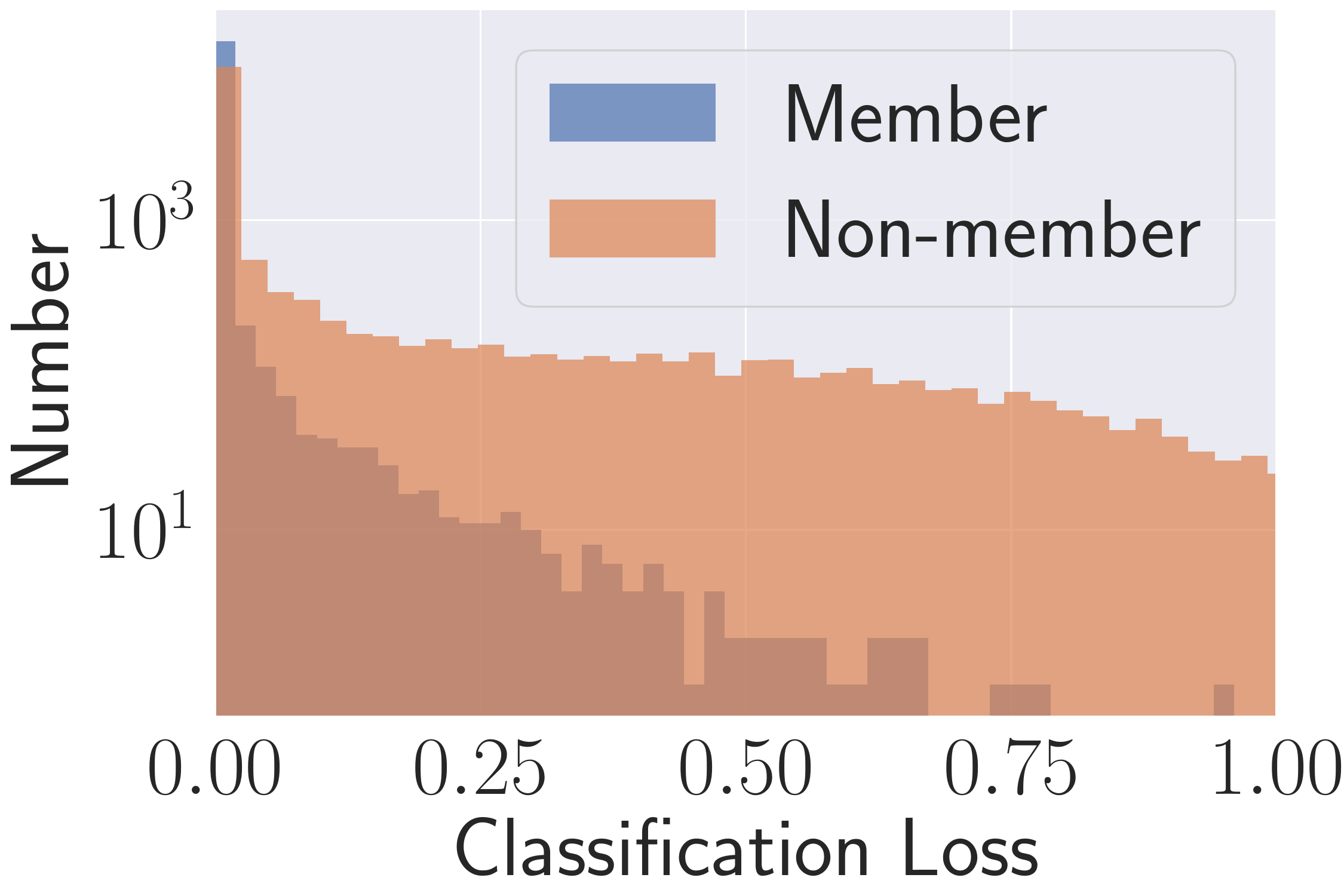}
\caption{Supervised Model}
\label{fig:loss_ce}
\end{subfigure}
\begin{subfigure}{0.47\columnwidth}
\includegraphics[width=\columnwidth]{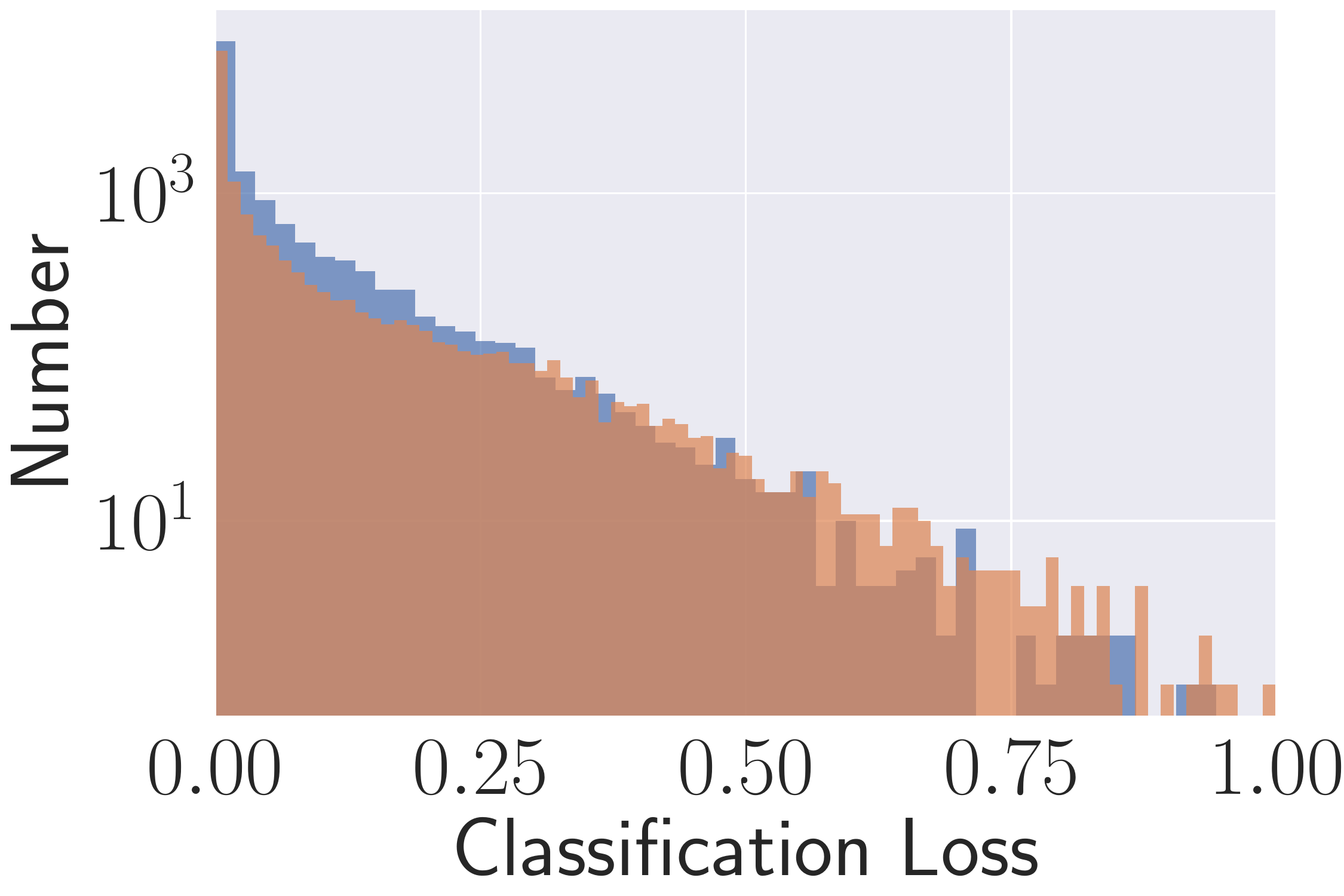}
\caption{Contrastive Model}
\label{fig:loss_simclr}
\end{subfigure}
\caption{The distribution of loss with respect to original classification tasks for member and non-member samples for both the supervised model and the contrastive model with ResNet-18 on CIFAR10.
The x-axis represents each sample's classification loss.
The y-axis represents the number of member and non-member samples.}
\label{figure:loss_dsitribution}
\end{figure}

\begin{figure}[!t]
\centering
\includegraphics[width=0.8\columnwidth]{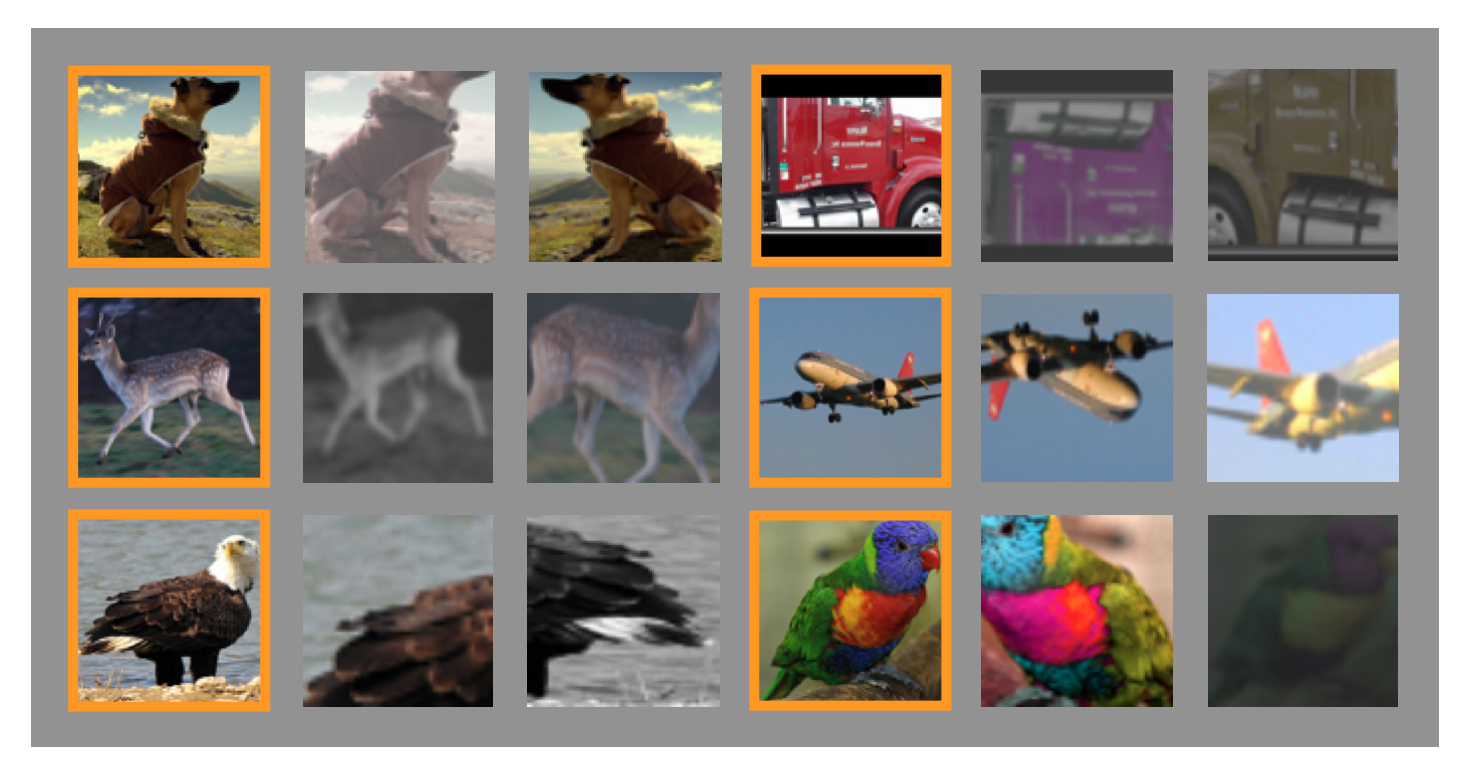}
\caption{Randomly selected images from STL10 and their augmented views used during the process of contrastive learning.
The first and fourth columns show the original images (bounded by orange boxes), and the rest columns show their augmented views.}
\label{figure:img_visualize}
\end{figure} 

\begin{figure*}[!t]
\centering
\begin{subfigure}{0.65\columnwidth}
\includegraphics[width=\columnwidth]{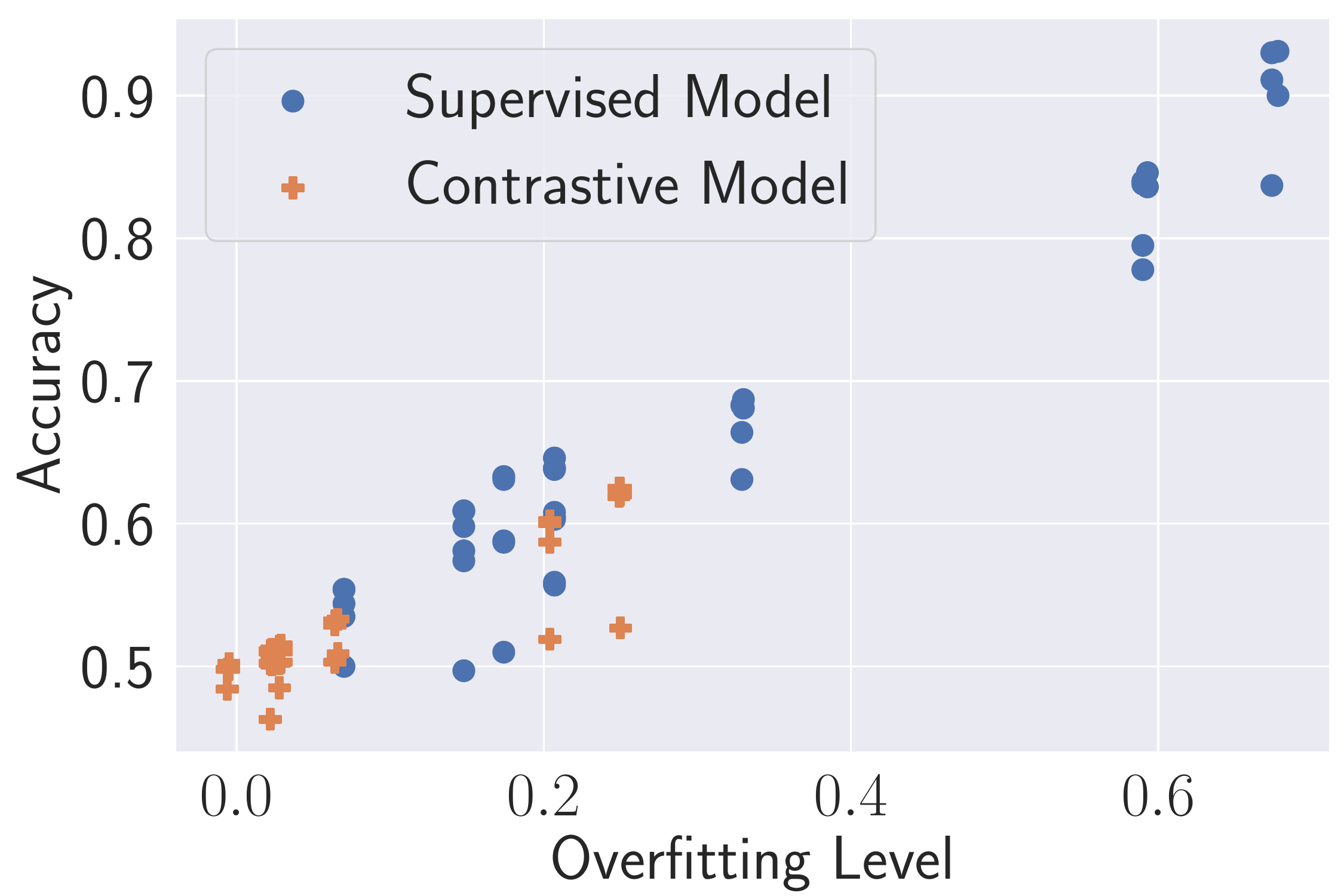}
\caption{MobileNetV2}
\label{figure:overfitting_mobilenet}
\end{subfigure}
\begin{subfigure}{0.65\columnwidth}
\includegraphics[width=\columnwidth]{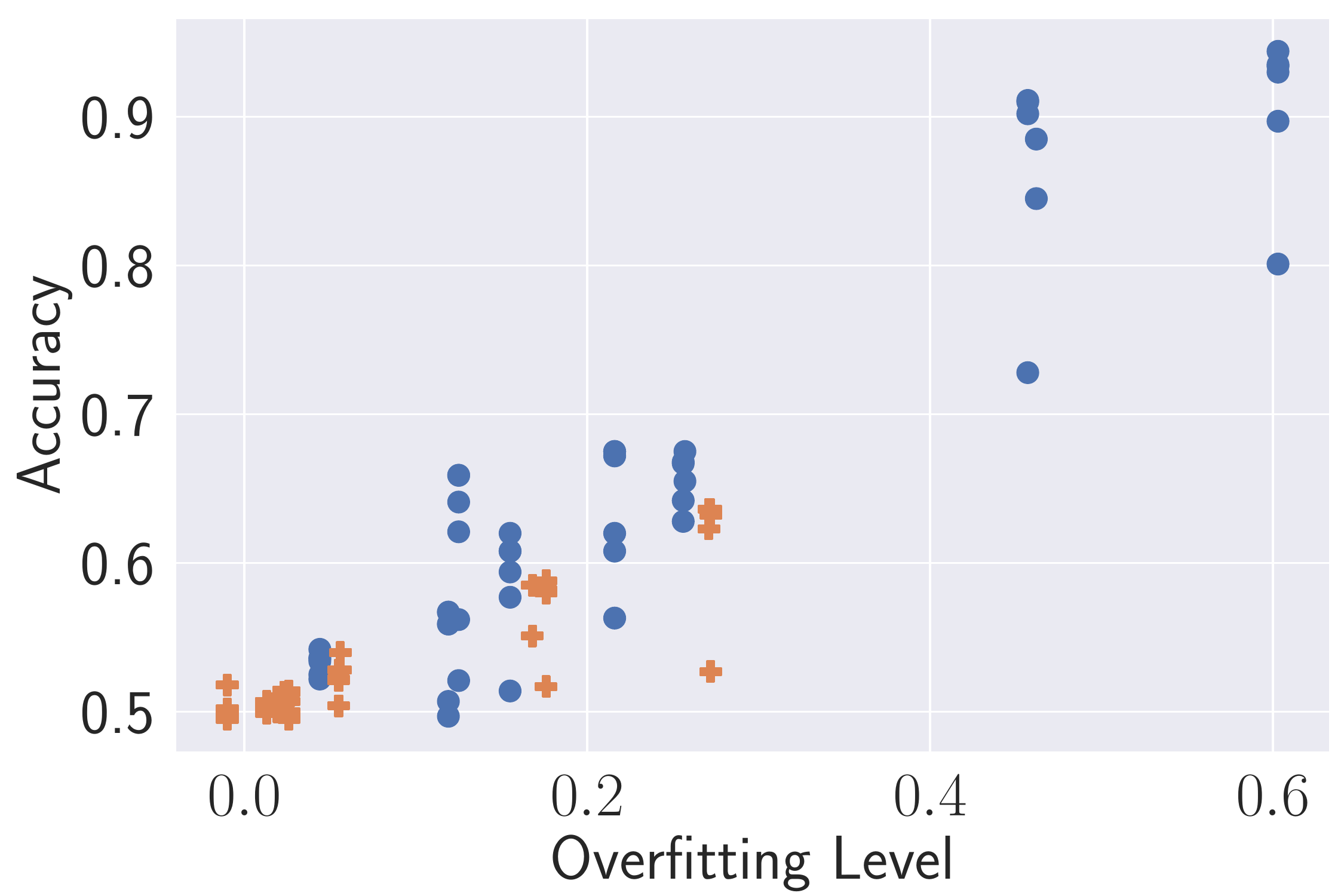}
\caption{ResNet-18}
\label{figure:overfitting_resnet18}
\end{subfigure}
\begin{subfigure}{0.65\columnwidth}
\includegraphics[width=\columnwidth]{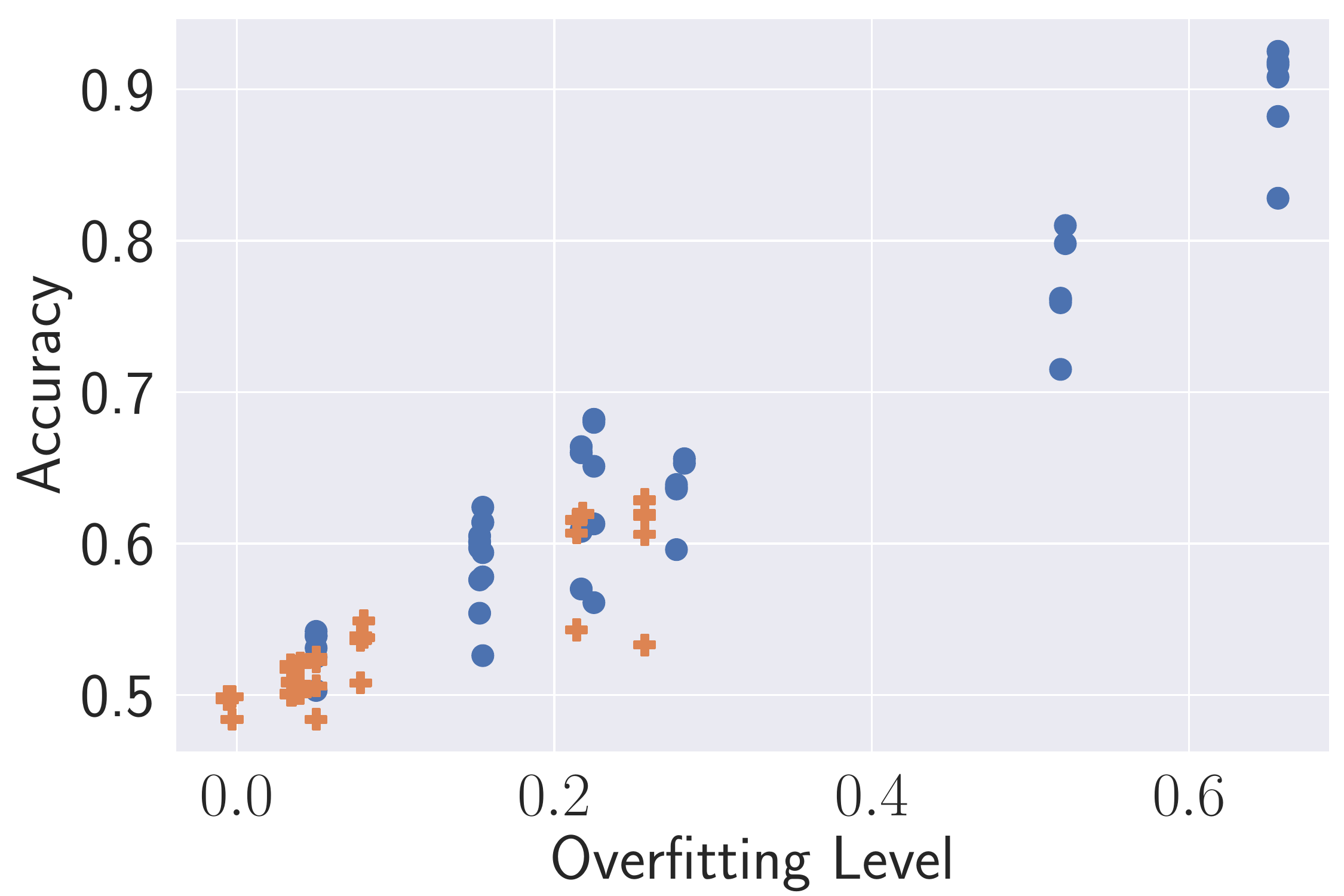}
\caption{ResNet-50}
\label{figure:overfitting_resnet50}
\end{subfigure}
\caption{The performance of membership inference attacks against both supervised models and contrastive models with MobileNetV2, ResNet-18, and ResNet-50 on 5 different datasets under different overfitting levels.
The x-axis represents different overfitting levels.
The y-axis represents membership inference attacks' accuracy.}
\label{figure:overfitting}
\end{figure*}

\begin{figure*}[!t]
\centering
\begin{subfigure}{0.65\columnwidth}
\includegraphics[width=\columnwidth]{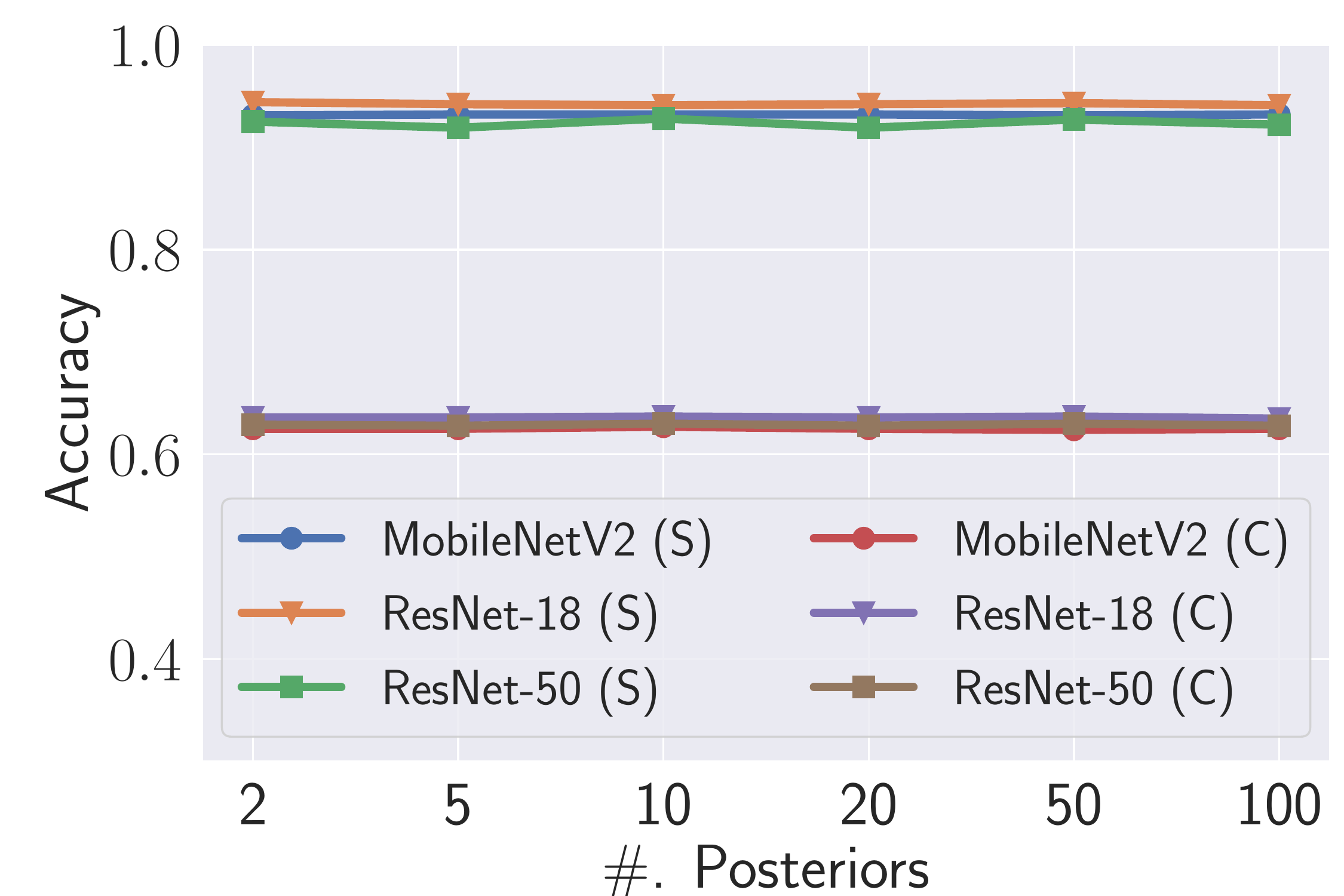}
\caption{NN-based}
\label{figure:diff_posteriors_nn-based}
\end{subfigure}
\begin{subfigure}{0.65\columnwidth}
\includegraphics[width=\columnwidth]{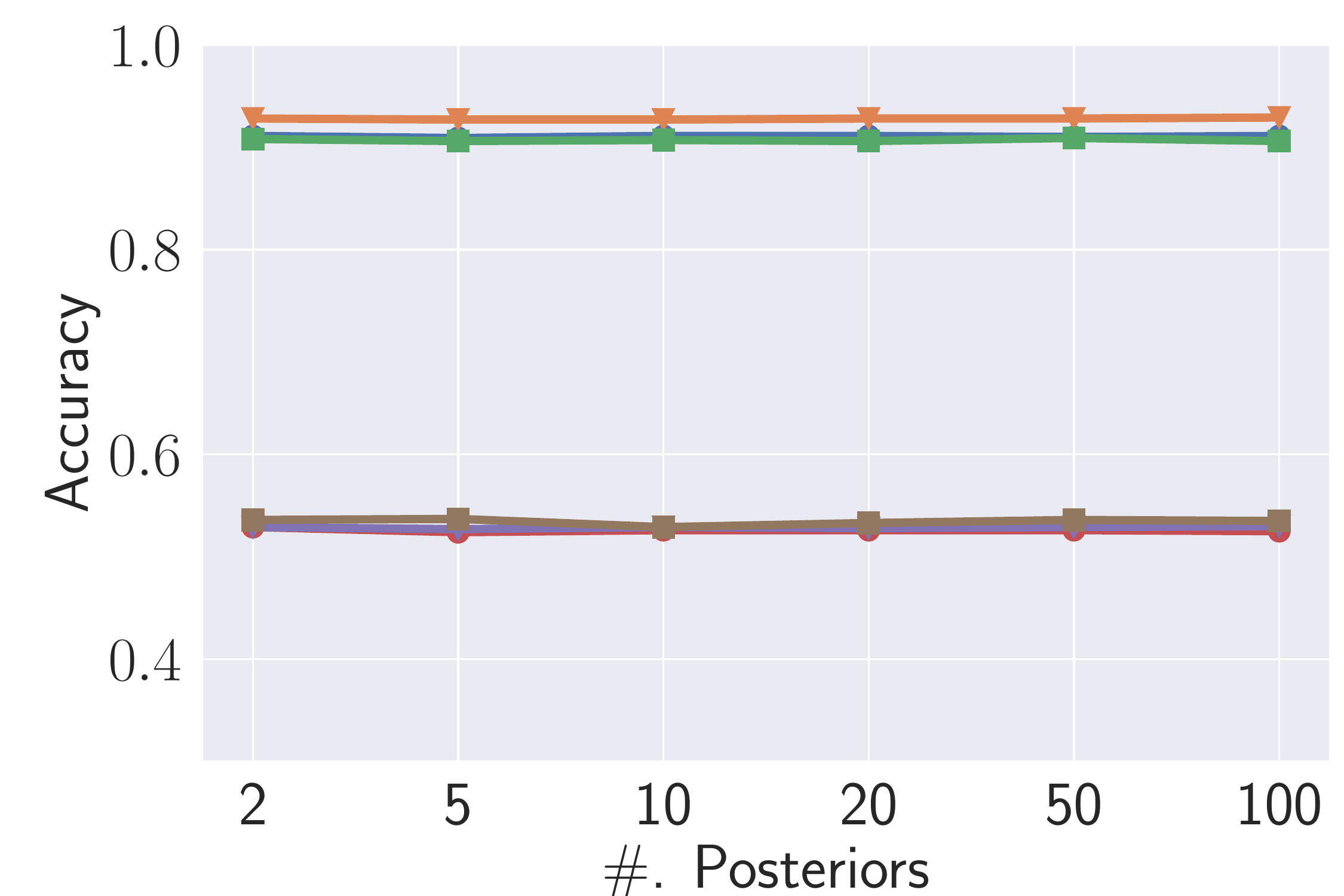}
\caption{Metric-ent}
\label{figure:diff_posteriors_metric-ent}
\end{subfigure}
\begin{subfigure}{0.65\columnwidth}
\includegraphics[width=\columnwidth]{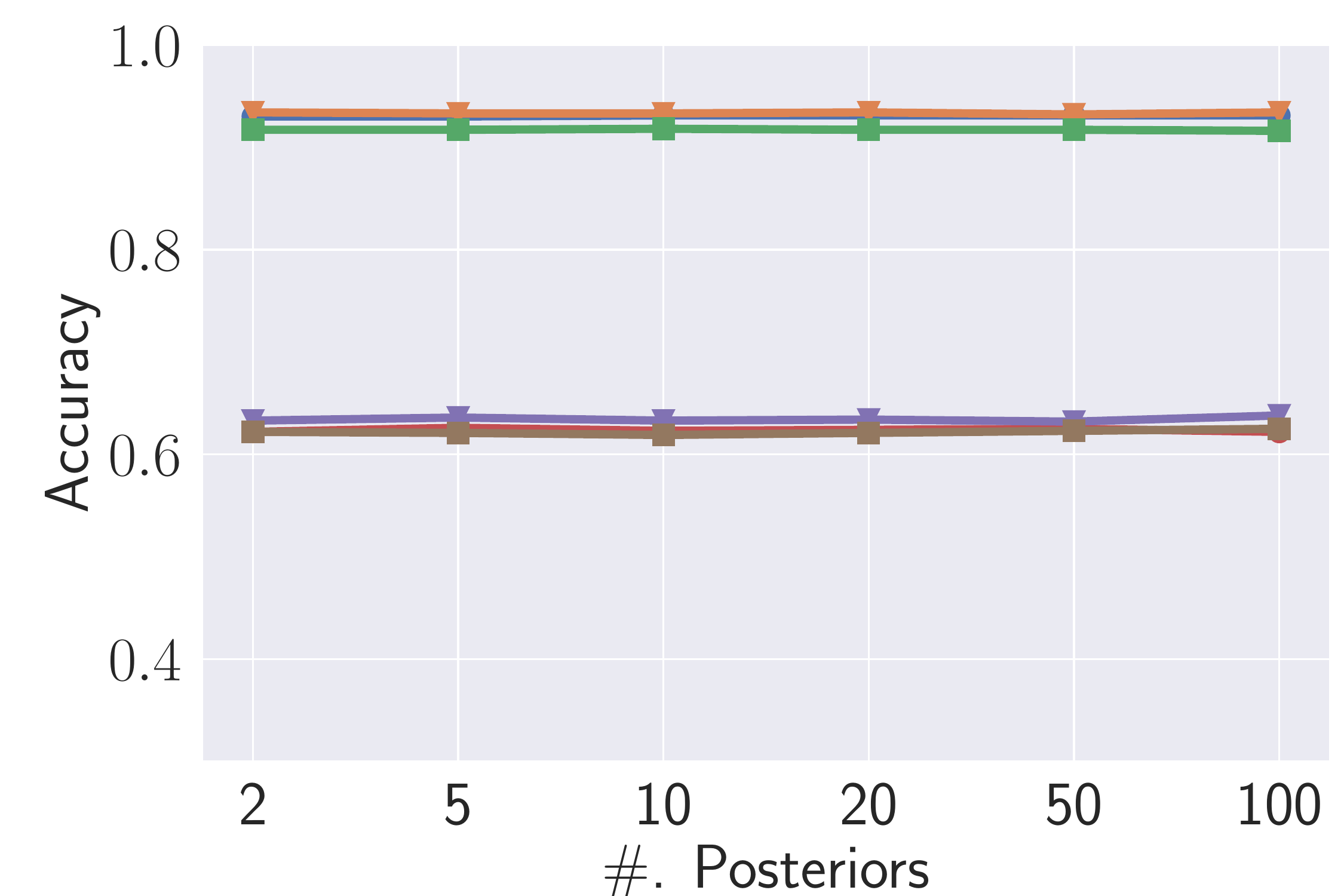}
\caption{Metric-ment}
\label{figure:diff_posteriors_metric-ment}
\end{subfigure}
\caption{The performance of NN-based, metric-ent, and metric-ment attacks against both supervised models and contrastive models with MobileNetV2, ResNet-18, and ResNet-50 on CIFAR100 under different numbers of posteriors given by the target models.
(S) and (C) denotes the supervised and contrastive models, respectively. 
The x-axis represents different numbers of posteriors.
The y-axis represents membership inference attacks' accuracy.
Note that we do not report the performance of metric-corr, metric-conf, and label-only attacks since the number of posteriors does not affect their performance.}
\label{figure:diff_posteriors}
\end{figure*}

\begin{figure}[!t]
\centering
\begin{subfigure}{0.45\columnwidth}
\includegraphics[width=\columnwidth]{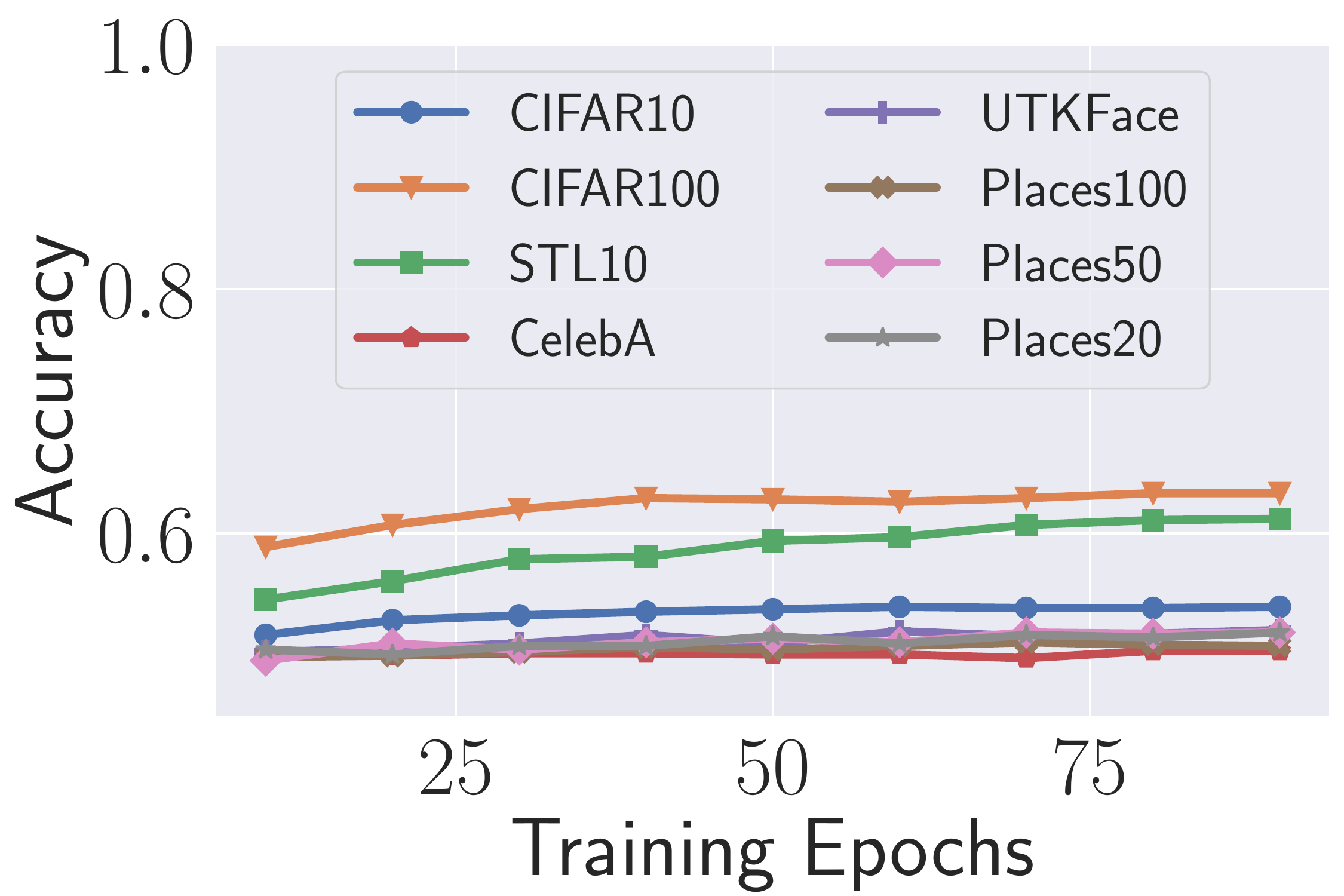}
\caption{NN-based}
\label{figure:diff_linear_nn-based}
\end{subfigure}
\begin{subfigure}{0.45\columnwidth}
\includegraphics[width=\columnwidth]{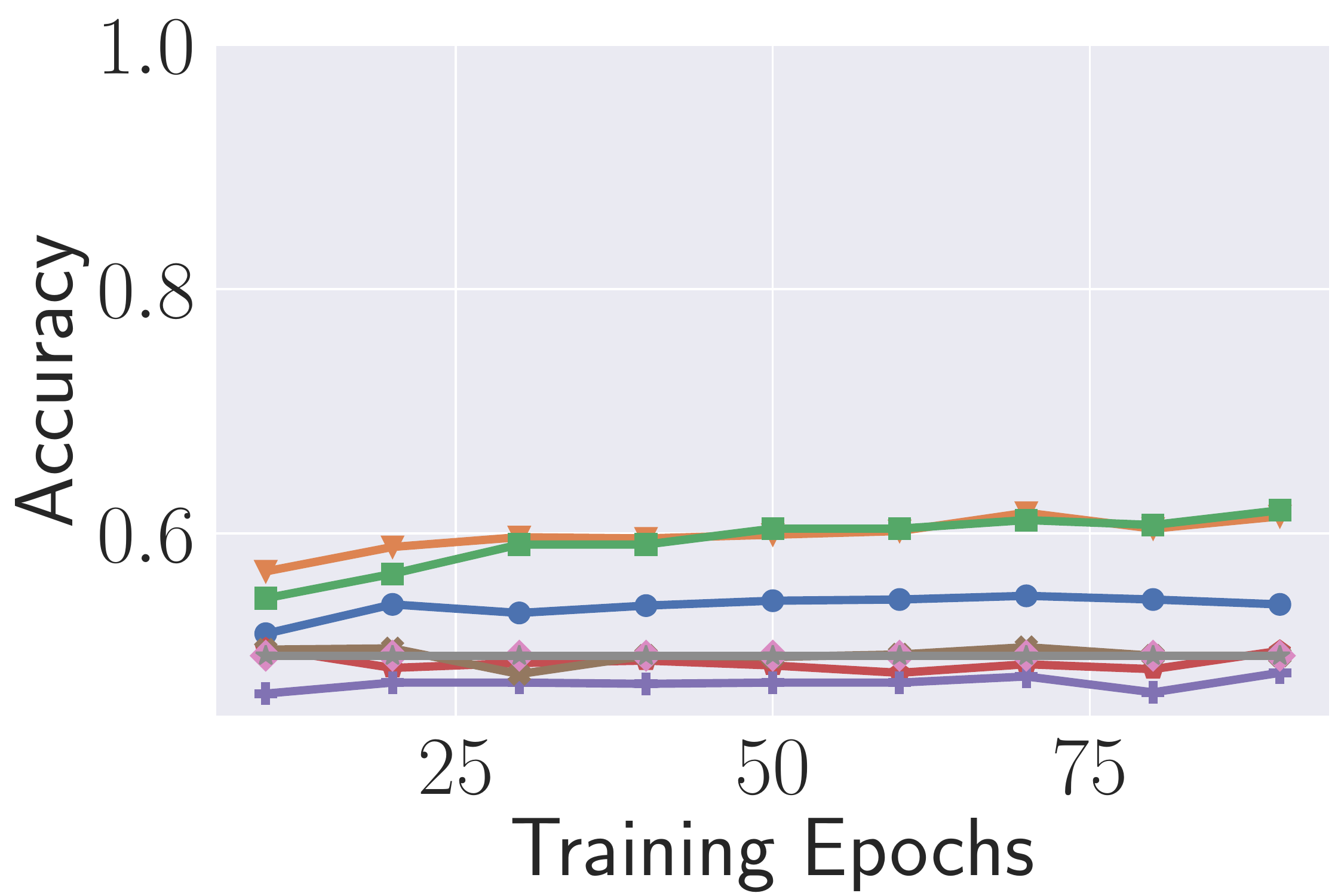}
\caption{Label-only}
\label{figure:diff_linear_label_only}
\end{subfigure}
\caption{The performance of NN-based and label-only membership inference attacks against contrastive models with ResNet-50 on 8 different datasets under different numbers of epochs for classification layer training.
The x-axis represents different numbers of epochs.
The y-axis represents membership inference attacks' accuracy.
Each line corresponds to a specific dataset.}
\label{figure:mia_diff_linear_all}
\end{figure} 

Regarding membership inference against supervised models and contrastive models, the results for MobileNetV2 are shown in \autoref{figure:mia_attack_performance_mobilenet}.
We also summarize the results for ResNet-18 (\autoref{figure:mia_attack_performance_resnet18}) and ResNet-50 (\autoref{figure:mia_attack_performance_resnet50}) in Appendix.
In \autoref{figure:mia_attack_performance_mobilenet}, we see that all the supervised models have higher attack accuracy than the contrastive models.
E.g., when the supervised model is MobileNetV2 trained on CIFAR100, the accuracy of NN-based attack is 0.931, while the accuracy for the corresponding contrastive model is only 0.625.

We observe that NN-based, metric-conf, and metric-ment attacks achieve the best performance in all cases.
The reason metric-conf and metric-ment achieving better performance than metric-corr and metric-ent is that metric-conf and metric-ment consider both prediction correctness and confidence while metric-corr (metric-ent) only considers prediction correctness (confidence).
Interestingly, for supervised models, metric-corr and metric-ent perform similarly, while for contrastive models, metric-ent is worse than metric-corr.
This reason is that the posteriors generated by contrastive models are more smooth compared to supervised model, which makes it harder to distinguish members and non-members through the posterior entropy.
Label-only attacks perform worse than NN-based attacks.
This is expected since the adversary has less information in these cases.
Note that label-only attacks do not perform well on binary classifiers, we will investigate the reason in the future.\footnote{Choquette-Choo et al.~\cite{CTCP20} also only perform label-only membership inference attacks against datasets with more than two classes.}

To further investigate why contrastive models are less vulnerable to membership inference, we analyze the loss distribution between members and non-members in both supervised models and contrastive models.
Due to space limitations, we only show the results of ResNet-18 trained on the CIFAR10 dataset in~\autoref{figure:loss_dsitribution}.
A clear trend is that compared to the contrastive model, the supervised model has a larger divergence between the classification loss (cross-entropy) for members and non-members.
Recall that contrastive learning uses two augmented views of each sample in each epoch to train its base encoder and the original sample to train its classification layer.
This indicates that each sample is generalized to multiple views during the contrastive model training process.
In this way, the contrastive model reduces its memorization of the original sample itself.

Interestingly, Song et al.~\cite{SSM19} observe that defense mechanisms for mitigating adversarial examples~\cite{BCMNSLGR13,PMJFCS16,TKPGBM17,CW17} increase the membership inference performance.
This means such defense and contrastive learning have different effects on membership privacy.
On the one hand, these defense mechanisms for adversarial examples use original samples and their visually imperceptible adversarial examples to train a model; in this way, the model learns to remember each original sample more accurately.
On the other hand, the augmented samples in contrastive learning are very different from their original samples (see \autoref{figure:img_visualize} for some examples).
Therefore, membership inference is less effective against contrastive models.

We notice that the attack performance varies on different models and different datasets.
We relate this to the different overfitting levels.
Similar to previous work~\cite{SSSS17,SZHBFB19}, we measure the overfitting level of a target model by calculating the difference between its training accuracy and testing accuracy.
In~\autoref{figure:overfitting}, we see that the overfitting level is highly correlated with the attack performance: if a model is more overfitted, it is more vulnerable to membership inference attacks.
For instance, in \autoref{figure:overfitting_mobilenet}, on CIFAR100, the contrastive model (upper right orange cross) has an overfitting level of 0.249, and the NN-based attack accuracy is 0.625, while the supervised model (upper right blue dot) has a larger overfitting level (0.678) and higher attack accuracy (0.931).
Another observation is that compared to the supervised models, the overfitting levels of the contrastive models reside in a smaller range.

NN-based method as well as some of the metric-based ones (metric-ent and metric-ment) require the target model to provide posteriors to launch the attacks.
We further investigate whether the number of posteriors provided by the target model can influence the attack performance.
Concretely, we vary the number of posteriors from 2 to 100 on CIFAR100 for both supervised and contrastive models.
\autoref{figure:diff_posteriors} shows that the number of posteriors does not have a strong influence on the attack performance.
We further measure the influence of the number of epochs used for training each contrastive model's classification layer on the attack performance.
\autoref{figure:mia_diff_linear_all} shows that the attack accuracy is rather stable (the performance of metric-based attacks are summarized in \autoref{figure:mia_diff_linear_metric-based} in Appendix).
These results show that contrastive models consistently reduce the membership threat.

In conclusion, contrastive models are less vulnerable to membership inference attacks compared to supervised models.
The reason is that contrastive models are less overfitted to their training samples than supervised models due to the design of the contrastive learning paradigm.

% ----------------------------------------------------
\section{Attribute Inference Attack}
\label{section:AI}
% ----------------------------------------------------

In this section, we take a different angle to measure the privacy risks of contrastive learning using attribute inference attack~\cite{MSCS19,SS20}.
Similar to membership inference attacks, we use existing attribute inference attacks~\cite{MSCS19,SS20} to measure the contrastive model's privacy risks instead of inventing new methods.

\begin{figure*}[!t]
\centering
\begin{subfigure}{0.5\columnwidth}
\includegraphics[width=\columnwidth]{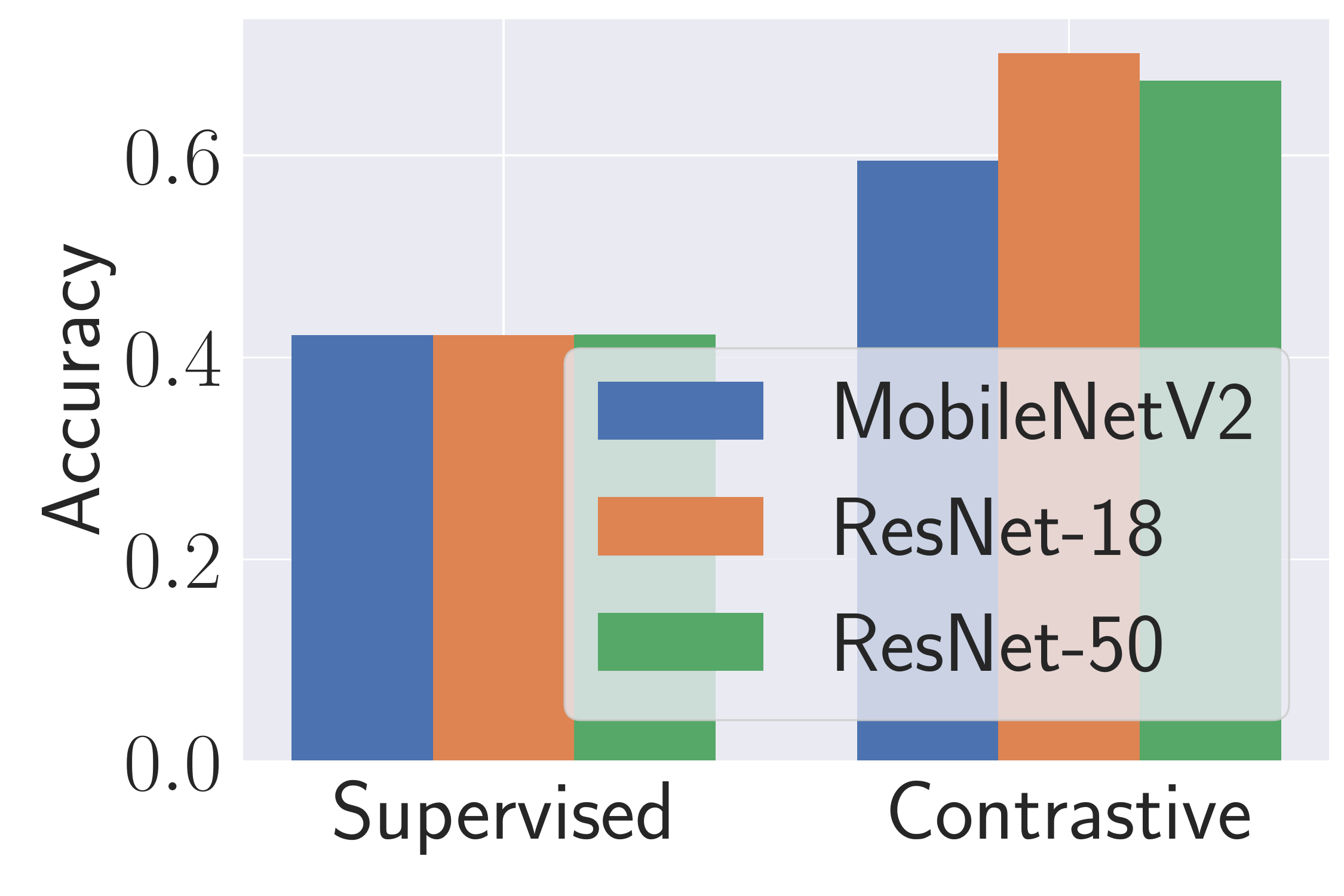}
\caption{UTKFace}
\label{figure:ai_performance_utkface}
\end{subfigure}
\begin{subfigure}{0.5\columnwidth}
\includegraphics[width=\columnwidth]{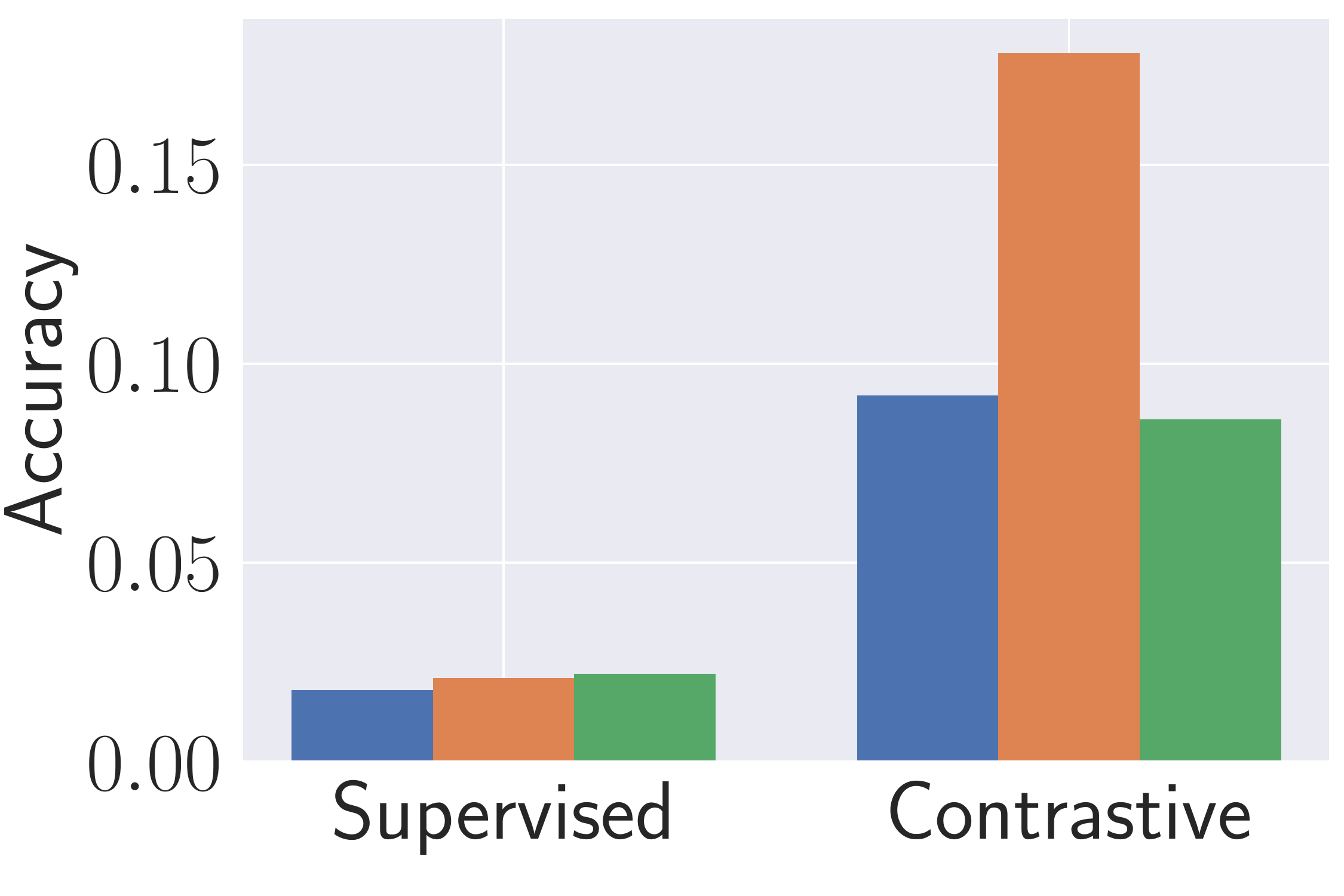}
\caption{Places100}
\label{figure:ai_performance_place100}
\end{subfigure}
\begin{subfigure}{0.5\columnwidth}
\includegraphics[width=\columnwidth]{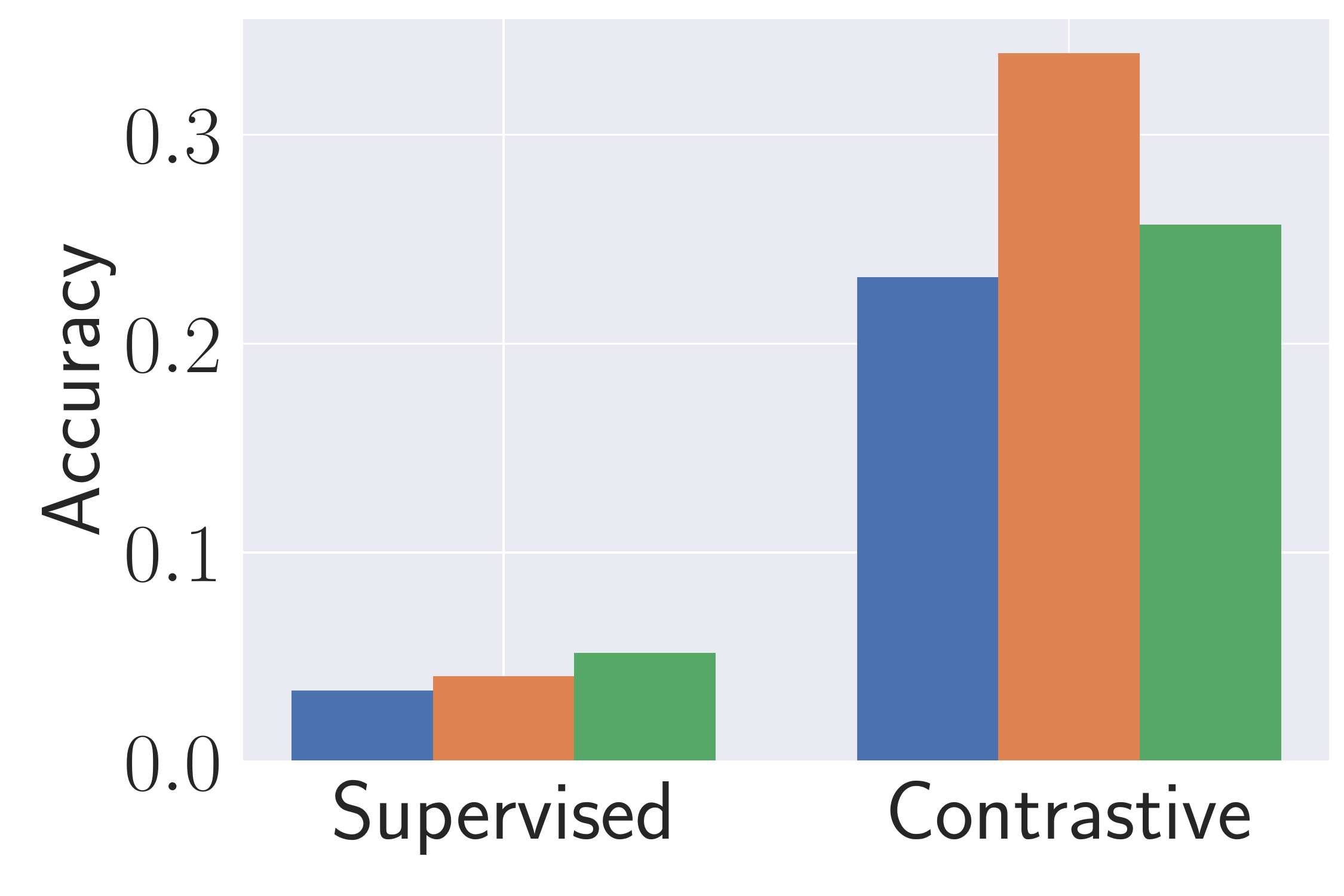}
\caption{Places50}
\label{figure:ai_performance_place50}
\end{subfigure}
\begin{subfigure}{0.5\columnwidth}
\includegraphics[width=\columnwidth]{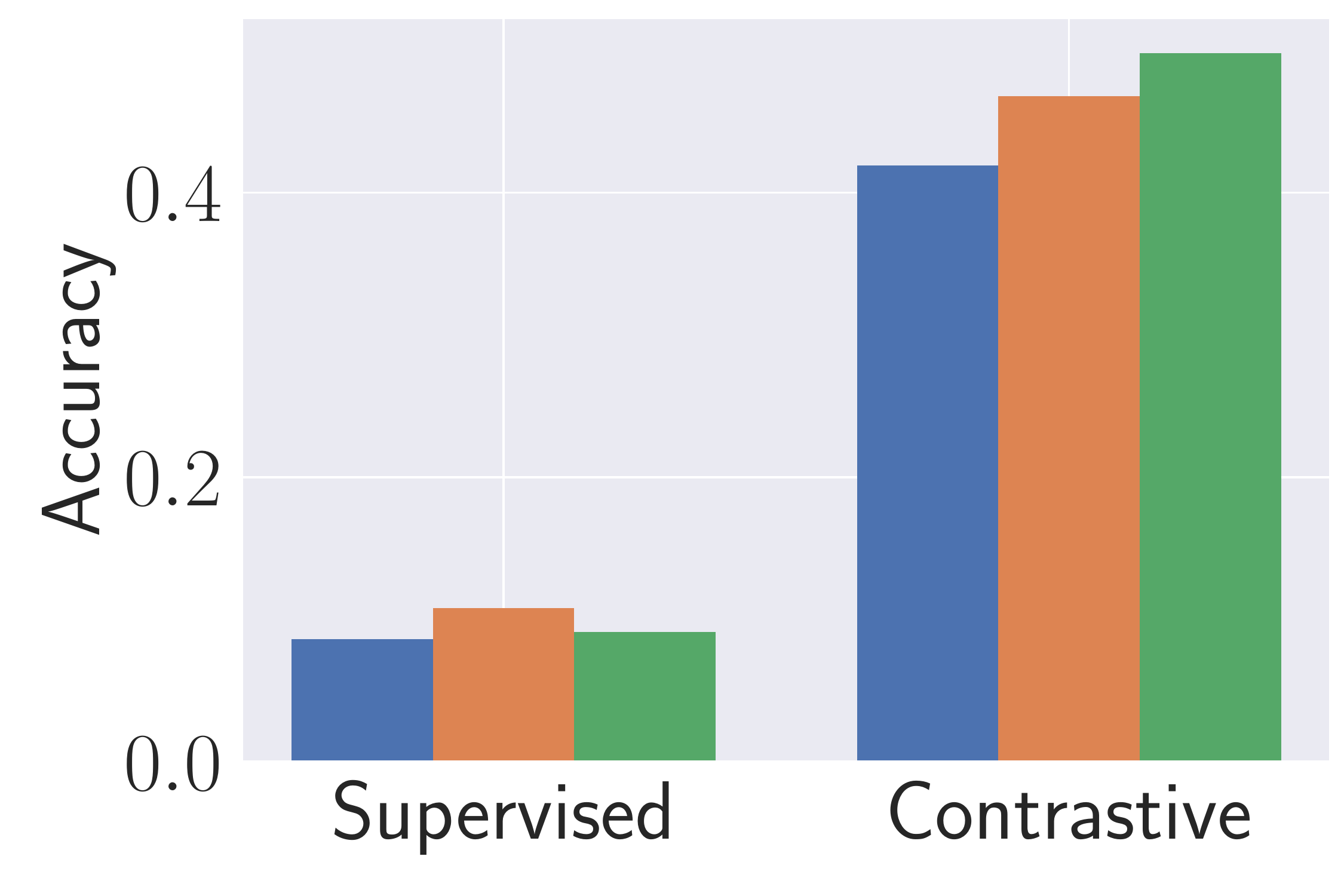}
\caption{Places20}
\label{figure:ai_performance_place20}
\end{subfigure}
\caption{The performance of attribute inference attacks against both supervised models and contrastive models with MobileNetV2, ResNet-18, and ResNet-50 on 4 different datasets.
The x-axis represents different models.
The y-axis represents attribute inference attacks' accuracy.}
\label{figure:ai_performance}
\end{figure*}

% ----------------------------------------------------
\subsection{Attack Definition and Threat Model}
\label{subsection:AIThreatModel}
% ----------------------------------------------------

In attribute inference, the adversary's goal is to infer a specific sensitive attribute of a data sample from its representation generated by a target model.
This sensitive attribute is not related to the target ML model's original classification task.
For instance, a target model is designed to classify an individual's age from their social network posts, while attribute inference aims to infer their educational background.

Attribute inference attacks have been successfully performed on supervised models~\cite{MSCS19,SS20}.
The reason behind this is the intrinsic overlearning property of ML models.
Overlearning means that an ML model trained for a certain task may also learn to represent other characteristics of data samples.
Such representation capability, in some cases, can be exploited by an adversary to infer data samples' sensitive attributes.

Once a contrastive model is trained, it can generate a representation for each sample with its base encoder $\Encoder$.
For a supervised model, we consider the whole model without the classification layer as its base encoder to generate a representation for each sample.
Note that the base encoder of contrastive model and supervised model has the same architecture.
 
For attribute inference, given a data sample's representation from a target model, denoted by $\RepresentationVector$, to conduct the attribute inference attack, the adversary trains an attack model $\AIAttackModel: \RepresentationVector \mapsto \SensitiveAttribute$,
where $\SensitiveAttribute$ represents the sensitive attribute.

We follow the same threat model as previous work~\cite{MSCS19,SS20}: the adversary only has access to the target sample’s embedding (representation), but not the target sample itself.
The adversary is also assumed to have a set of samples' embeddings and their sensitive attributes; this dataset is termed as an auxiliary dataset $\AuxDataset$.
As shown by previous work, attribute inference can be applied in both federated learning~\cite{MSCS19} and model partitioning~\cite{SS20} settings.

% ----------------------------------------------------
\subsection{Methodology}
\label{subsection:AIMethodology}
% ----------------------------------------------------

We generalize the methodology of attribute inference attacks against supervised models~\cite{MSCS19,SS20} to contrastive models.
The attack process can be partitioned into two stages, i.e., attack model training and attribute inference.

\mypara{Attack Model Training}
For each $(\RepresentationVector,\SensitiveAttribute) \in \AuxDataset$, the adversary takes the representation $\RepresentationVector$ as the input and the corresponding sensitive attribute $\SensitiveAttribute$ as the label to train the attack model.

\mypara{Attribute Inference}
To determine the sensitive attribute of a data sample's representation $\RepresentationVector$, the adversary queries the attack model $\AIAttackModel$ with $\RepresentationVector$ and obtains the result.

% ----------------------------------------------------
\subsection{Experimental Setting}
\label{subsection:AIExperimentalSetting}
% ----------------------------------------------------

\begin{figure*}[!t]
\centering
\begin{subfigure}{0.45\columnwidth}
\includegraphics[width=\columnwidth]{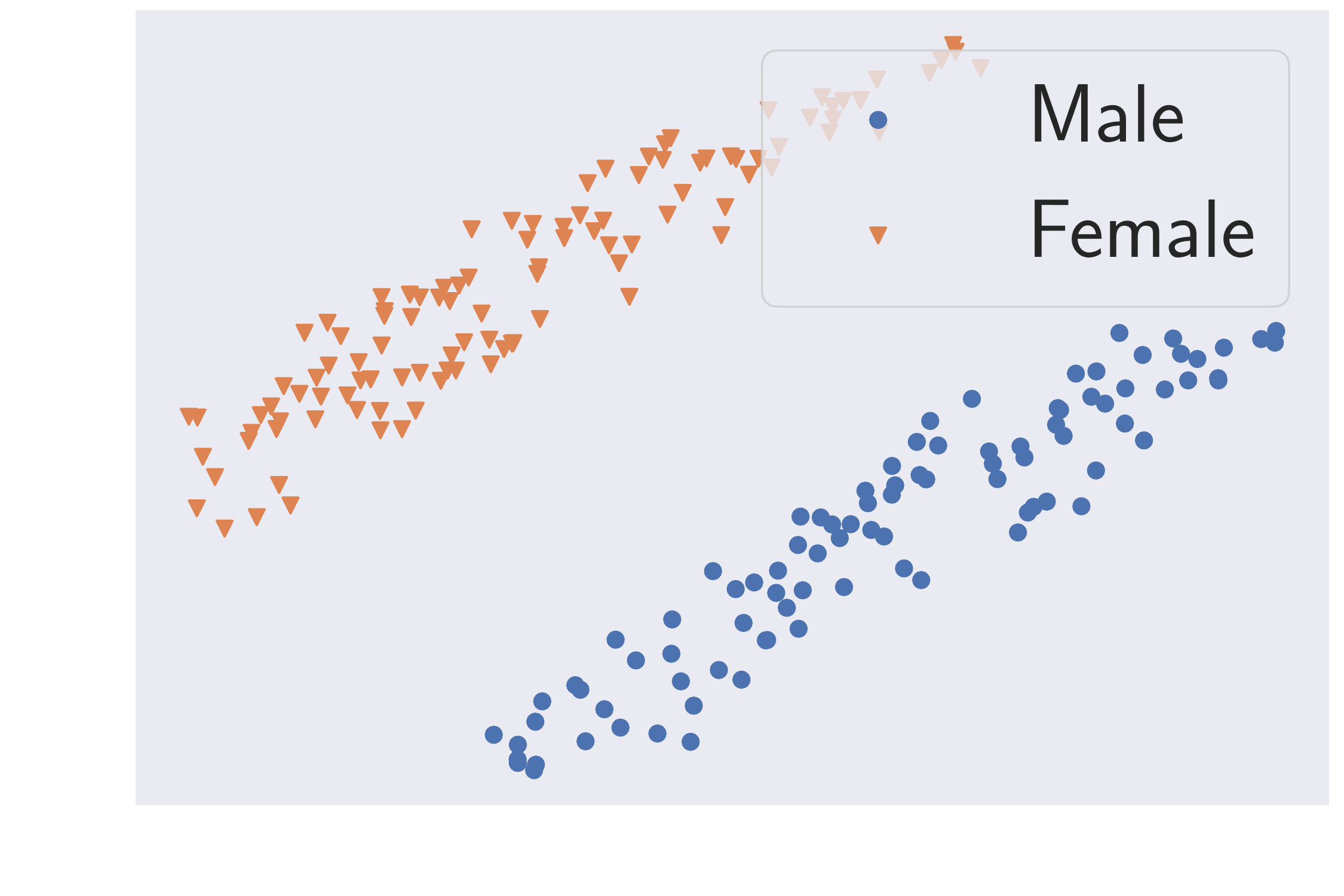}
\caption{Original Task (S)}
\label{figure:tsne_1}
\end{subfigure}
\begin{subfigure}{0.45\columnwidth}
\includegraphics[width=\columnwidth]{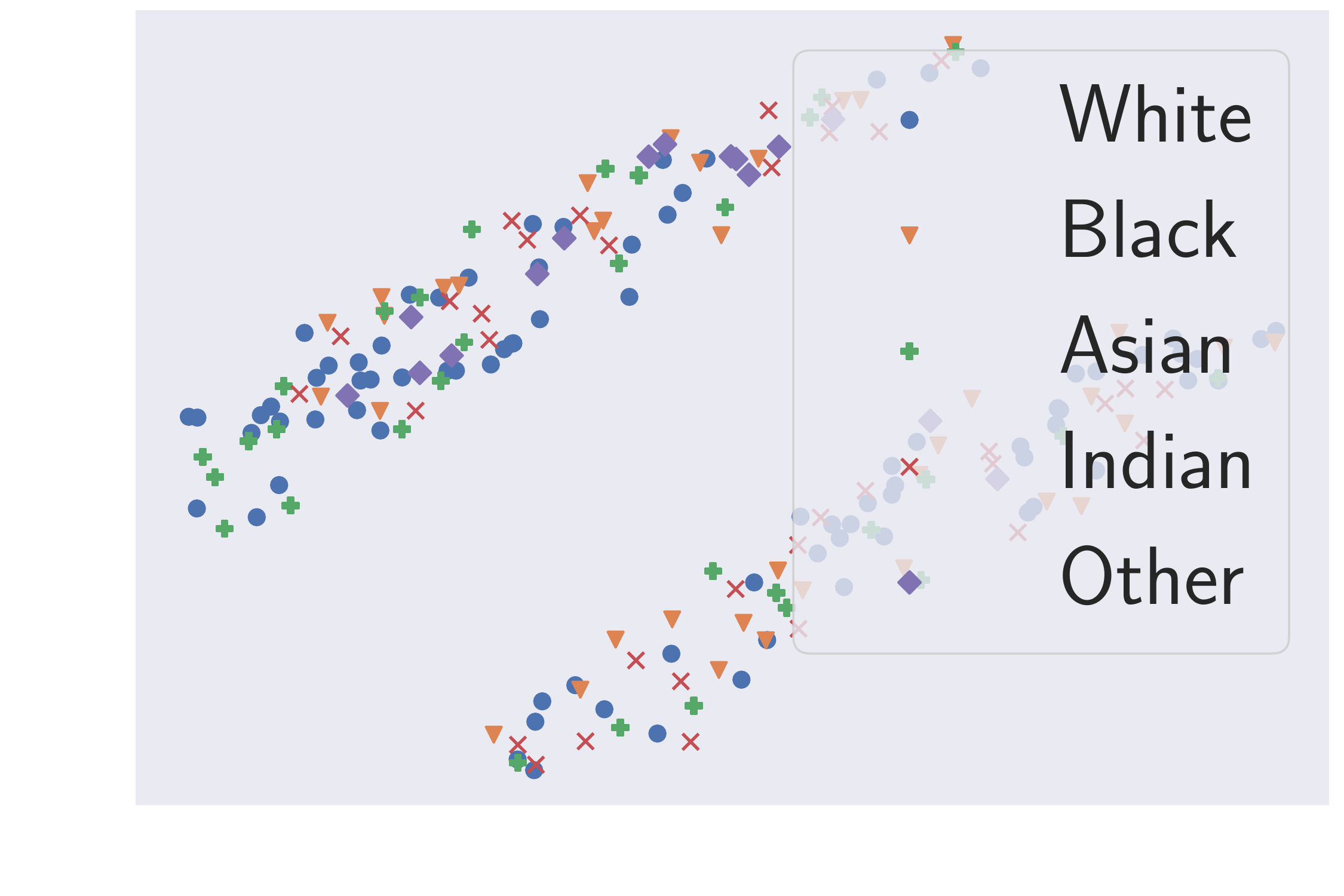}
\caption{Sensitive Attribute (S)}
\label{figure:tsne_2}
\end{subfigure}
\begin{subfigure}{0.45\columnwidth}
\includegraphics[width=\columnwidth]{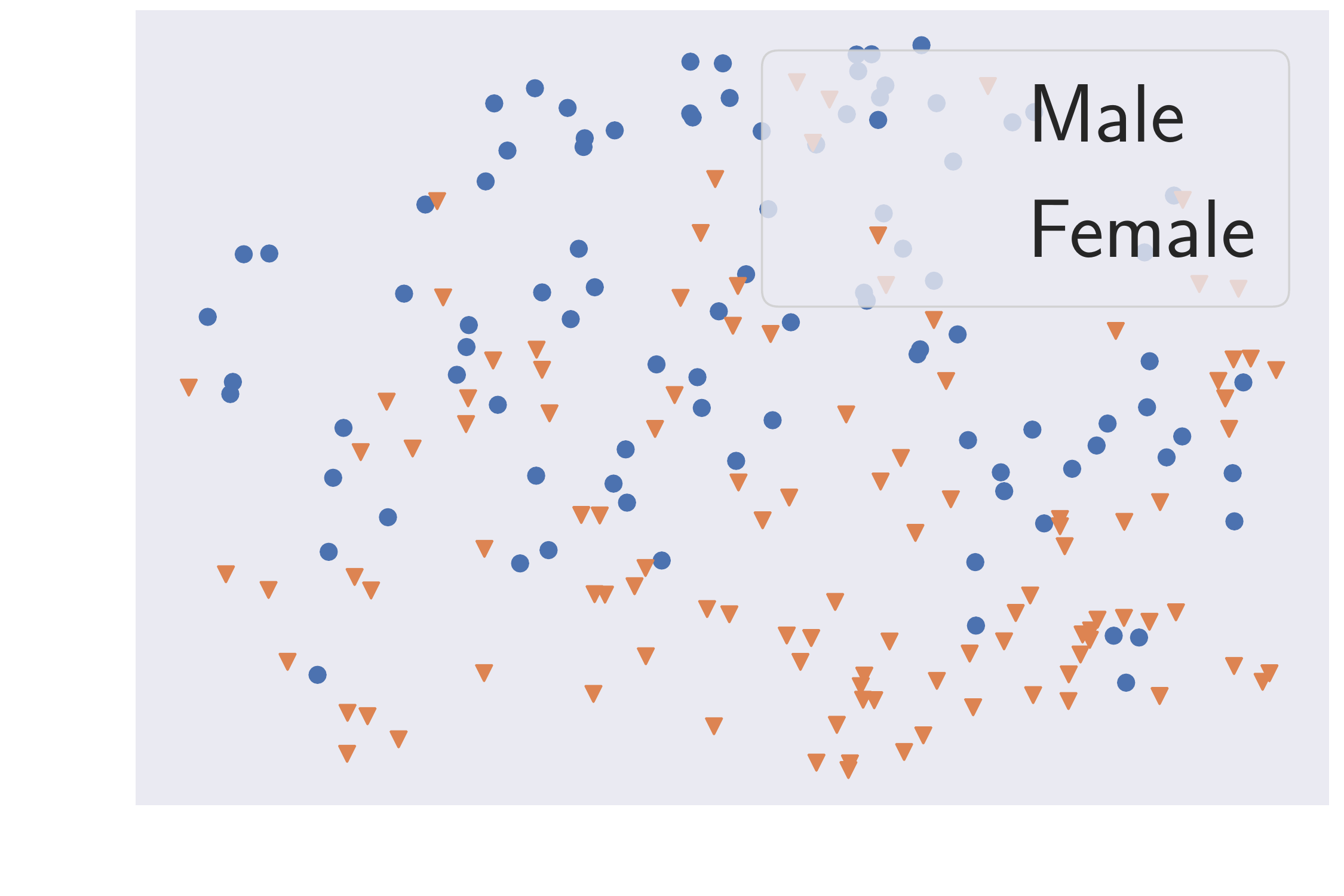}
\caption{Original Task (C)}
\label{figure:tsne_3}
\end{subfigure}
% x
\begin{subfigure}{0.45\columnwidth}
\includegraphics[width=\columnwidth]{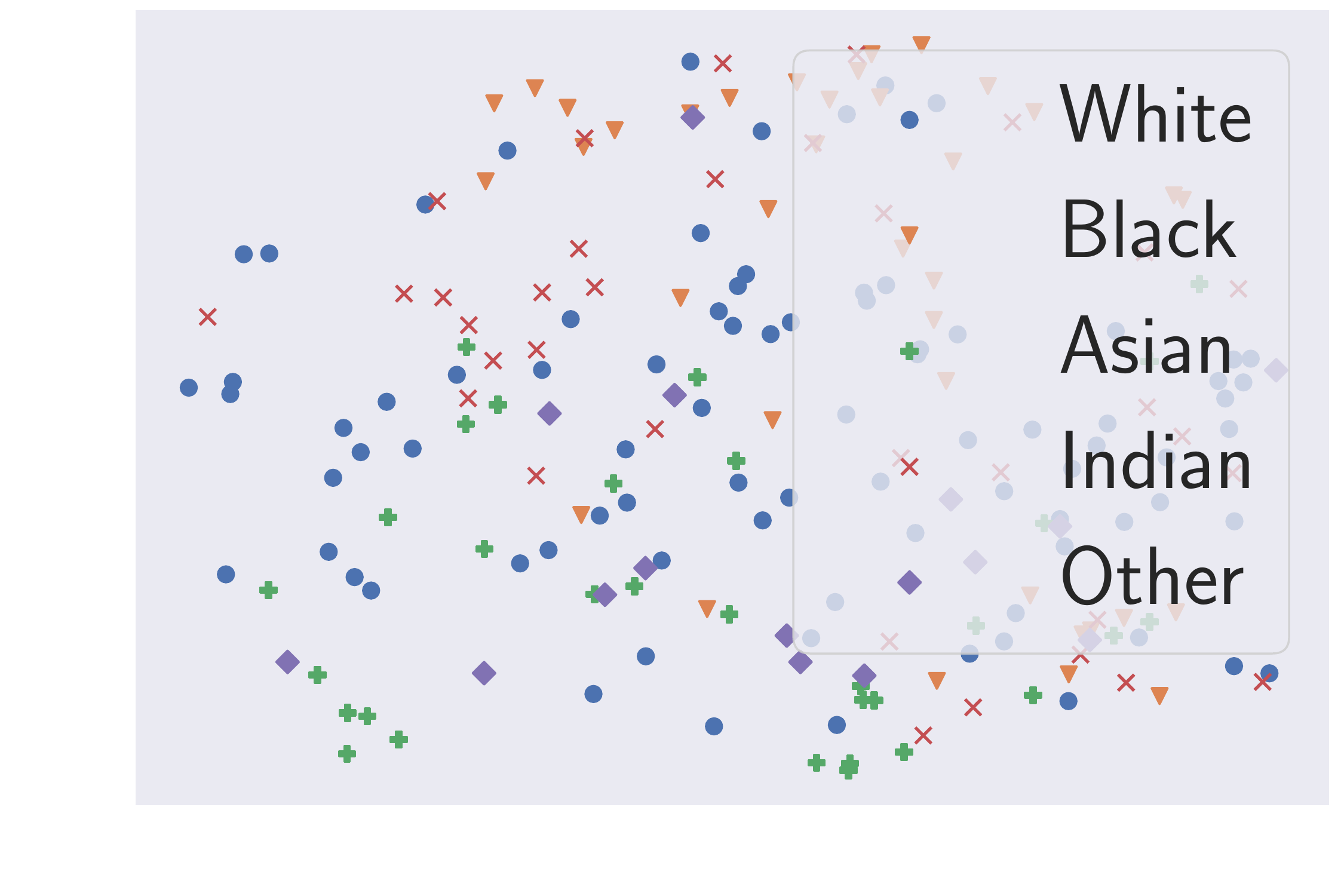}
\caption{Sensitive (C)}
\label{figure:tsne_4}
\end{subfigure}
\caption{The representations for 200 randomly selected samples generated by both the supervised model and the contrastive model with ResNet-18 on UTKFace projected into a 2-dimension space using t-SNE.
(S) and (C) denotes the supervised and contrastive models, respectively.
Each point represents a sample.}
\label{figure:images}
\end{figure*}

\begin{figure*}[!t]
\centering
\begin{subfigure}{0.45\columnwidth}
\includegraphics[width=\columnwidth]{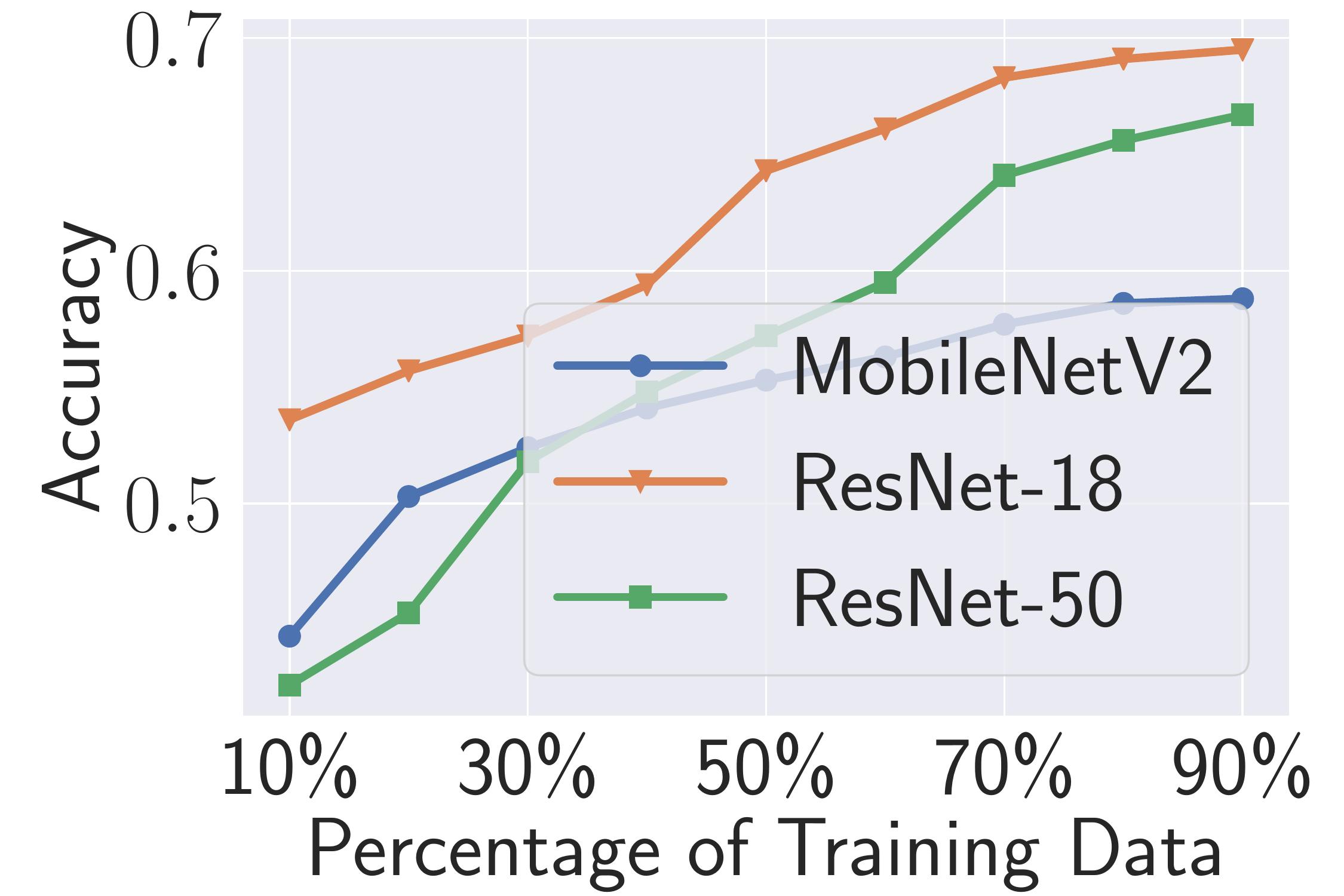}
\caption{UTKFace}
\label{figure:ai_diff_ratio_utkface}
\end{subfigure}
\begin{subfigure}{0.45\columnwidth}
\includegraphics[width=\columnwidth]{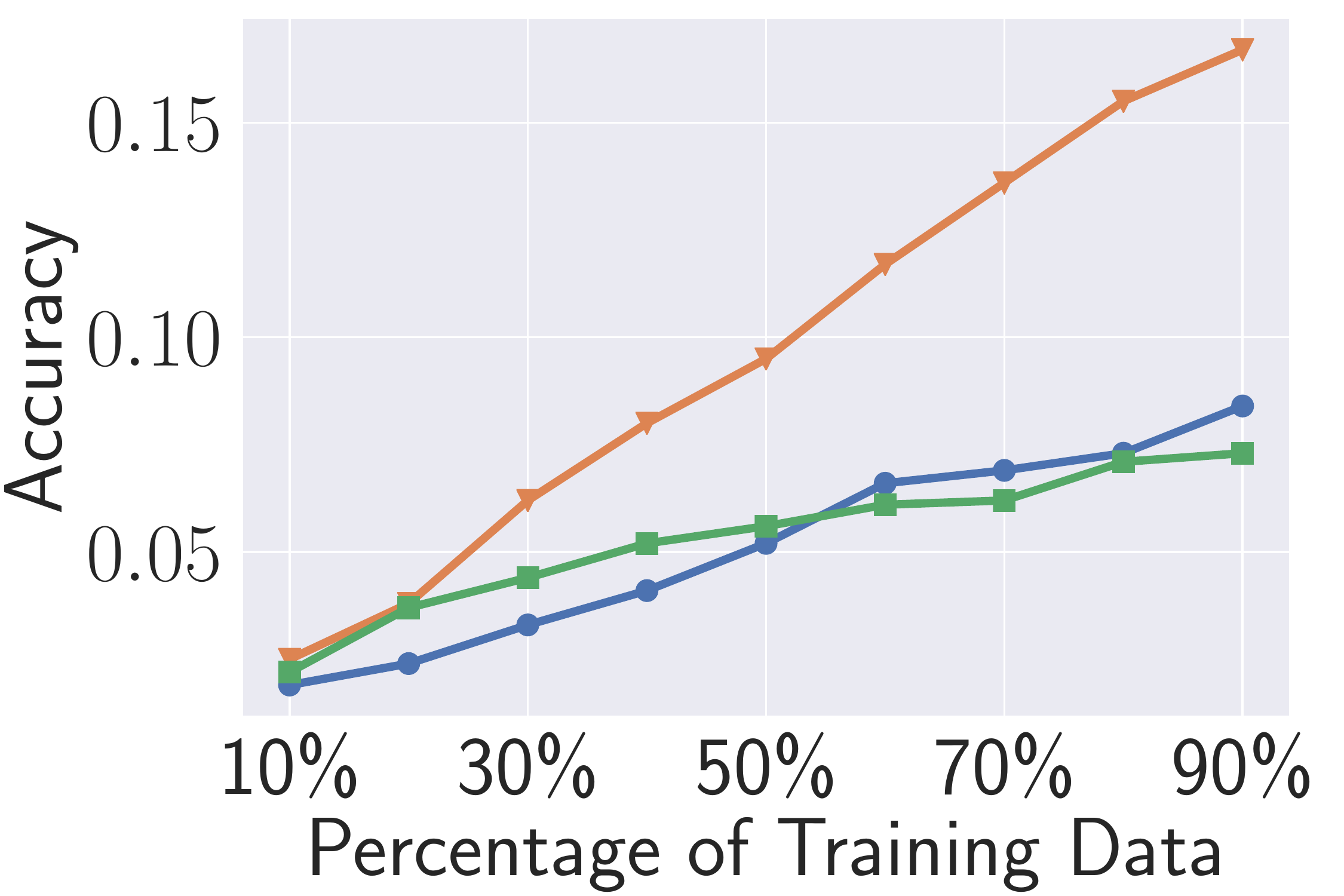}
\caption{Places100}
\label{figure:ai_diff_ratio_place100}
\end{subfigure}
\begin{subfigure}{0.45\columnwidth}
\includegraphics[width=\columnwidth]{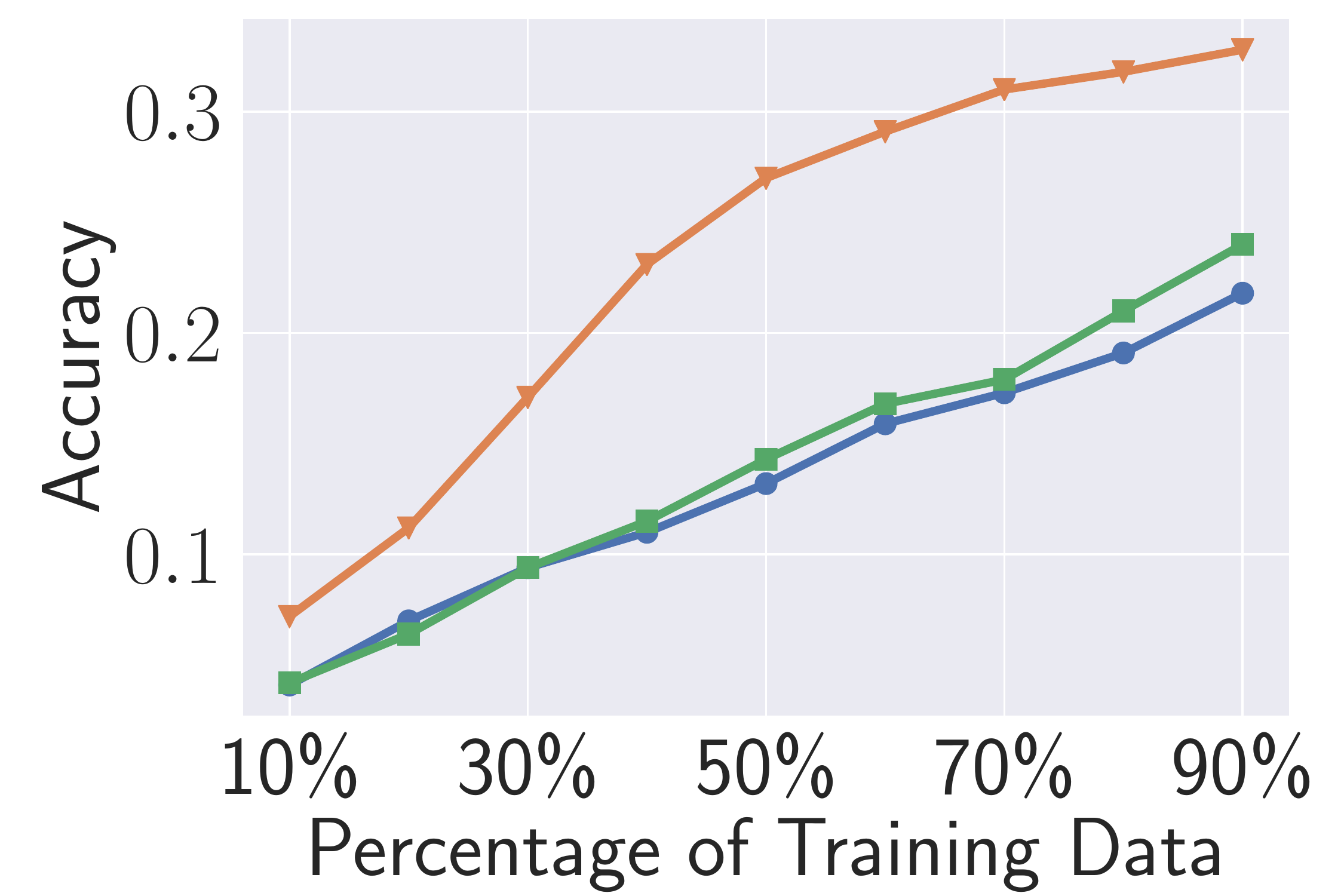}
\caption{Places50}
\label{figure:ai_diff_ratio_place50}
\end{subfigure}
\begin{subfigure}{0.45\columnwidth}
\includegraphics[width=\columnwidth]{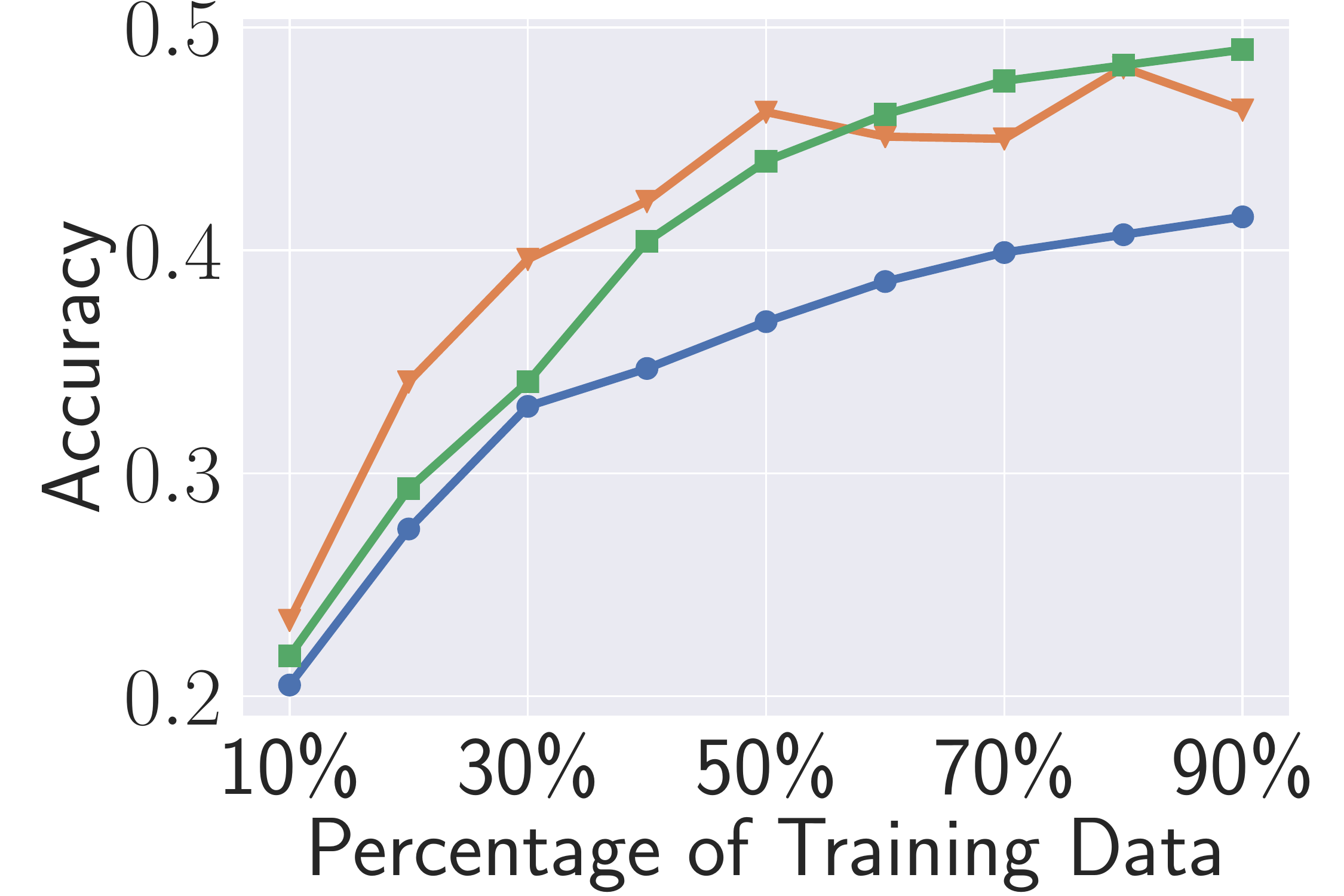}
\caption{Places20}
\label{figure:ai_diff_ratio_place20}
\end{subfigure}
\caption{The performance of attribute inference attacks against contrastive models on 4 different datasets under different percentages of the attack training dataset.
The x-axis represents different percentages of the attack training dataset.
The y-axis represents attribute inference attacks' accuracy.}
\label{figure:ai_diff_ratio} 
\end{figure*}

\begin{figure*}[!t]
\centering
\begin{subfigure}{0.45\columnwidth}
\includegraphics[width=\columnwidth]{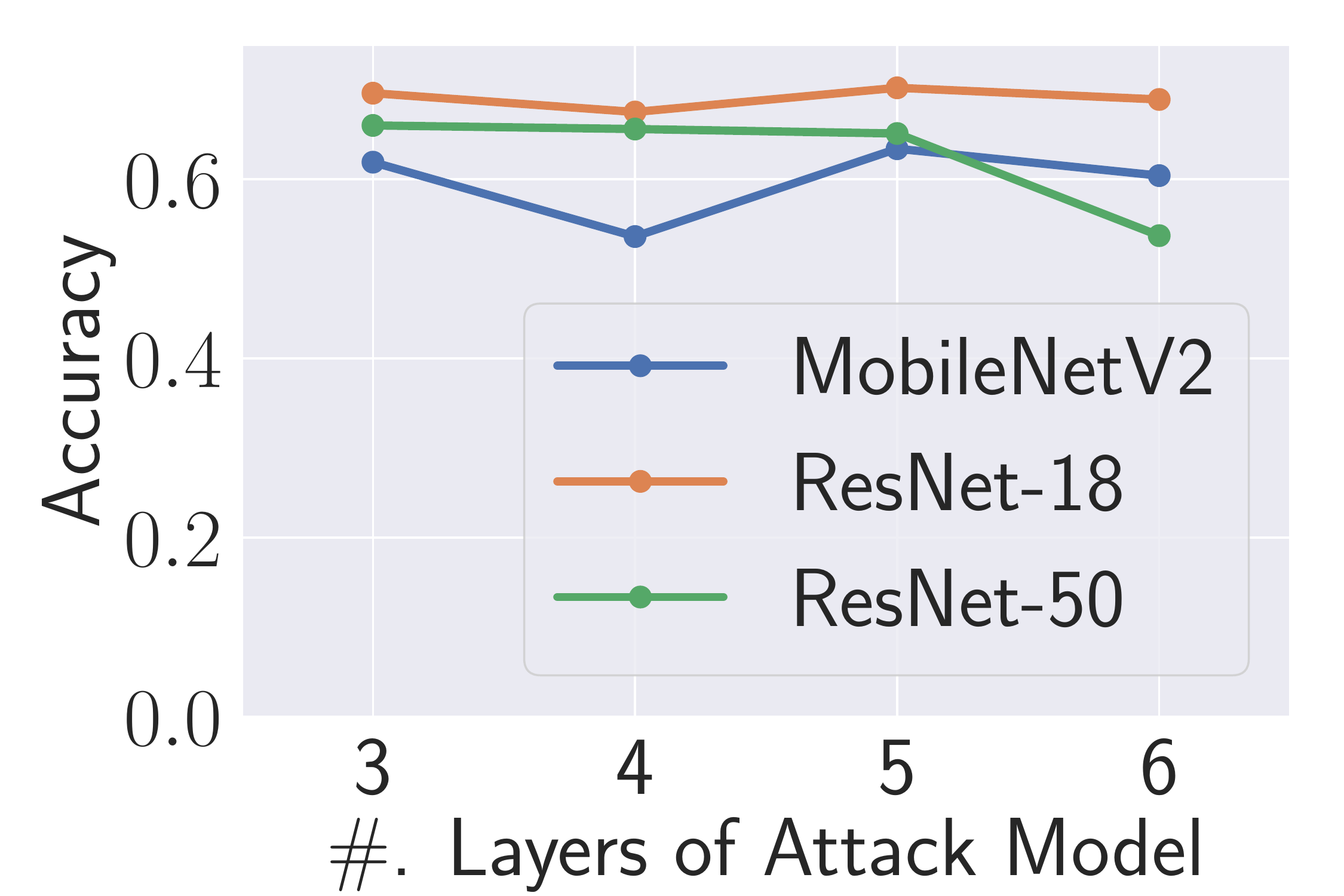}
\caption{UTKFace}
\label{figure:ai_diff_attack_layer_utkface_SimCLR}
\end{subfigure}
\begin{subfigure}{0.45\columnwidth}
\includegraphics[width=\columnwidth]{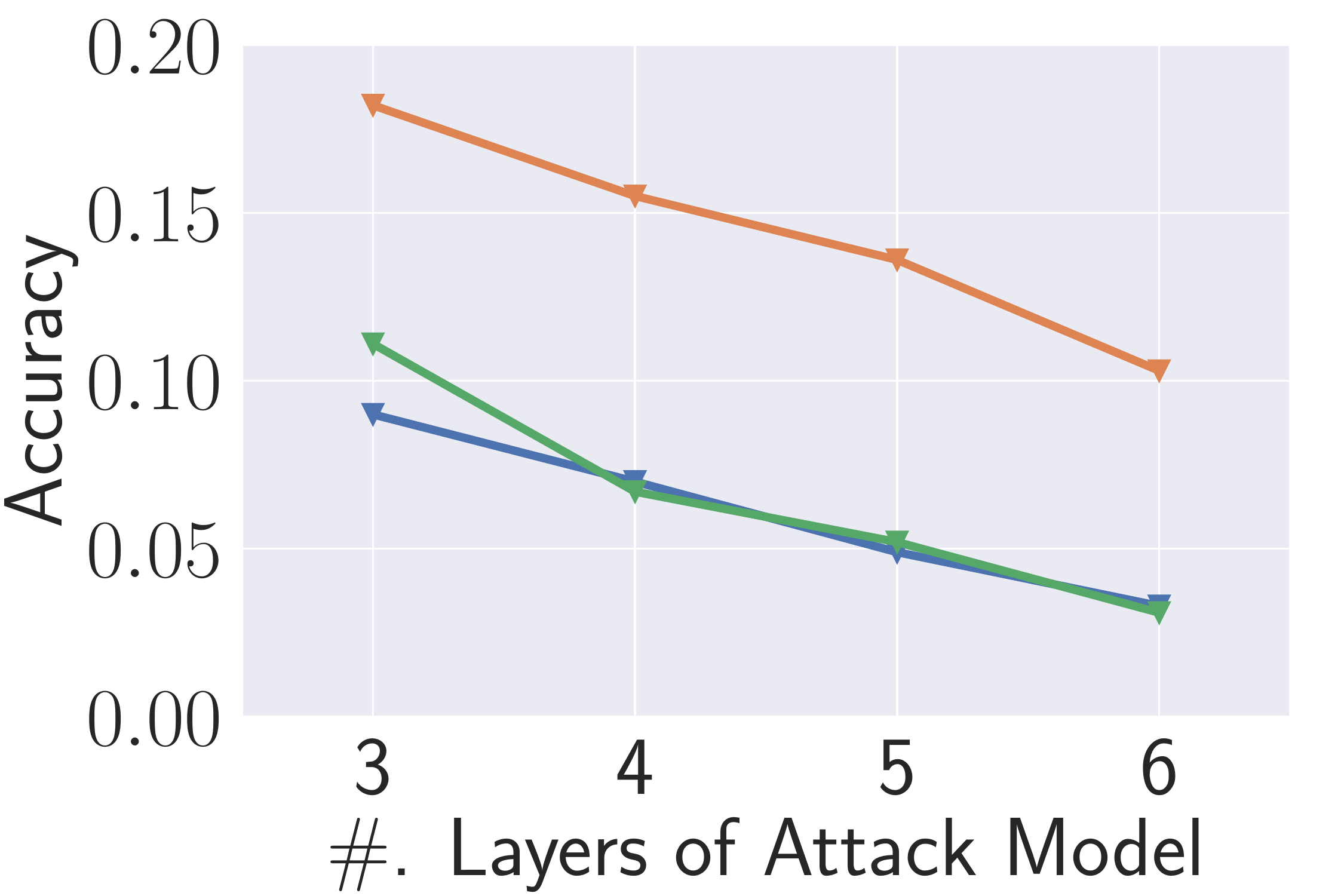}
\caption{Places100}
\label{figure:ai_diff_attack_layer_place100_SimCLR}
\end{subfigure}
\begin{subfigure}{0.45\columnwidth}
\includegraphics[width=\columnwidth]{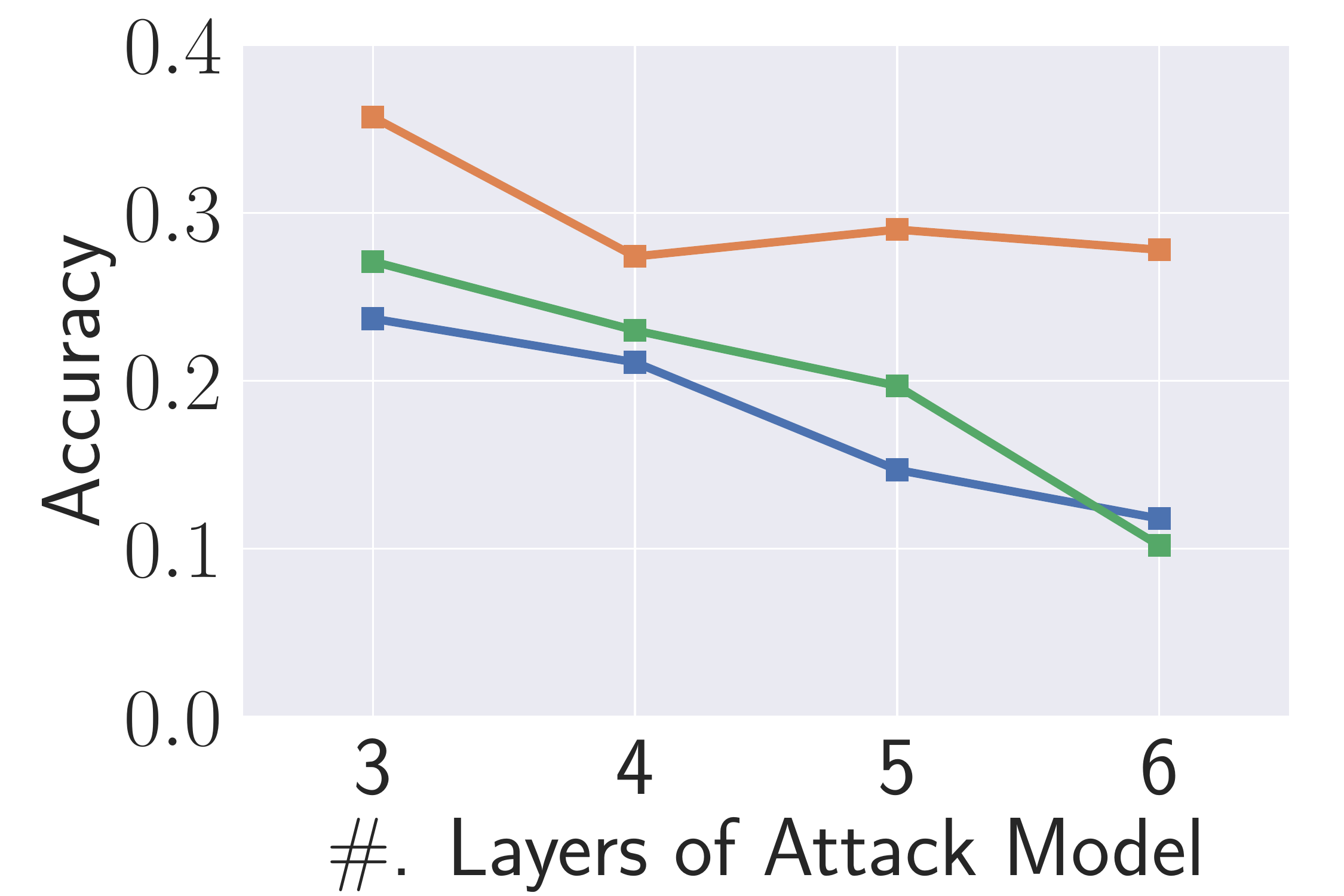}
\caption{Places50}
\label{figure:ai_diff_attack_layer_place50_SimCLR}
\end{subfigure}
\begin{subfigure}{0.45\columnwidth}
\includegraphics[width=\columnwidth]{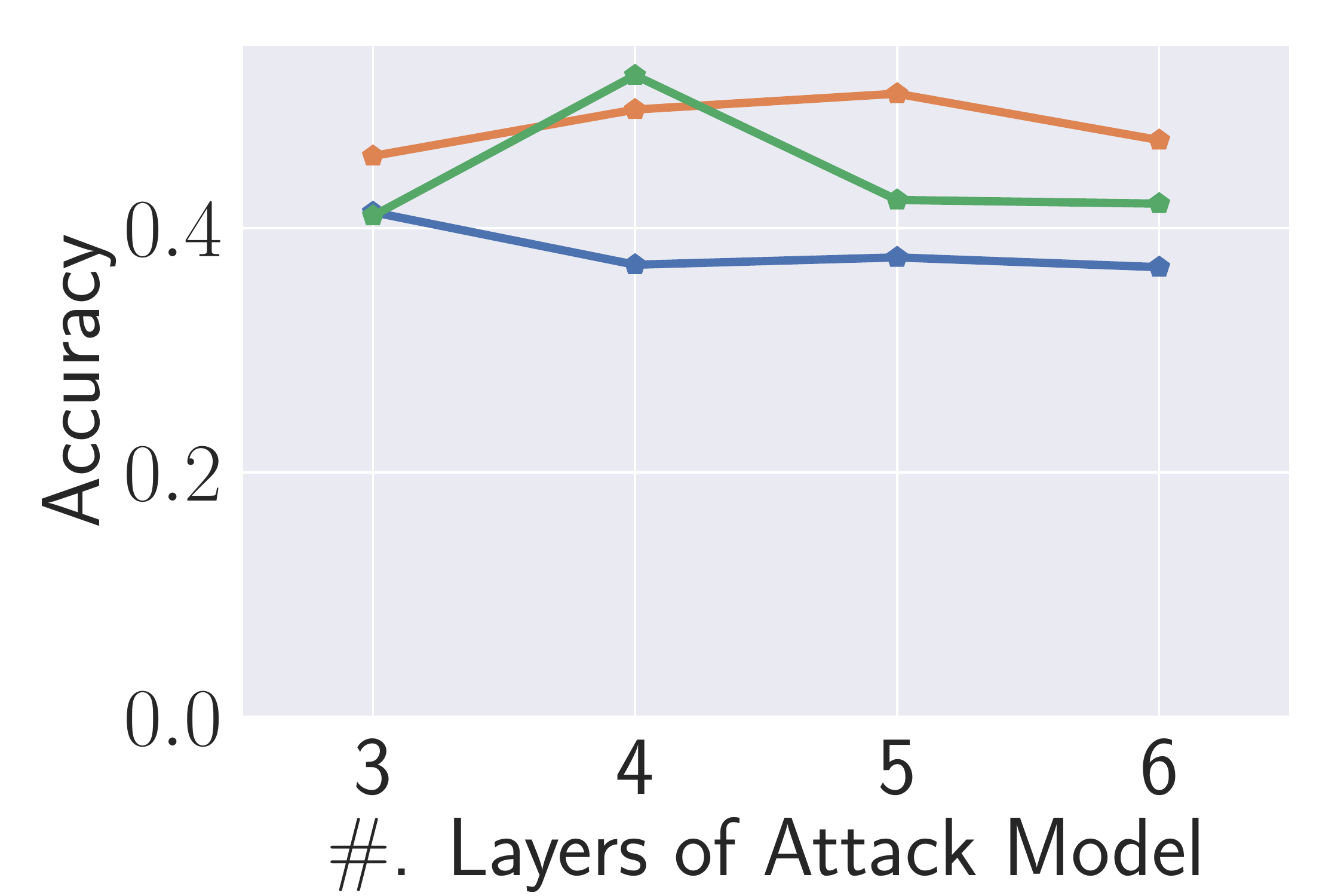}
\caption{Places20}
\label{figure:ai_diff_attack_layer_place20_SimCLR}
\end{subfigure}
\caption{The performance of attribute inference attacks against contrastive models on 4 different datasets under attack models with different layers.
The x-axis represents attack models' layers.
The y-axis represents attribute inference attacks' accuracy.}
\label{figure:ai_diff_attack_layer_SimCLR} 
\end{figure*}

\mypara{Datasets}
We utilize UTKFace, Places100, Places50, and Places20 to evaluate attribute inference attacks as they contain extra attributes that can be considered as sensitive attributes for our experiments (see \autoref{subsection:MIAExperimentalSetting}).
In UTKFace, the target model's classification task is gender classification, and the sensitive attribute is race (Black, White, Asian, Indian, and Other).
In Places100, Places50, and Places20, the target classification task is whether the scene is indoor or outdoor, and the sensitive attribmaxcute is scene categories.
Similar to Song and Shmatikov~\cite{SS20}, we take $\TargetTrain$ to generate the auxiliary dataset and train the attack model, and take $\TargetTest$ to test the attack performance.

\mypara{Metric}
We adopt accuracy as the metric to evaluate attribute inference attacks following previous work~\cite{MSCS19,SS20}.

\mypara{Models}
All the target models' architectures are the same as those for membership inference attacks.
For the attack model, we leverage a 3-layer MLP with the number of neurons in the hidden layer set to 128.
We use cross-entropy as the loss function and SGD as the optimizer with a learning rate of 0.01.
The attack model is trained for 100 epochs.
The dimension of each sample's representation from the base encoder, i.e., the attack model's input, is 1,280 for MobileNetV2, 512 for ResNet-18, and 2,048 for ResNet-50.

% ----------------------------------------------------
\subsection{Results}
\label{subsection:AIResults}
% ----------------------------------------------------

\begin{table}[!t]
\caption{The baseline accuracy (random guessing based on majority class labels) of attribute inference attack on different datasets.}
\label{table:ai_random_guess_baseline}
\centering
\begin{tabular}{l | c c }
\toprule
Dataset & \#. Class & Baseline Accuracy\\
\midrule
UTKFace & 5 & 0.421 \\
Places100 & 100 & 0.012 \\
Places50 & 50 & 0.023 \\
Places20 & 20 & 0.053 \\
\bottomrule
\end{tabular}
\end{table}

The performance of attribute inference attacks is depicted in \autoref{figure:ai_performance}.
First, we observe that, in general, attribute inference achieves effective performance except for the supervised model trained on UTKFace dataset (close to the prior sensitive attribute distribution in the attack training dataset as shown in \autoref{table:ai_random_guess_baseline}). 
Second, compared to the supervised models, the contrastive models are more vulnerable to attribute inference attacks.
For instance, on the UTKFace dataset with ResNet-18, we can achieve an attack accuracy of 0.701 on the contrastive model while only 0.422 on the supervised model.
To better understand this, we extract samples' representations (512-dimension) from ResNet-18 on UTKFace for both the supervised model and the contrastive model and project them into a 2-dimension space using t-Distributed Neighbor Embedding (t-SNE)~\cite{MH08}: \autoref{figure:tsne_1} shows the results for the supervised model on the original classification task, i.e., gender classification; \autoref{figure:tsne_2} shows the results for the supervised model on attribute inference, i.e., race.
We see that in \autoref{figure:tsne_1}, male samples (blue) and female samples (orange) reside in completely different regions, which can be separated perfectly (the gender classification accuracy is 0.875 in \autoref{figure:target_performance}).
However, for the sensitive attribute (\autoref{figure:tsne_2}), samples of different classes are clustered tightly, which increases the difficulty for attribute inference.
\autoref{figure:tsne_3} and \autoref{figure:tsne_4} show the corresponding results for the contrastive model.
We observe that different samples' representations on the contrastive model are less separable with respect to the original classification task compared to the supervised model (see \autoref{figure:tsne_3} and \autoref{figure:tsne_1}), but we can still successfully separate most of them correctly (the gender classification accuracy is 0.858 in \autoref{figure:target_performance}) since most of the male samples (blue) lie in the upper area while the female samples (orange) are in the lower area.
On the other hand, for the sensitive attribute, compared to the supervised model (\autoref{figure:tsne_2}), representations generated by the contrastive model (\autoref{figure:tsne_4}) are more distinguishable.
Our finding reveals that the representations generated by the contrastive model are more informative, which can be exploited not only for the original classification tasks but also for attribute inference.

To study the effect of training dataset size on the attack model $\AIAttackModel$, we randomly select from 10\% to 90\% of the training dataset to train the attack model and evaluate the performance using all the testing dataset; the results for contrastive models are summarized in \autoref{figure:ai_diff_ratio}.
By jointly considering \autoref{figure:ai_performance} and \autoref{figure:ai_diff_ratio}, we can observe that, in most of the cases, even using 10\% of the training dataset, the contrastive models are still more vulnerable to attribute inference attack than the supervised models when the attack model is trained with its \textit{full} training dataset.
On the other hand, the attack performance on supervised models is not significantly influenced by the training dataset size (see \autoref{figure:ai_diff_ratio_ce}).
This further shows the privacy risks of contrastive learning.

Recall our attack model is a 3-layer MLP.
We further investigate whether more complex attack models would improve the attack performance.
To this end, we increase the attack model's layer from 3 to 6 and summarize the corresponding attack performance for contrastive and supervised models in \autoref{figure:ai_diff_attack_layer_SimCLR} and \autoref{figure:ai_diff_attack_layer_CE} (in Appendix), respectively.
The results show that 3-layer attack models can achieve the best performance in most of the cases. 
With more layers, the attack performance may degrade or keep stable, which indicates that even simple models are enough to launch effective attacks.
This further shows that informative representations learned by contrastive models can be easily exploited by the adversary to infer samples' attributes.

We also observe that attribute inference attacks over contrastive models are more effective against smaller embedding size (see \autoref{figure:ai_performance} and \autoref{figure:ai_diff_ratio}).
For instance, ResNet-18 (512) leak more information than MobileNetV2 (1,280) and ResNet-50 (2,048).
We conjecture that a larger embedding size represents each sample in a more complex space in the contrastive setting, which is harder for the attack model to decode. 
However, the effect of embedding size on attribute inference attacks against the supervised models is less pronounced (see \autoref{figure:ai_performance} and \autoref{figure:ai_diff_ratio_ce} in the Appendix).
This further shows the difference between supervised models and contrastive models with respect to representing samples.

In conclusion, contrastive models are more vulnerable to attribute inference attacks compared to supervised models.

% ----------------------------------------------------
\section{Defense}
\label{section:Defense}
% ----------------------------------------------------

So far, we have demonstrated that compared to supervised models, contrastive models are more vulnerable to attribute inference attacks (\autoref{section:AI}) but less vulnerable to membership inference attacks (\autoref{section:MIA}).
In this section, we propose the first privacy-preserving contrastive learning mechanism, namely \Talos, which aims to reduce the risks of attribute inference for contrastive models while maintaining their membership privacy and model utility.

% ----------------------------------------------------
\subsection{Methodology}
\label{subsection:Talosology}
% ----------------------------------------------------

\mypara{Intuition}
As shown in \autoref{section:AI}, the reason for a contrastive model to be vulnerable to attribute inference attacks is that the model's base encoder $\Encoder$ learns informative representations for data samples, which can be exploited by an adversary.
To mitigate such a threat, we aim for a new training paradigm for contrastive learning which can eliminate data samples' sensitive attributes from their representations.
Meanwhile, the base encoder of the contrastive model still needs to represent data samples expressively for preserving model utility.
These two objectives are in conflict, and our defense mechanism should consider both simultaneously.

\mypara{Methodology}
Our defense mechanism, namely \Talos, can be modeled as a mini-max game, and we rely on adversarial training ~\cite{GPMXWOCB14,ES16,XDDHN17,EG18,CNC18} to realize it.
Similar to the original contrastive model, \Talos also leverages a base encoder and a projection head to learn informative representations for data samples.
Besides, \Talos introduces an adversarial classifier $\AdvClassifier$, which is used to censor sensitive attributes from data samples' representations.

The adversarial classifier of \Talos is essentially designed for attribute inference.
Similar to the original contrastive learning process (see \autoref{section:Preliminary}), \Talos is trained with mini-batches.
Given a mini-batch of $2N$ augmented data samples (generated from $N$ original samples), we define the loss of the adversarial classifier $\AdvClassifier$ as follows.
\begin{equation}
\label{equation:adversarial_loss}
\LossFunction_{\AdvClassifier} = \frac{1}{2N} \sum_{k=1}^{N} [\CELoss(\SensitiveAttribute_{k}, \AdvClassifier(\Encoder(\tilde{\DataPoint}_{2k-1}))) + \CELoss(\SensitiveAttribute_{k}, \AdvClassifier(\Encoder(\tilde{\DataPoint}_{2k})))]
\end{equation}
where $\tilde{\DataPoint}_{2k-1}$ and $\tilde{\DataPoint}_{2k}$ are the two augmented samples of an original sample $\DataPoint_{k}$, $\SensitiveAttribute_{k}$ represents $\DataPoint_{k}$'s sensitive attribute, $\Encoder$ is the base encoder, and $\CELoss$ is the cross-entropy loss (\autoref{equation:CELoss}).
We consider $\tilde{\DataPoint}_{2k-1}$ and $\tilde{\DataPoint}_{2k}$ sharing the same sensitive attribute as $\DataPoint_{k}$.
Note that we take the output of the base encoder instead of the projection head as the input to the adversarial classifier.
Since the projection head is discarded after the first phase of training the contrastive model, directly optimizing the base encoder with the adversarial classifier loss would maintain the effect of adversarial training.

\begin{algorithm}[!t]
\caption{The training process of \Talos.}
\label{algorithm:defense}
\SetAlgoLined
\LinesNumbered
\textbf{Input}: Target training dataset $\TargetTrain$ with sensitive attribute $\SensitiveAttribute$, base encoder $\Encoder$, projection head $\ProjectionHead$, adversarial classifier $\AdvClassifier$, and adversarial factor $\lambda$.\\
Initialize $\Encoder$, $\ProjectionHead$, and $\AdvClassifier$'s parameters.\\
\For{\textbf{each epoch}}{
\For{\textbf{each mini-batch}}{
Sample a mini-batch with $N$ training data samples and its corresponding sensitive attributes $\{(x_1,s_1), (x_2,s_2), ..., (x_N,s_N)\}$ from $\TargetTrain$\\
Generate augmented data samples:
$\{(\tilde{\DataPoint}_1,s_1), (\tilde{\DataPoint}_2,s_1), ..., (\tilde{\DataPoint}_{2N},s_N)\}$, where $\tilde{\DataPoint}_{2k-1}$ and $\tilde{\DataPoint}_{2k}$ are the two augmented views of $\DataPoint_k$\\
Feed augmented data samples into the base encoder $\Encoder$ and the projection head $\ProjectionHead$ to calculate the contrastive loss: 
$\FinalSimCLRLoss=\frac{1}{2 N} \sum_{k=1}^{N}[\ell(2 k-1,2 k)+\ell(2 k, 2 k-1)]$\\
Feed the representations generated by the base encoder $\Encoder$ into the adversarial classifier $\AdvClassifier$ to calculate the adversarial classifier loss: 
$\LossFunction_{\AdvClassifier} = \frac{1}{2N} \sum_{k=1}^{N} [\CELoss(\SensitiveAttribute_{k}, \AdvClassifier(\Encoder(\tilde{\DataPoint}_{2k-1}))) + \CELoss(\SensitiveAttribute_{k}, \AdvClassifier(\Encoder(\tilde{\DataPoint}_{2k})))]$\\
\uIf{$\textbf{epoch} \mod 2 \neq 0$}{
Optimize adversarial classifier $\AdvClassifier$'s parameters with the adversarial classifier loss: $\mathcal{L}_{\AdvClassifier}$
}
\Else{
Optimize projection head $\ProjectionHead$'s parameters with the contrastive loss:  $\FinalSimCLRLoss$\\
Optimize base encoder $\Encoder$'s parameters with adversarial training loss: 
$\PrivacyPreservingLoss=\FinalSimCLRLoss - \lambda \mathcal{L}_{\AdvClassifier}$
}
}
}
\textbf{Return}: Base encoder $\Encoder$
\end{algorithm}

\Talos also adopts the original contrastive loss $\FinalSimCLRLoss$ (\autoref{equation:contrastive_loss}).
By jointly considering the adversarial classifier loss and the contrastive loss, \Talos's loss function is defined as follows:
\begin{equation}
\label{equation:defense_loss}
\PrivacyPreservingLoss = \FinalSimCLRLoss - \lambda \mathcal{L}_{\AdvClassifier}
\end{equation}
where $\lambda$ is the \textit{adversarial factor} to balance the two losses.
We refer to a model trained with \Talos as a \Talos model.

\autoref{algorithm:defense} presents the training process of \Talos.
In each mini-batch, given $N$ training samples, we first generate $2N$ augmented views (Line 6) and feed them into the base encoder. 
The generated representations are then fed into the projection head (Line 7) and the adversarial classifier (Line 8) simultaneously.
Note that the adversarial classifier and contrastive model are updated alternately by epoch.
We First optimize the adversarial classifier with the cross-entropy loss (Line 10).
Then we optimize the projection head with the contrastive loss (Line 12) and the base encoder with the loss function of \Talos, i.e., \autoref{equation:defense_loss} (Line 13).

To implement this in practice, we utilize the gradient reversal layer (GRL) proposed by Ganin et al.~\cite{GL15}.
GRL is a layer that can be added between the base encoder $\Encoder$ and the adversarial classifier $\AdvClassifier$.
In the forward propagation, GRL acts as an identity transform that simply copies the input as the output.
During the backpropagation, GRL takes the gradients passed through it from the adversarial classifier $\AdvClassifier$, multiplies the gradients by $-\lambda$, and passes them to the base encoder $\Encoder$.
Such operation lets the base encoder receive the opposite direction of gradients from the adversarial classifier.
In this way, the base encoder $\Encoder$ is able to learn informative representations for samples while censoring their sensitive attributes.

\begin{figure*}[!t]
\centering
\begin{subfigure}{0.5\columnwidth}
\includegraphics[width=\columnwidth]{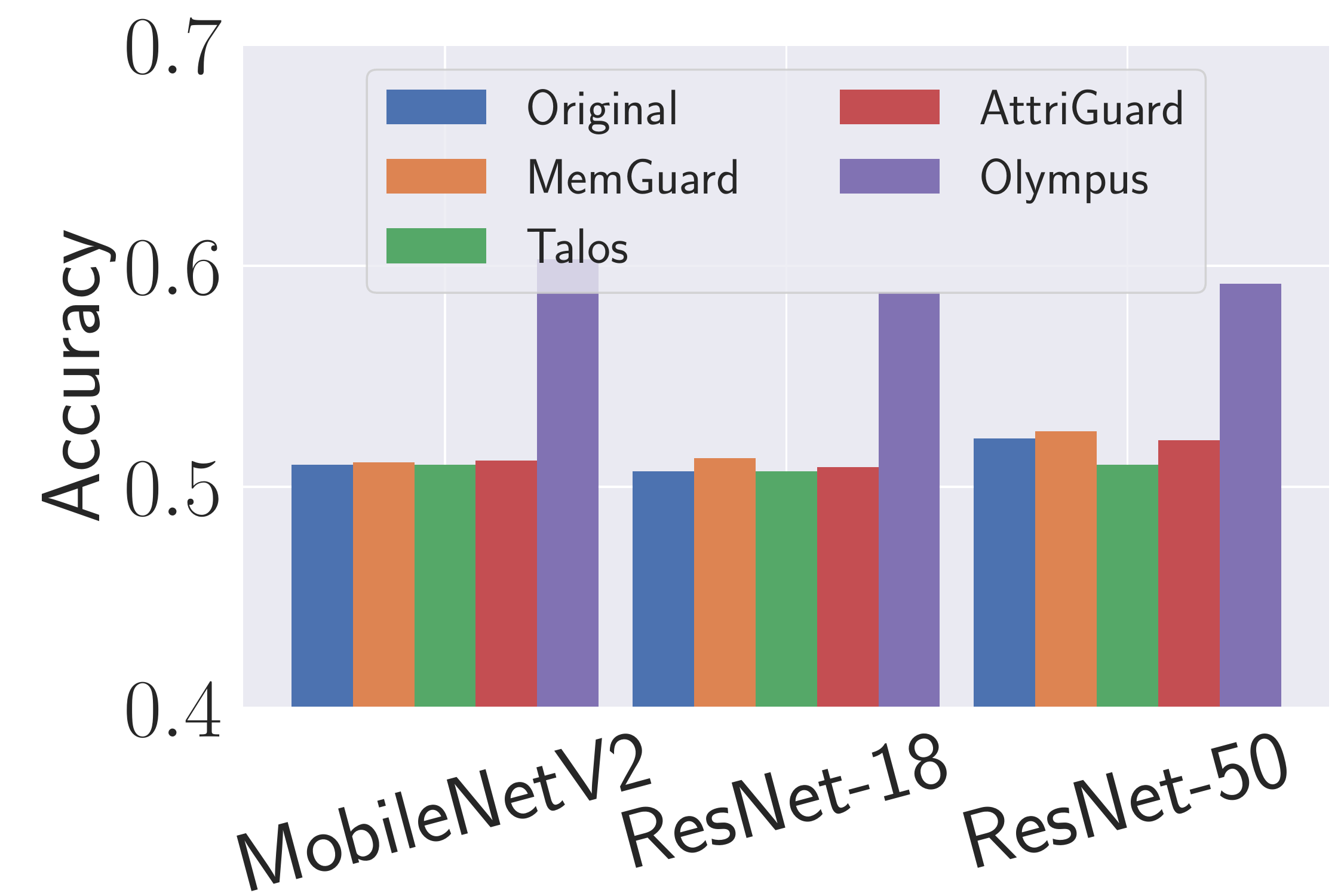}
\caption{UTKFace}
\label{figure:defense_comparison_mia_UTKFace}
\end{subfigure}
\begin{subfigure}{0.5\columnwidth}
\includegraphics[width=\columnwidth]{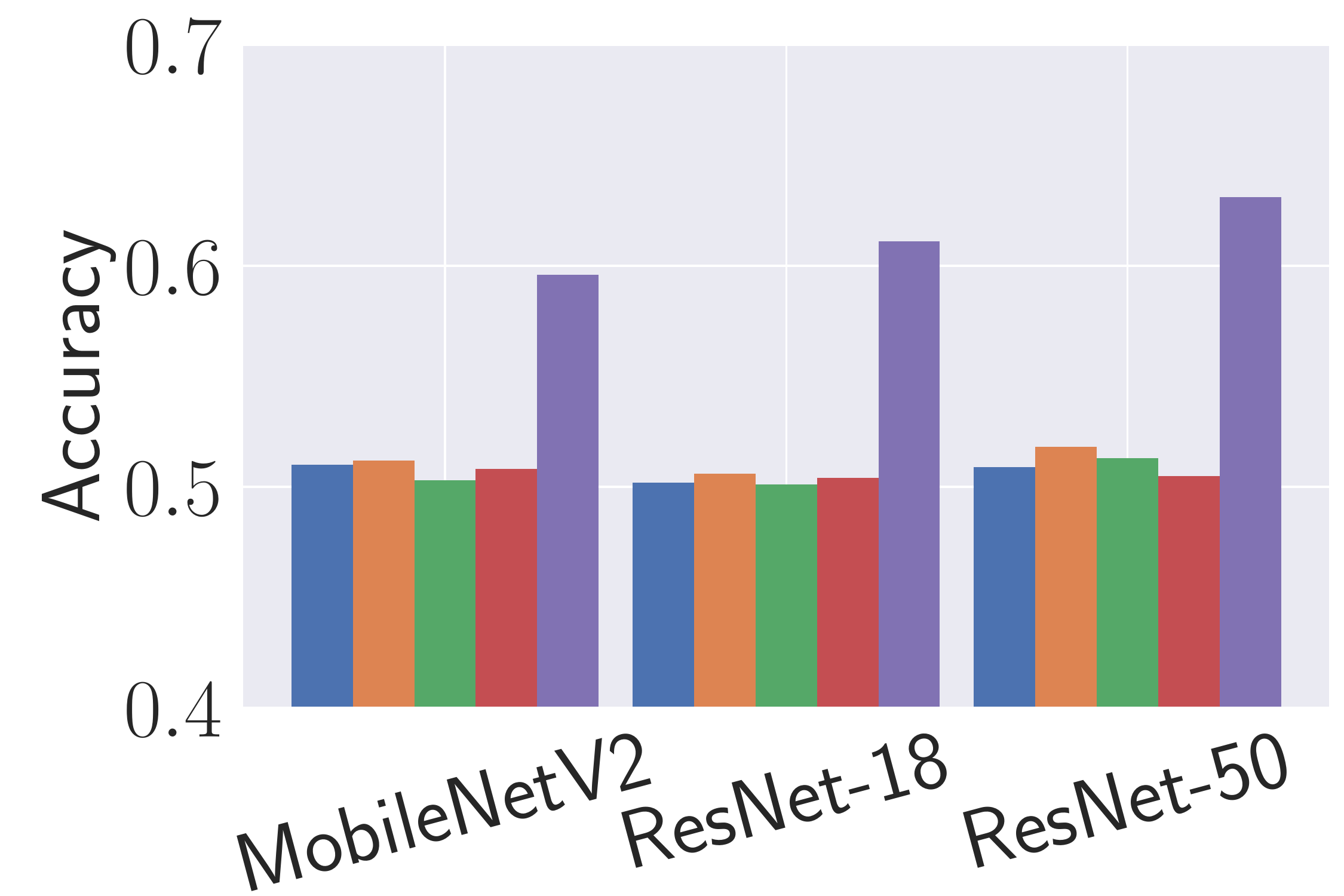}
\caption{Places100}
\label{figure:defense_comparison_mia_Places100}
\end{subfigure}
\begin{subfigure}{0.5\columnwidth}
\includegraphics[width=\columnwidth]{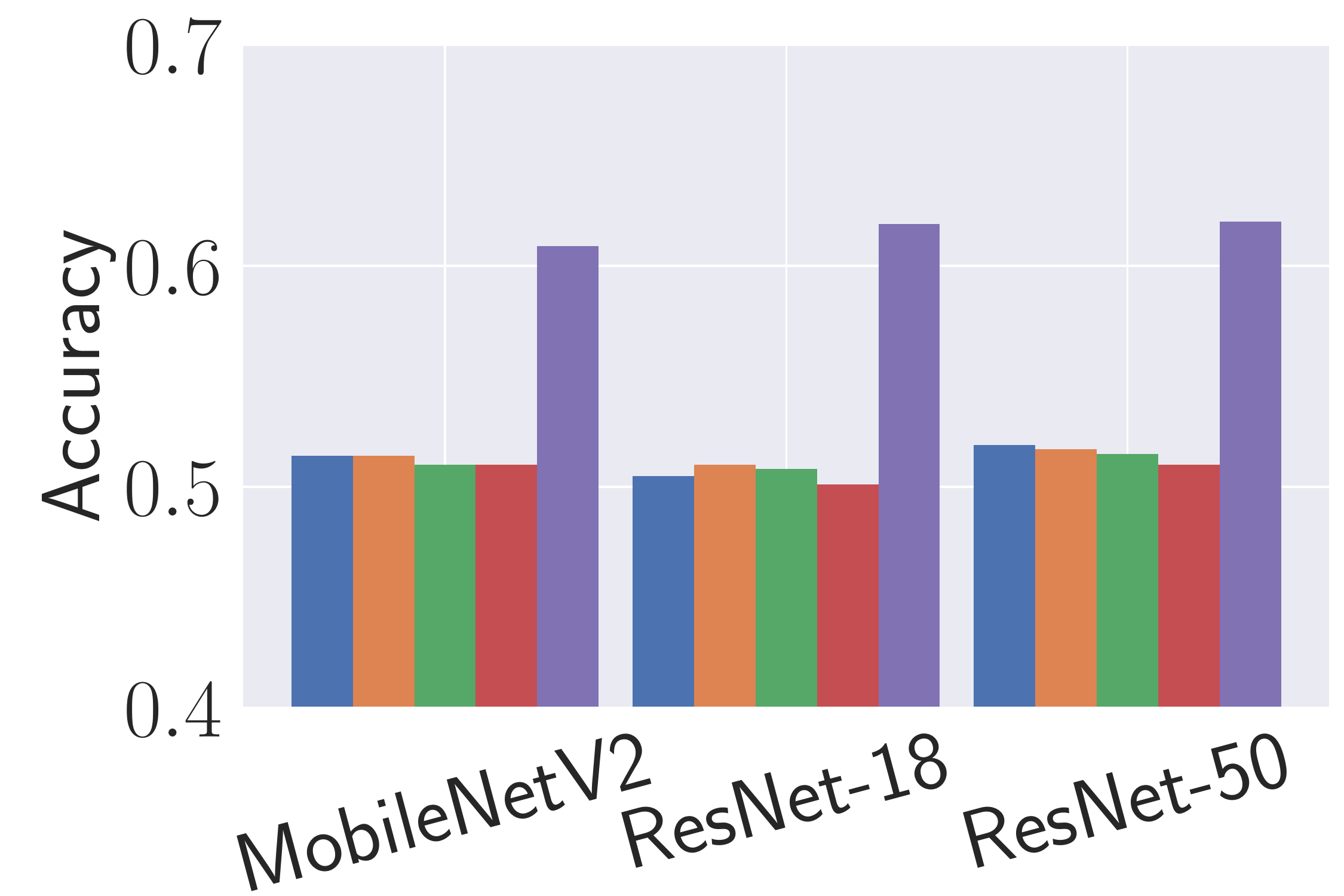}
\caption{Places50}
\label{figure:defense_comparison_mia_Places50}
\end{subfigure}
\begin{subfigure}{0.5\columnwidth}
\includegraphics[width=\columnwidth]{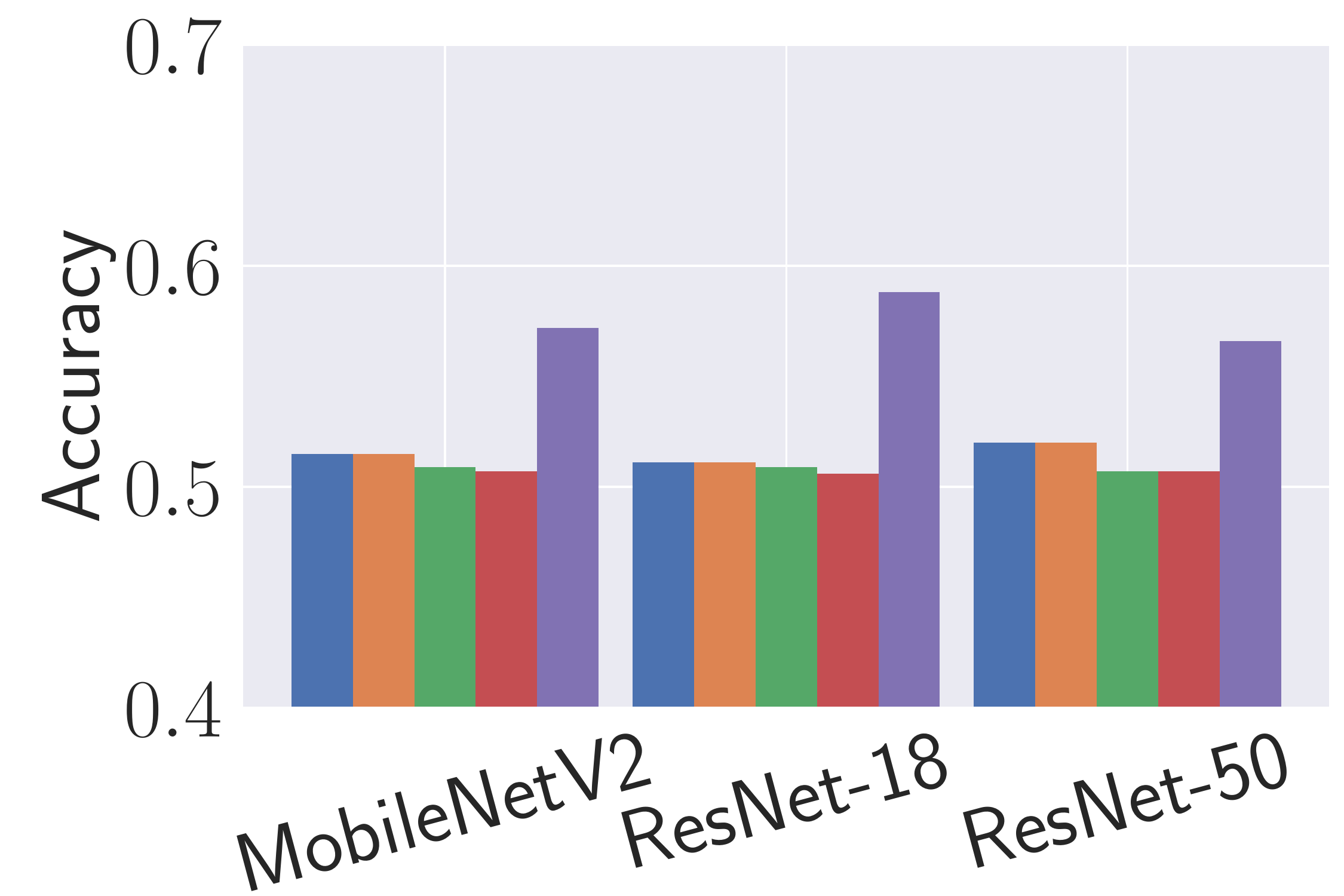}
\caption{Places20}
\label{figure:defense_comparison_mia_Places20}
\end{subfigure}
\caption{
The performance of NN-based membership inference attacks against original contrastive models, \Talos, \MemGuard, \Olympus, and \AttriGuard with MobileNetV2, ResNet-18, and ResNet-50 on 4 different datasets.
The x-axis represents different models.
The y-axis represents the accuracy of NN-based membership inference attacks.}
\label{figure:defense_comparison_mia_NN-based}
\end{figure*}

Note that our adversarial training is performed only on the process of training the base encoder $\Encoder$.
The training for the classification layer of the contrastive model remains unchanged.
As we show in \autoref{section:MIA}, the classification layer generalizes well on the contrastive models, i.e., less overfitting. 
Therefore, models trained by \Talos should be robust against membership inference attacks as well.
Our evaluation shows that this is indeed the case (see \autoref{figure:defense_comparison_mia_NN-based}).

\mypara{Adaptive Attacks}
An adversary needs to establish a shadow model to mount membership inference attacks.
To evaluate membership privacy risks of \Talos, we consider an adaptive (and stronger) adversary~\cite{JSBZG19}.
Concretely, we assume that the adversary knows the training details of \Talos and trains their shadow model in the same way.
For attribute inference, the attack model is trained on embeddings generated by \Talos, thus, our attribute inference attack considered in the evaluation of \Talos is also adaptive.

% ----------------------------------------------------
\subsection{Experimental Setting}
\label{subsection:DefenseSetting}
% ----------------------------------------------------

We follow the same experimental setting, including datasets, metrics, target models, and attack models (both attribute inference and membership inference), as those in \autoref{subsection:MIAExperimentalSetting} and \autoref{subsection:AIExperimentalSetting}. 
As mentioned before, both membership inference and attribute inference attacks are performed in an adaptive way.
Regarding the adversarial classifier of \Talos, we leverage a 3-layer MLP with 64 neurons in the hidden layer, which is smaller than the attribute inference attack model.

\begin{figure*}[!t]
\centering
\begin{subfigure}{0.5\columnwidth}
\includegraphics[width=\columnwidth]{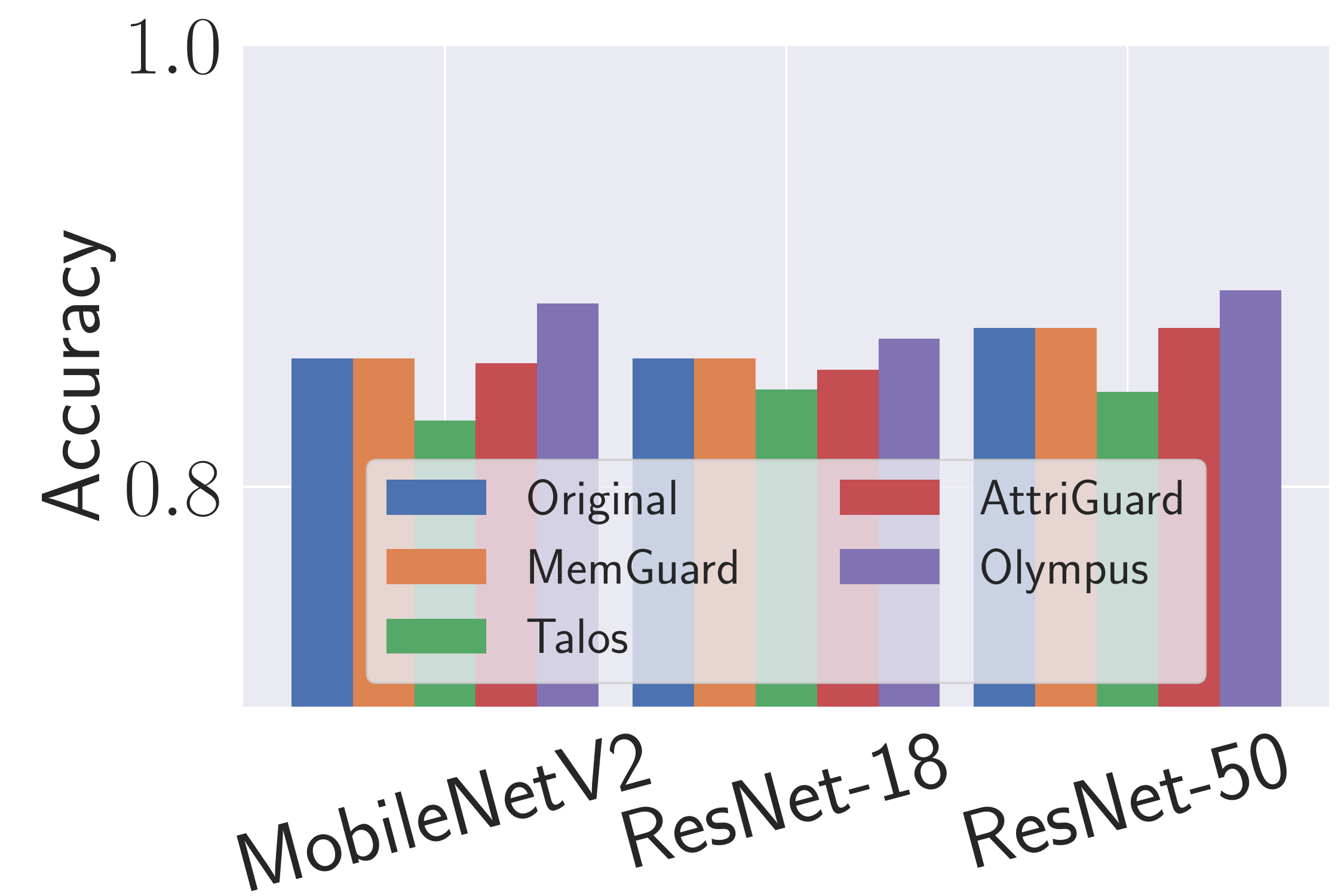}
\caption{UTKFace}
\label{figure:defense_comparison_target_UTKFace}
\end{subfigure}
\begin{subfigure}{0.5\columnwidth}
\includegraphics[width=\columnwidth]{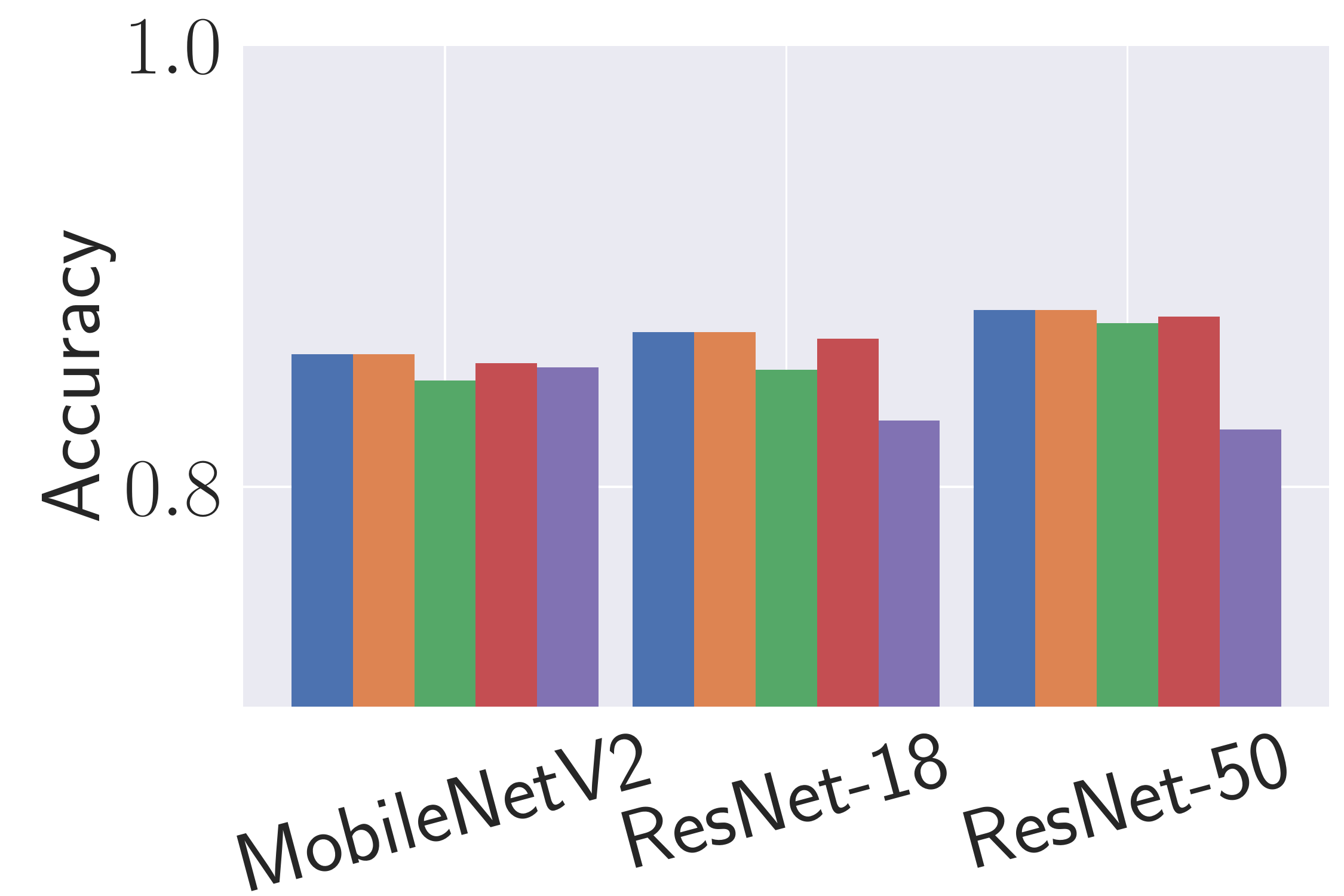}
\caption{Places100}
\label{figure:defense_comparison_target_Places100}
\end{subfigure}
\begin{subfigure}{0.5\columnwidth}
\includegraphics[width=\columnwidth]{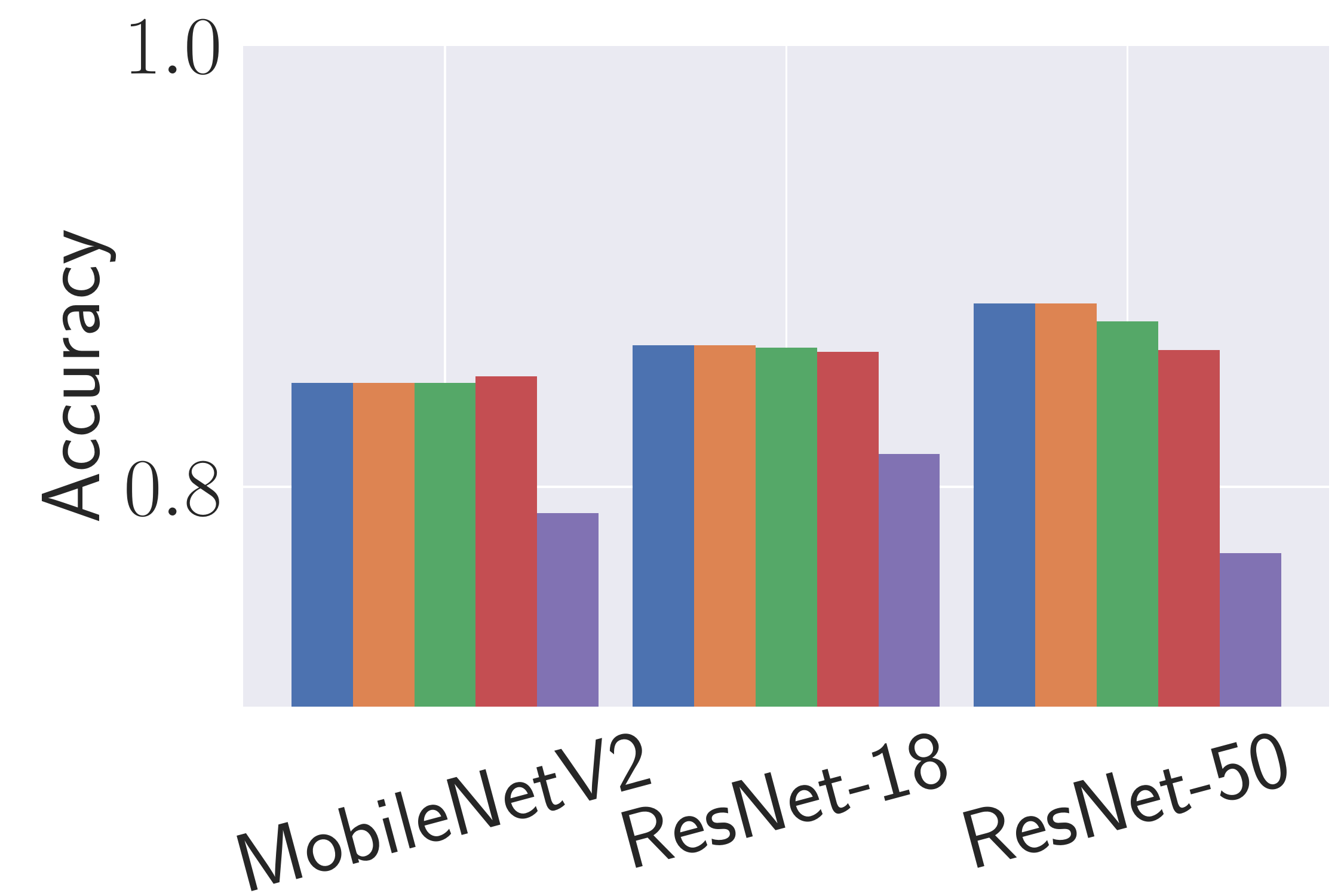}
\caption{Places50}
\label{figure:defense_comparison_target_Places50}
\end{subfigure}
\begin{subfigure}{0.5\columnwidth}
\includegraphics[width=\columnwidth]{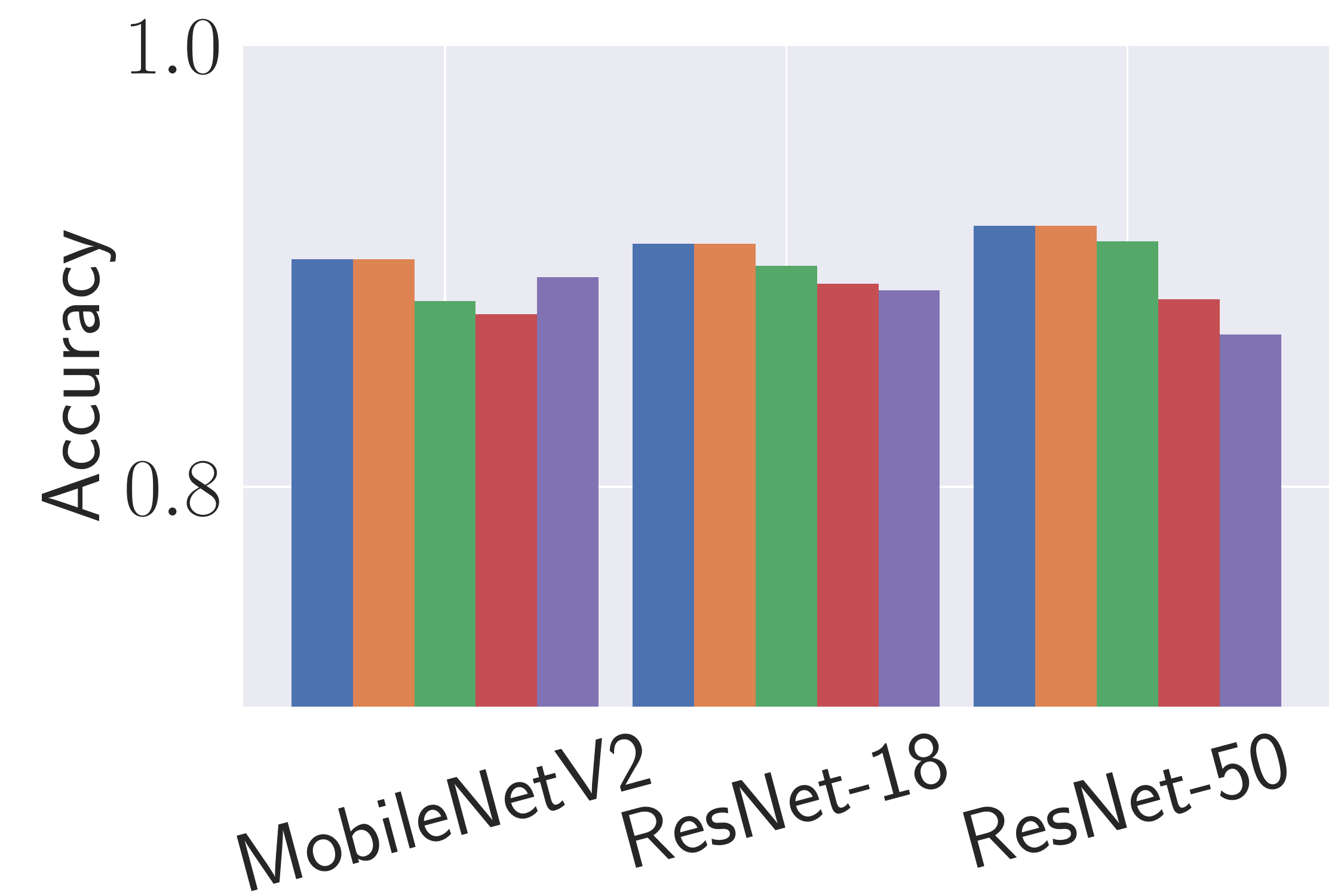}
\caption{Places20}
\label{figure:defense_comparison_target_Places20}
\end{subfigure}
\caption{
The performance of original classification tasks against original contrastive models, \Talos, \MemGuard, \Olympus, and \AttriGuard with MobileNetV2, ResNet-18, and ResNet-50 on 4 different datasets.
The x-axis represents different models.
The y-axis represents the accuracy of original classification tasks.}
\label{figure:defense_comparison_target}
\end{figure*}

\begin{figure*}[!t]
\centering
\begin{subfigure}{0.5\columnwidth}
\includegraphics[width=\columnwidth]{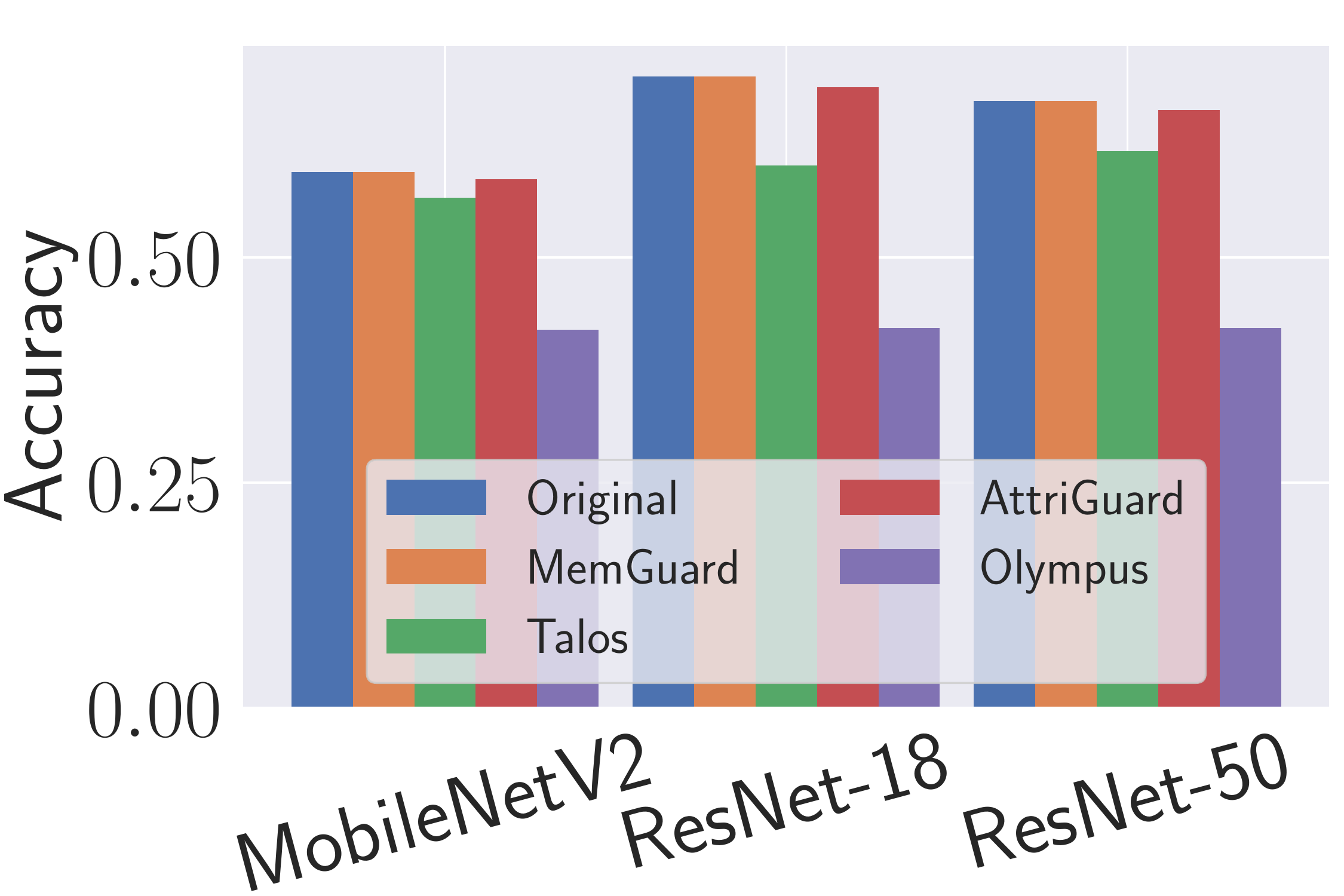}
\caption{UTKFace}
\label{figure:defense_comparison_ai_UTKFace}
\end{subfigure}
\begin{subfigure}{0.5\columnwidth}
\includegraphics[width=\columnwidth]{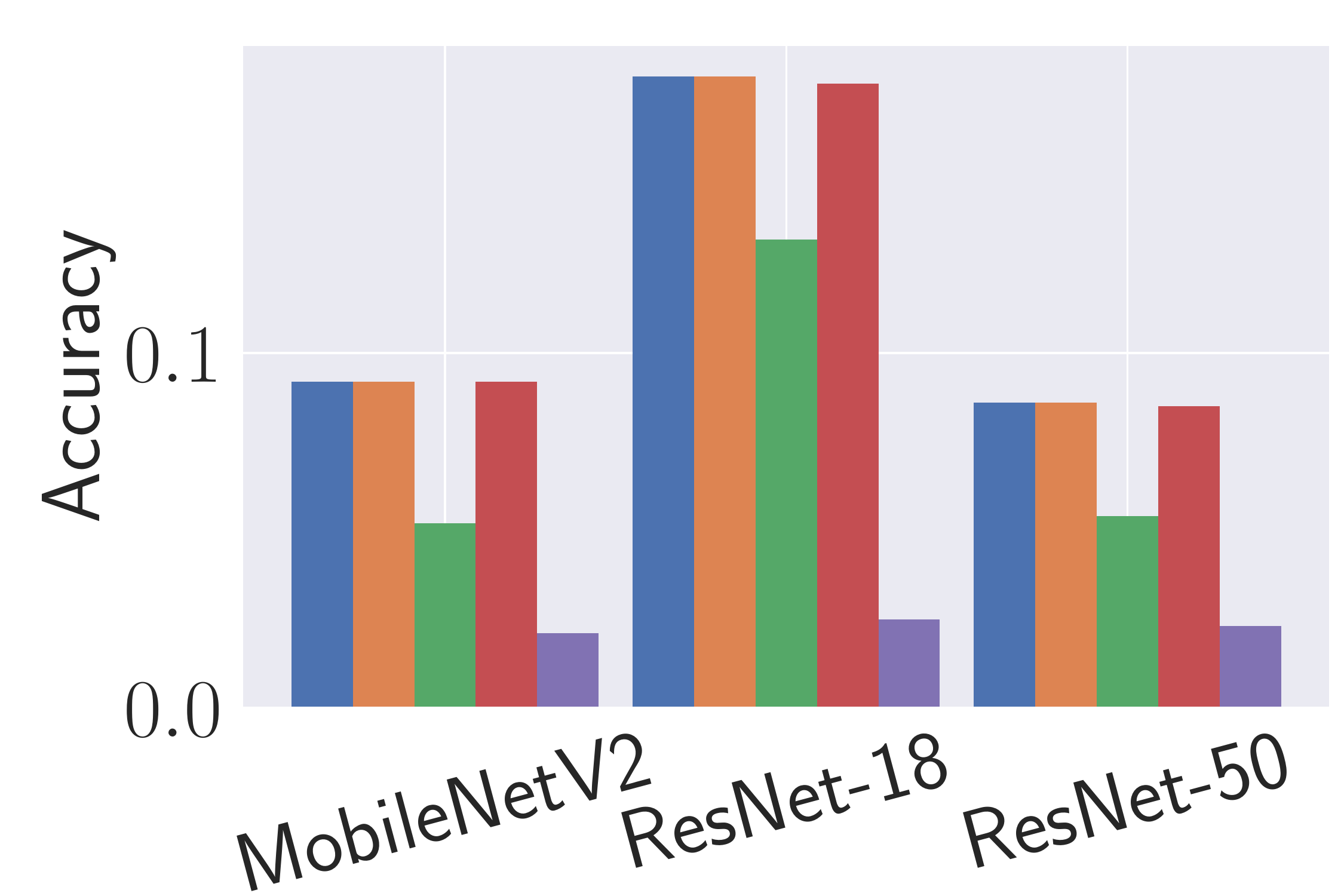}
\caption{Places100}
\label{figure:defense_comparison_ai_Places100}
\end{subfigure}
\begin{subfigure}{0.5\columnwidth}
\includegraphics[width=\columnwidth]{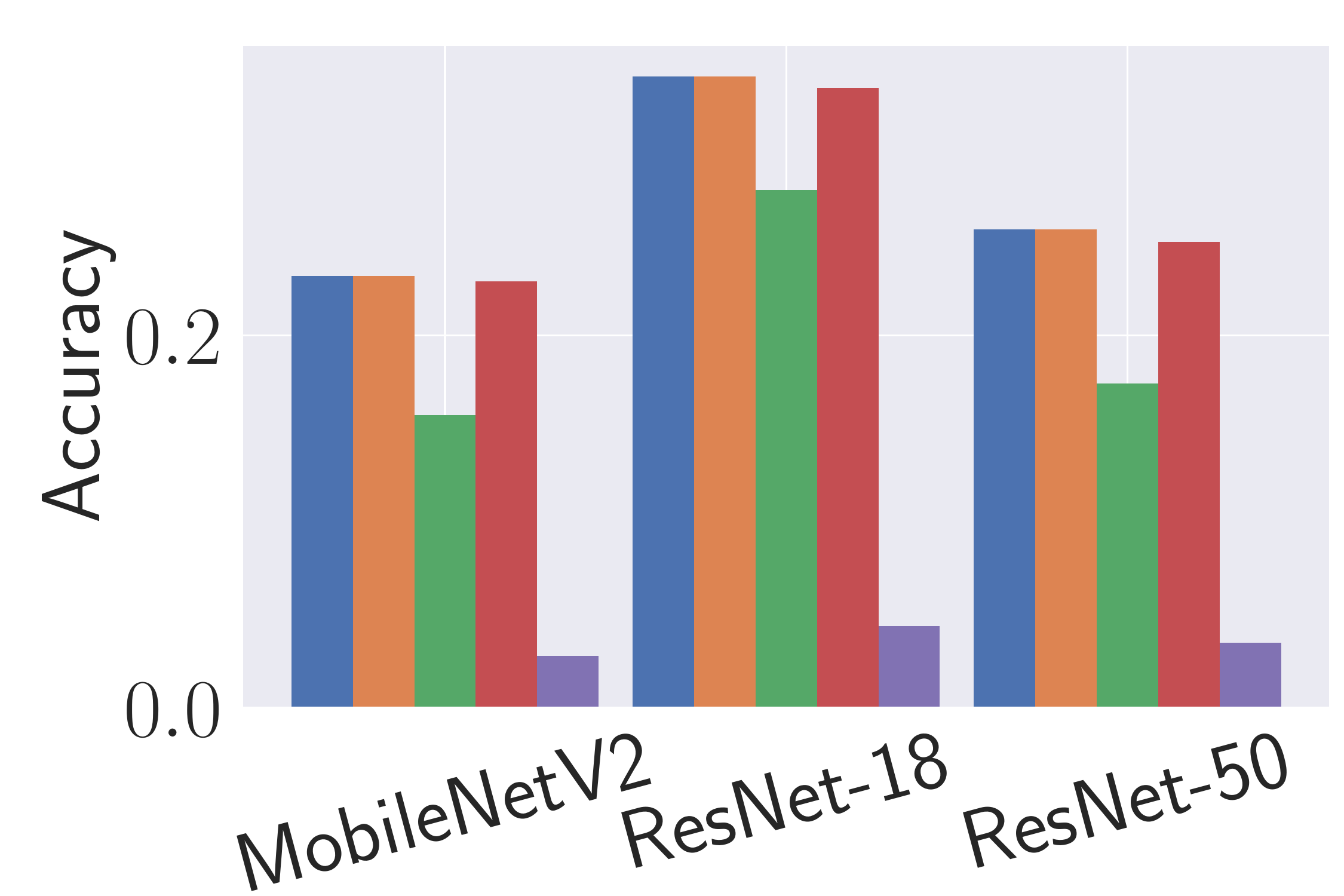}
\caption{Places50}
\label{figure:defense_comparison_ai_Places50}
\end{subfigure}
\begin{subfigure}{0.5\columnwidth}
\includegraphics[width=\columnwidth]{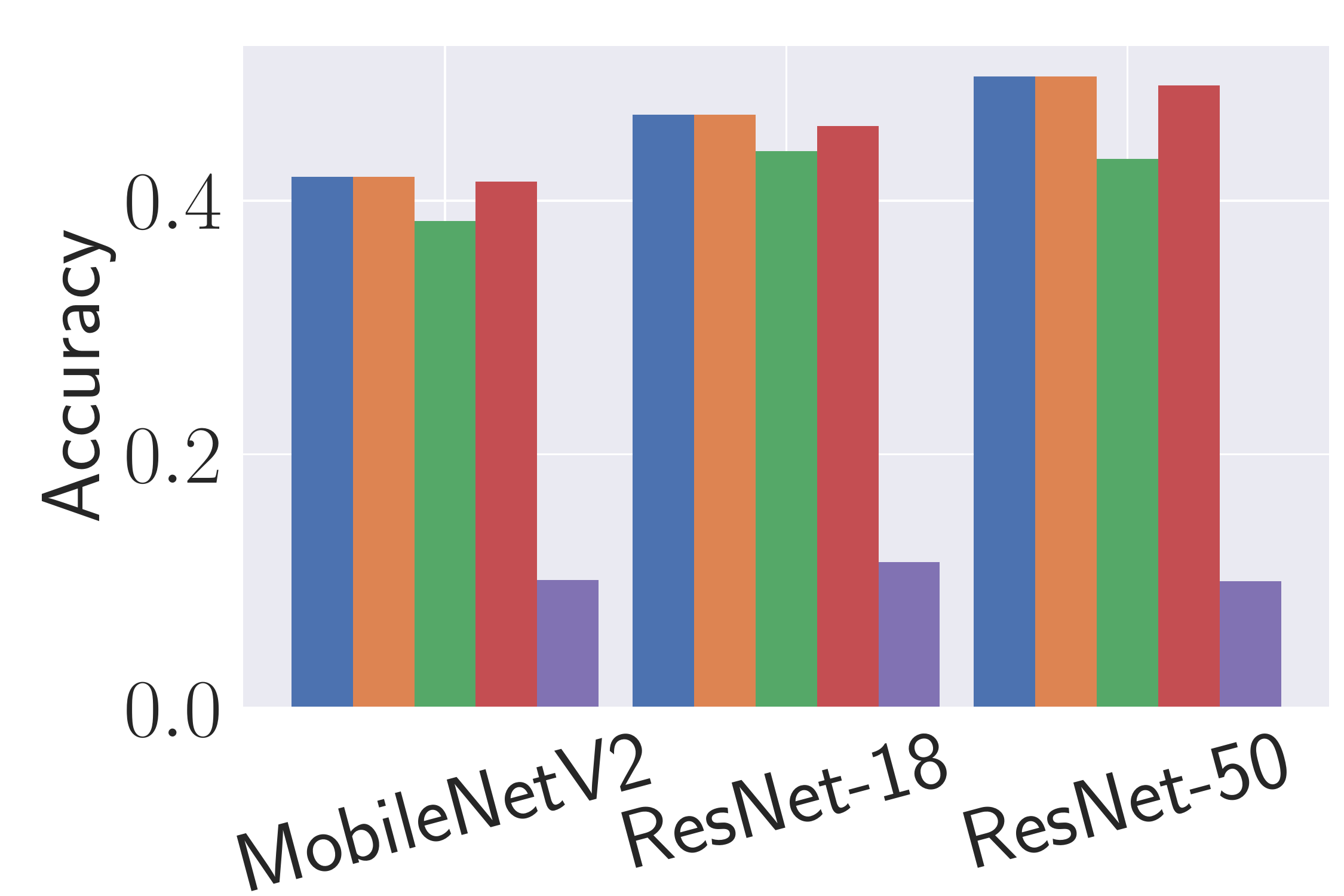}
\caption{Places20}
\label{figure:defense_comparison_ai_Places20}
\end{subfigure}
\caption{
The performance of attribute inference attacks against original contrastive models, \Talos, \MemGuard, \Olympus, and \AttriGuard with MobileNetV2, ResNet-18, and ResNet-50 on 4 different datasets.
The x-axis represents different models.
The y-axis represents the accuracy of attribute inference attacks.}
\label{figure:defense_comparison_ai}
\end{figure*}

\mypara{Baseline}
We consider three state-of-the-art defenses, one for membership inference (\MemGuard~\cite{JSBZG19}) and two for attribute inference (\Olympus~\cite{RMP19} and \AttriGuard~\cite{JG18}) as the baseline models.
\MemGuard, \Olympus, and \AttriGuard are originally designed for supervised models, here, we adapt them to contrastive models.
Since the input to the attribute inference attack is each sample's representation, we further consider a sample's representation as the input to \Olympus and \AttriGuard.

\MemGuard is a two-phase defense for membership inference.
In phase I, the defender generates a noise vector to perturb the posteriors of a target sample, so that the adversary's membership classifier is likely to give a random guess for the perturbed posteriors.
In phase II, the defender adds the noise vector to the posteriors with certain probability. 

\Olympus, designed for attribute inference, has three basic components: an autoencoder to transfer the original representation into the perturbed one, a classifier to perform the original task over the perturbed representation, and an adversarial classifier to infer the sensitive attribute from the perturbed representation.
\Olympus optimizes the three components using adversarial training to preserve the model utility while protecting samples' sensitive attributes.
To perform \Olympus on contrastive models, we first train a base encoder following the original contrastive learning process.
Then, we add an autoencoder between the base encoder and the classification layer, and fine-tune the whole model using the original training samples with \Olympus's losses.

\AttriGuard is a two-phase defense for attribute inference.
In phase I, for each representation, the defender generates an adversarial example for each possible value of the sensitive attribute by adapting the existing evasion attack techniques.
In phase II, the defender samples one sensitive attribute value based on a probability distribution and selects the corresponding adversarial example found in phase I as the new representation.

The adversarial classifier used in \AttriGuard and \Olympus shares the same architecutre as the one in \Talos.
For \MemGuard, we follow Jia et al.~\cite{JSBZG19} to generate the noise in Phase I.
For the autoencoder of \Olympus, we set its encoder (decoder) as a 2-layer MLP with 256 and 128 (128 and 256) neurons in the hidden layers. 
For \AttriGuard, we leverage the C\&W attack~\cite{CW17} with the $L_{inf}$ norm in phase I.

% ----------------------------------------------------
\subsection{Results}
\label{subsection:DefenseResult}
% ----------------------------------------------------

We compare the performance of the original classification tasks, NN-based membership inference attacks, and attribute inference attacks for the original contrastive model and the models defended by \Talos, \MemGuard, \Olympus, and \AttriGuard.
The results are depicted in \autoref{figure:defense_comparison_target}, \autoref{figure:defense_comparison_mia_NN-based}, and \autoref{figure:defense_comparison_ai}, respectively.
Note that we also perform metric-based and label-only membership inference attacks and the results are summarized in \autoref{figure:defense_comparison_mia_Metric-corr}, \autoref{figure:defense_comparison_mia_Metric-conf}, \autoref{figure:defense_comparison_mia_Metric-ent}, \autoref{figure:defense_comparison_mia_Metric-ment}, and \autoref{figure:defense_comparison_mia_Label-only} in Appendix.

In \autoref{figure:defense_comparison_ai}, we find that \Talos indeed reduces the attribute inference accuracy compared to the original contrastive learning.
For instance, the attribute inference accuracy is 0.701 on the original contrastive model with ResNet-18 on the UTKFace dataset, while only 0.602 on the \Talos model.
Meanwhile, the testing accuracy of the original classification task for the \Talos model is also preserved (\autoref{figure:defense_comparison_target}).

For different defense mechanisms, we find that \Olympus reduces attribute inference attacks the most (see \autoref{figure:defense_comparison_ai}).
However, it jeopardizes the membership privacy to a large extent (see \autoref{figure:defense_comparison_mia_NN-based}).
For instance, the membership inference accuracy of the \Talos model (ResNet-50) on Place100 is 0.513 while the corresponding \Olympus model's value is 0.631.
The reason is that \Olympus's training process utilizes the original training samples to fine-tune the whole model, which leads to the model memorizing these samples with the model's full capacity.
On the other hand, as mentioned in \autoref{subsection:Talosology}, \Talos is only performed on the training process of the base encoder $\Encoder$ which considers each sample's augmented views.
The original samples are only used to fine-tune the final classification layer, the same as training a normal contrastive model.
In other words, the \Talos model memorizes its training samples with only its one-layer capacity.
Therefore, \Talos models are less prone to membership inference.
In addition, \Olympus jeopardizes the target model's utility in multiple cases (see \autoref{figure:defense_comparison_target_Places100}, \autoref{figure:defense_comparison_target_Places50}, and \autoref{figure:defense_comparison_target_Places20}), the reason again lies in the training process of \Olympus.
More specifically, \Olympus needs to fine-tune the whole model in a supervised way, this reduces the effect of contrastive learning in the final model.
Meanwhile, \Talos preserves the contrastive learning process to a large extent as its adversarial loss is applied together with the contrastive loss during the training of the base encoder.
Since membership privacy, attribute privacy, and model utility are equally important, we believe \Talos is a better choice than \Olympus.

We also find that \Talos, \MemGuard, and \AttriGuard models can achieve similar utility as the original contrastive models (see~\autoref{figure:defense_comparison_target}).
However, \Talos can mitigate attribute inference attacks to a larger extent than \AttriGuard and \MemGuard (see \autoref{figure:defense_comparison_ai}).
For instance, the attribute inference accuracy is only 0.132 on the \Talos model with ResNet-18 on the Places100 dataset, while 0.176 and 0.178 on the \AttriGuard and \MemGuard models.
Also, as the contrastive learning procedure is preserved for \Talos, \AttriGuard, and \MemGuard, we observe that all these defenses are robust against membership inference attacks (see \autoref{figure:defense_comparison_mia_NN-based}).

We also investigate the effect of the adversarial factor $\lambda$ on the performance of original classification tasks, membership inference attacks, and attribute inference attacks.
The results are summarized in \autoref{figure:adv_training_target_performance}, \autoref{figure:adv_training_mia_performance}, and \autoref{figure:adv_training_ai_performance}.
First of all, we observe that the performance of original classification tasks (\autoref{figure:adv_training_target_performance}) and membership inference attacks ( \autoref{figure:adv_training_mia_performance}) are relatively stable with respect to different adversarial factors.
However, for different datasets or different model architectures, the best $\lambda$ to defense attributes inference attack may vary (\autoref{figure:adv_training_ai_performance}).
In general, we notice that setting $\lambda$ to 2 or 3 can achieve nearly the best defense performance on most datasets and model architectures. 
To perform \Talos in practice, we believe the model owner needs to tune the $\lambda$ on their validation dataset. 
During the process, concentrating more on model utility or defense effectiveness depends on the ML model owner’s purpose.

In conclusion, \Talos can successfully defend attribute inference attacks for contrastive models without jeopardizing their membership privacy and model utility.

% ----------------------------------------------------
\section{Related Work}
\label{section:RelatedWork}
% ----------------------------------------------------

\mypara{Contrastive Learning}
Contrastive learning is one of the most popular self-supervised learning paradigms~\cite{HFWXG20,GH10,OLV18,CKNH20,YCSCWS20,JXZZZZ20}.
Oord et al.~\cite{OLV18} propose contrastive predictive coding, which leverages autoregressive models to predict future observations for data samples.
Wu et al.~\cite{WXYL18} utilize a memory bank to save instance representation and k-nearest neighbors to conduct prediction.
He et al.~\cite{HFWXG20} introduce MoCo, which relies on momentum to update the key encoder with the query encoder to maintain consistency. 
Chen et al.~\cite{CKNH20} propose SimCLR, which leverages data augmentation and the projection head to enhance the performance of contrastive models.
SimCLR is the most prominent contrastive learning paradigm at the moment~\cite{LZHWMZT20}, thus we concentrate on it in this paper.

\mypara{Membership Inference Attack}
In membership inference, the adversary's goal is to infer whether a given data sample is used to train a target model.
Right now, membership inference is one of the major means to measure privacy risks of machine learning models~\cite{SSSS17,YGFJ18,HMDC19,SZHBFB19,NSH19,SSM19,LF20,HWWBSZ21}.
Shokri et al.~\cite{SSSS17} propose the first membership inference attack in the black-box setting.
Specifically, they rely on training multiple shadow models to mimic the behavior of a target model to derive the data for training their attack models.
Salem et al.~\cite{SZHBFB19} further relax the assumptions made by Shokri et al.~\cite{SSSS17} and propose three novel attacks.
Later, Nasr et al.~\cite{NSH19} conduct a comprehensive analysis of membership privacy under both black-box and white-box settings for centralized as well as federated learning scenarios.
Song et al.~\cite{SSM19} study the synergy between adversarial example and membership inference and show that membership privacy risks increase when a model owner applies measures to defend against adversarial example attacks.
To mitigate membership inference, many defense mechanisms have been proposed~\cite{NSH18,SZHBFB19,JSBZG19}.
Nasr et al.~\cite{NSH18} introduce an adversarial regularization term into a target model's loss function.
Salem et al.~\cite{SZHBFB19} propose to use dropout and model stacking to reduce model overfitting, the main reason behind the success of membership inference.
Jia et al.~\cite{JSBZG19} rely on adversarial examples to craft noise to add to a target sample's posteriors.
Also, deferentially private methods~\cite{PSMRTE18,NSTPC21} are introduced to mitigate membership inference.

\mypara{Attribute Inference Attack}
Another major type of privacy attack against ML models is attribute inference.
Here, an adversary aims to infer a specific sensitive attribute of a data sample from its representation generated by a target model~\cite{MSCS19,SS20}.
Melis et al.~\cite{MSCS19} propose the first attribute inference attack against machine learning, in particular federated learning.
Song and Shmatikov~\cite{SS20} later show that attribute inference attacks are also effective against another training paradigm, namely model partitioning.
They further demonstrate that the success of attribute inference is due to the overlearning behavior of ML models.
More recently, Song and Raghunathan~\cite{SR20} demonstrate that language models are also vulnerable to attribute inference.

\mypara{Other Attacks Against Machine Learning Models}
Besides membership inference and attribute inference, there exist a plethora of other attacks against ML models~\cite{SRS17,PMSW18,JOBLNL18,SSTS20,PZJY20,BWTRPOKB20,SBBFZ20,CTWJHLRBSEOR20,LMXZ20,HJBGZ21}.
One major attack is adversarial example~\cite{BCMNSLGR13,PMJFCS16,TKPGBM17,CW17}, where an adversary aims to add imperceptible noises to data samples to evade a target ML model.
Another representative attack in this domain is model extraction, the goal of which is to learn a target model's parameters~\cite{TZJRR16,OSF19,JCBKP20,KTPPI20} or hyperparameters~\cite{WG18,OASF18}.

% ----------------------------------------------------
\section{Discussion}
\label{section:Discussion}
% ----------------------------------------------------

\mypara{Other Types of Datasets}
In this paper, we only focus on image datasets, as most of the current efforts on contrastive learning concentrate on the image domain. 
For other types of datasets like texts or graphs, the main challenge is to define a suitable augmentation method for the input sample. 
There indeed exist some preliminary works of contrastive learning over texts or graphs~\cite{GNWB21,YCSCWS20}. 
However, the effectiveness of these methods still needs to be further evaluated. 
We believe it is straightforward to extend our analysis to contrastive models trained on other types of data.

\mypara{Novel Membership Inference Attacks Against Contrastive Models}
Traditional membership inference attacks use the original data samples to query the model and get the corresponding posteriors to launch the attacks.
However, such attacks is less effective on contrastive models as shown in our paper.
Since the contrastive model is trained with some augmented views of each data sample, the model itself may remember these augmented views as well.
This inspires us to use the augmented views of the original training sample to query the contrastive model to obtain multiple posteriors (one for each augmented version), and aggregate these posteriors as the input to the membership inference attack model.
However, our initial attempt in this direction does not achieve a stronger attack.
One reason might be our aggregation method is not optimal (we have tried averaging and concatenation).
In the future, we plan to investigate more advanced aggregation operations to establish a membership inference attack tailored to contrastive models.

% ----------------------------------------------------
\section{Conclusion}
\label{section:Conclusion}
% ----------------------------------------------------

In this paper, we perform the first privacy quantification of the most representative self-supervised learning paradigm, i.e., contrastive learning.
Concretely, we investigate the privacy risks of contrastive models trained on image datasets through the lens of membership inference and attribute inference.
Empirical evaluation shows that contrastive models are less vulnerable to membership inference attacks compared to supervised models.
This is due to the fact that contrastive models are normally less overfitted.
Meanwhile, contrastive models are more prone to attribute inference attacks.
We posit this is because contrastive models can generate more informative representations for data samples, which can be exploited by an adversary to achieve effective attribute inference.

To reduce the risks of attribute inference stemming from contrastive models, we propose the first privacy-preserving contrastive learning mechanism, namely \Talos.
Specifically, \Talos introduces an adversarial classifier to censor the sensitive attributes learned by the contrastive models under the adversarial training framework.
Our evaluation shows that \Talos can effectively mitigate the attribute inference risks for contrastive models while maintaining their membership privacy and model utility.

% ----------------------------------------------------
\section*{Acknowledgements}
\label{section:Acknowledgements}
% ----------------------------------------------------
This work is partially funded by the Helmholtz Association within the project ``Trustworthy Federated Data Analytics'' (TFDA) (funding number ZT-I-OO1 4).

% ----------------------------------------------------
\bibliographystyle{plain}
\bibliography{normal_generated_py3}
% ----------------------------------------------------

% ----------------------------------------------------
\newpage
\appendix
\section{Appendix}
\label{section:Appendix}
% ----------------------------------------------------

\begin{figure}[htbp]
\centering
\begin{subfigure}{\columnwidth}
\includegraphics[width=\columnwidth]{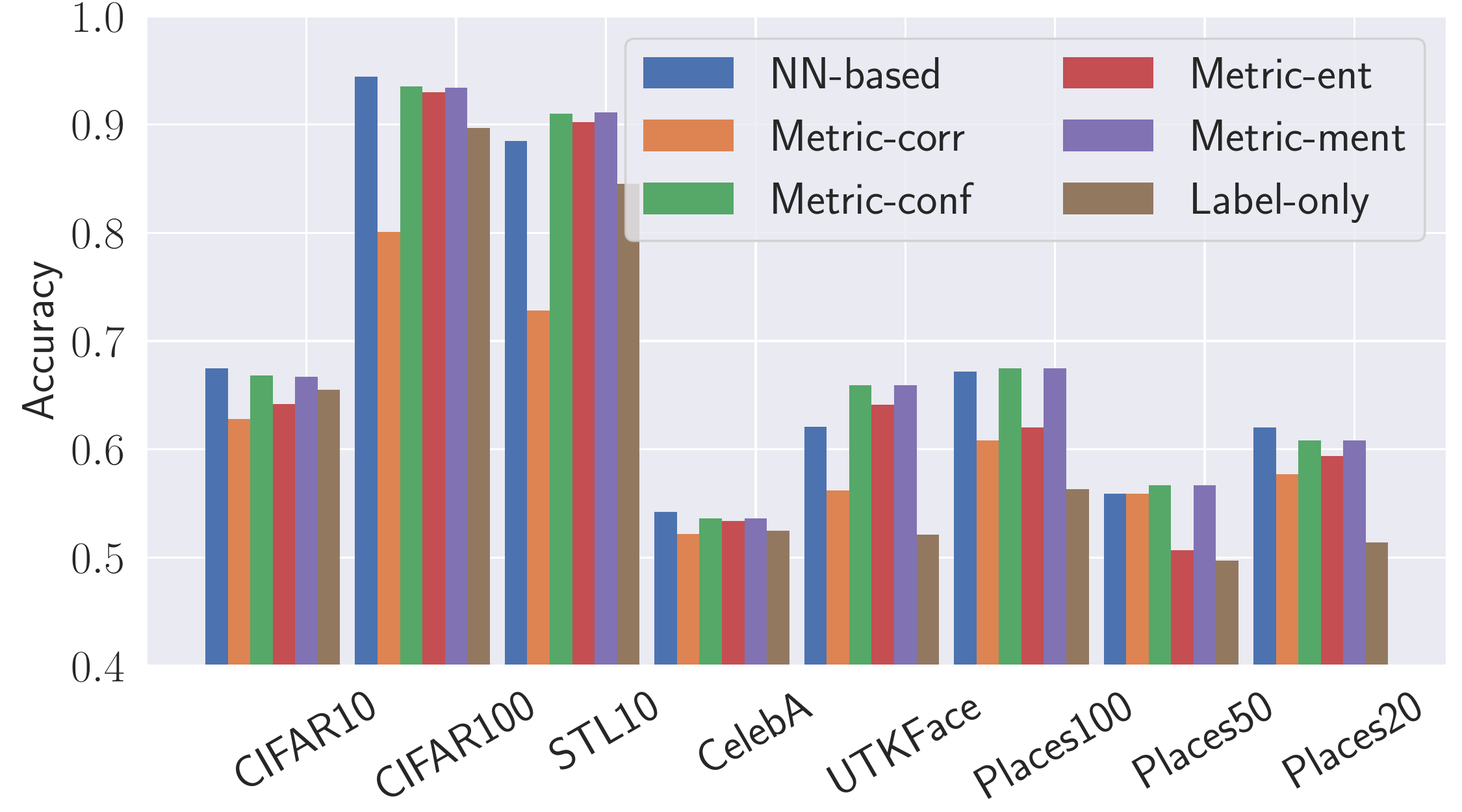}
\caption{Supervised Model}
\label{figure:mia_attack_supervised_resnet18}
\end{subfigure}
\begin{subfigure}{\columnwidth}
\includegraphics[width=\columnwidth]{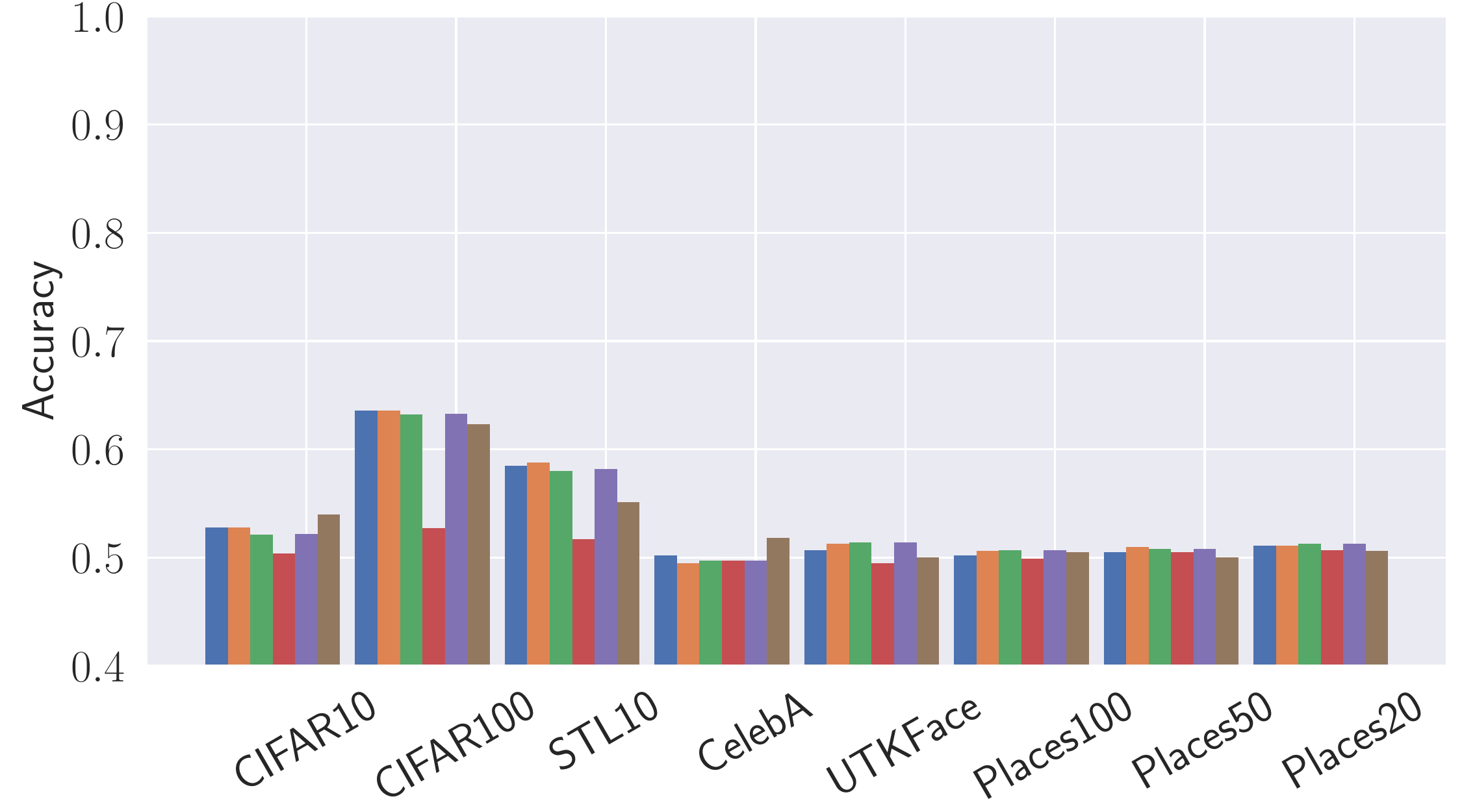}
\caption{Contrastive Model}
\label{figure:mia_attack_contrastive_resnet18}
\end{subfigure}
\caption{The performance of different membership inference attacks against both supervised models and contrastive models with ResNet-18 on 8 different datasets.
The x-axis represents different datasets.
The y-axis represents membership inference attacks' accuracy.}
\label{figure:mia_attack_performance_resnet18}
\end{figure}

\begin{figure*}[!ht]
\centering
\begin{subfigure}{\columnwidth}
\includegraphics[width=\columnwidth]{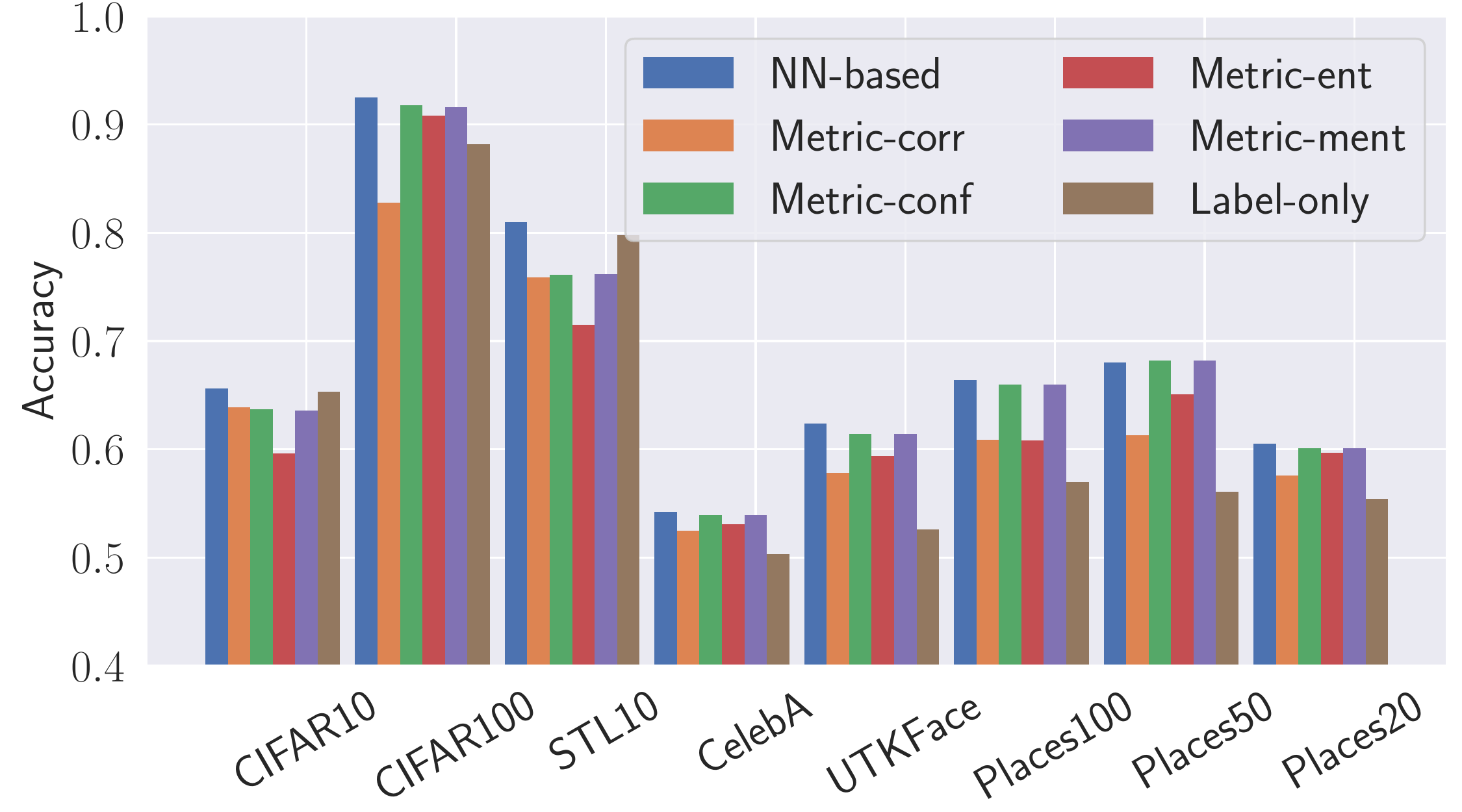}
\caption{Supervised Model}
\label{figure:mia_attack_supervised_resnet50}
\end{subfigure}
\begin{subfigure}{\columnwidth}
\includegraphics[width=\columnwidth]{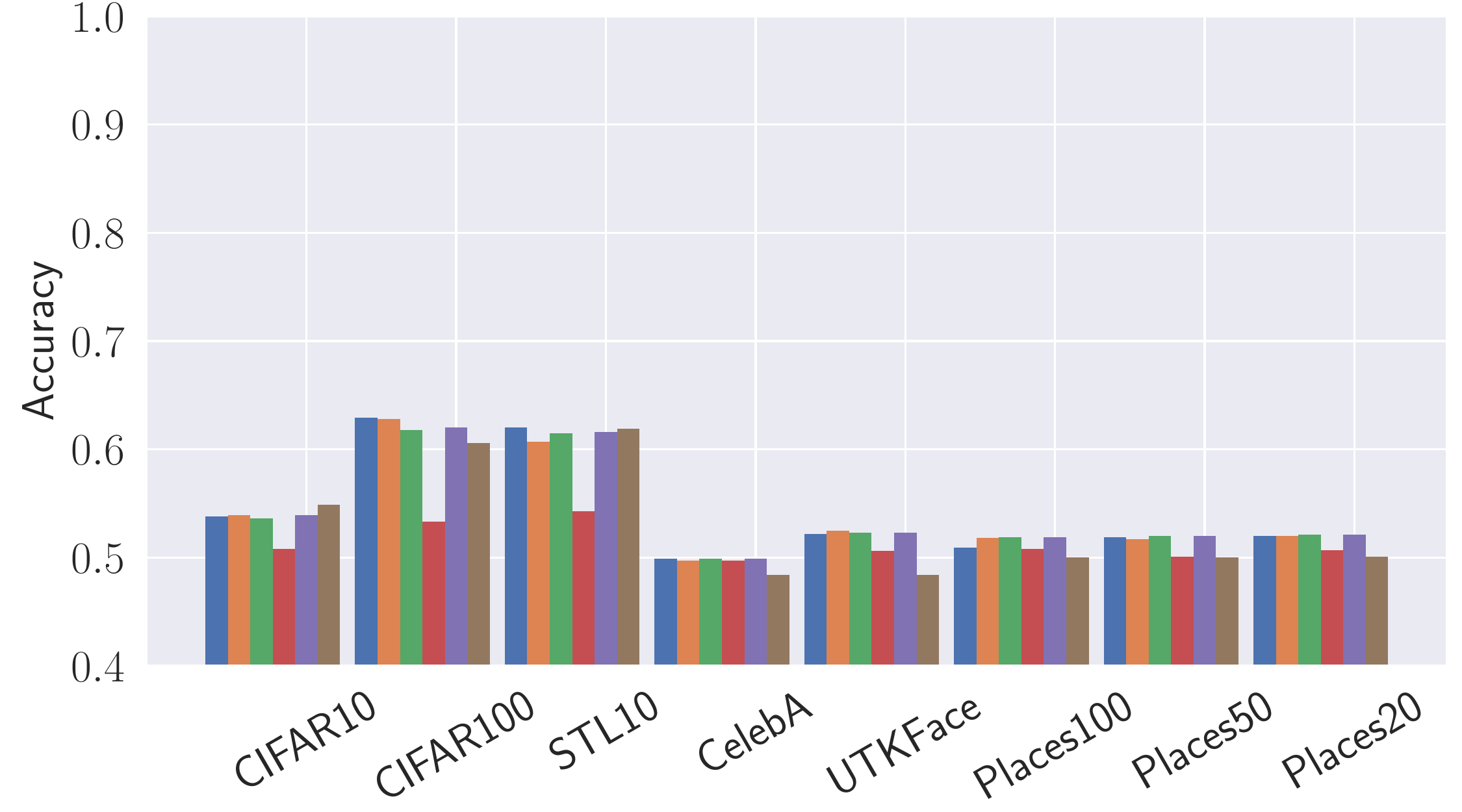}
\caption{Contrastive Model}
\label{figure:mia_attack_contrastive_resnet50}
\end{subfigure}
\caption{The performance of different membership inference attacks against both supervised models and contrastive models with ResNet-50 on 8 different datasets.
The x-axis represents different datasets.
The y-axis represents membership inference attacks' accuracy.}
\label{figure:mia_attack_performance_resnet50}
\end{figure*}

\begin{figure*}[!ht]
\centering
\begin{subfigure}{0.5\columnwidth}
\includegraphics[width=\columnwidth]{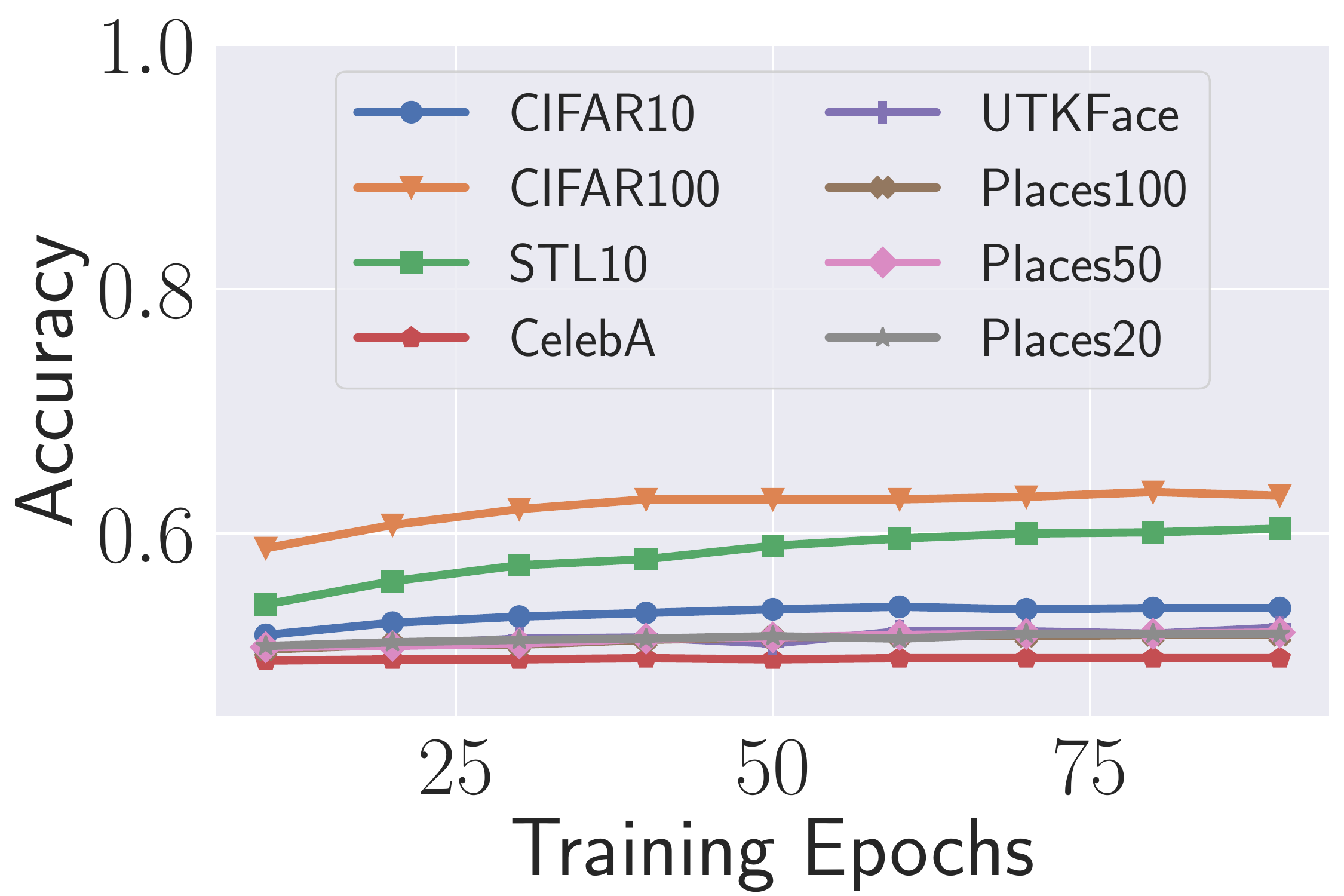}
\caption{Metric-corr}
\label{figure:mia_diff_linear_metric-corr}
\end{subfigure}
\begin{subfigure}{0.5\columnwidth}
\includegraphics[width=\columnwidth]{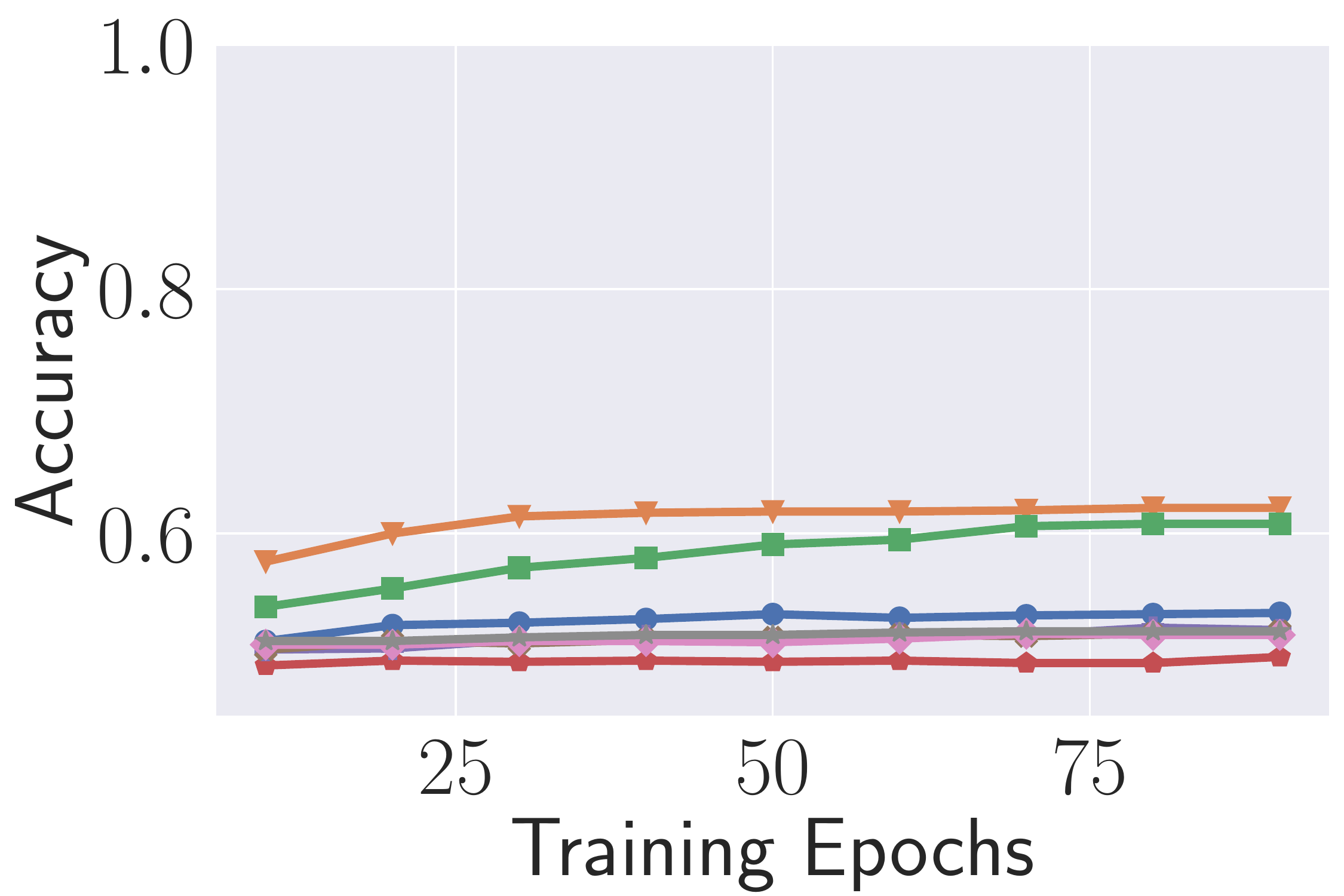}
\caption{Metric-conf}
\label{figure:mia_diff_linear_metric-conf}
\end{subfigure}
\begin{subfigure}{0.5\columnwidth}
\includegraphics[width=\columnwidth]{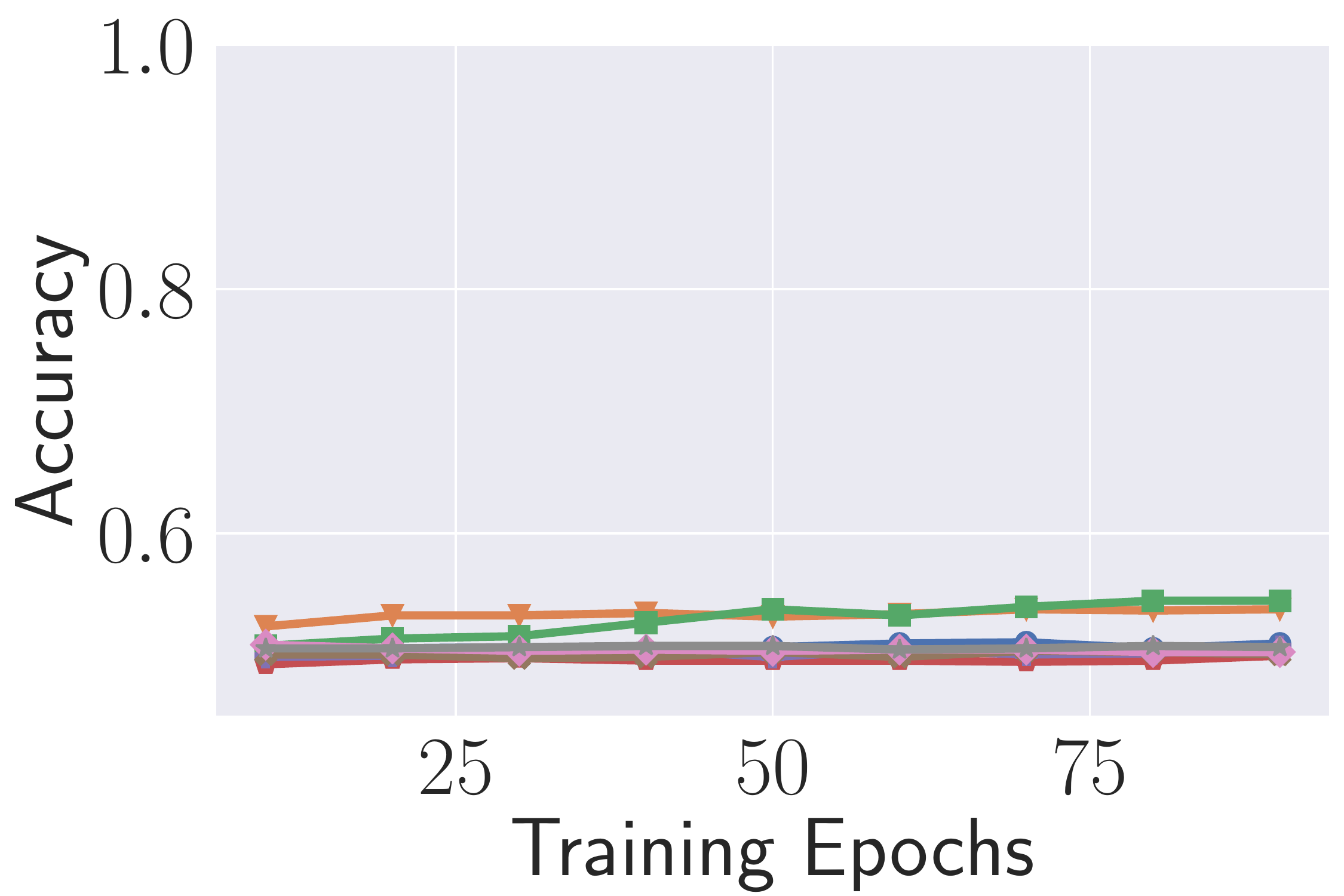}
\caption{Metric-ent}
\label{figure:mia_diff_linear_metric-ent}
\end{subfigure}
\begin{subfigure}{0.5\columnwidth}
\includegraphics[width=\columnwidth]{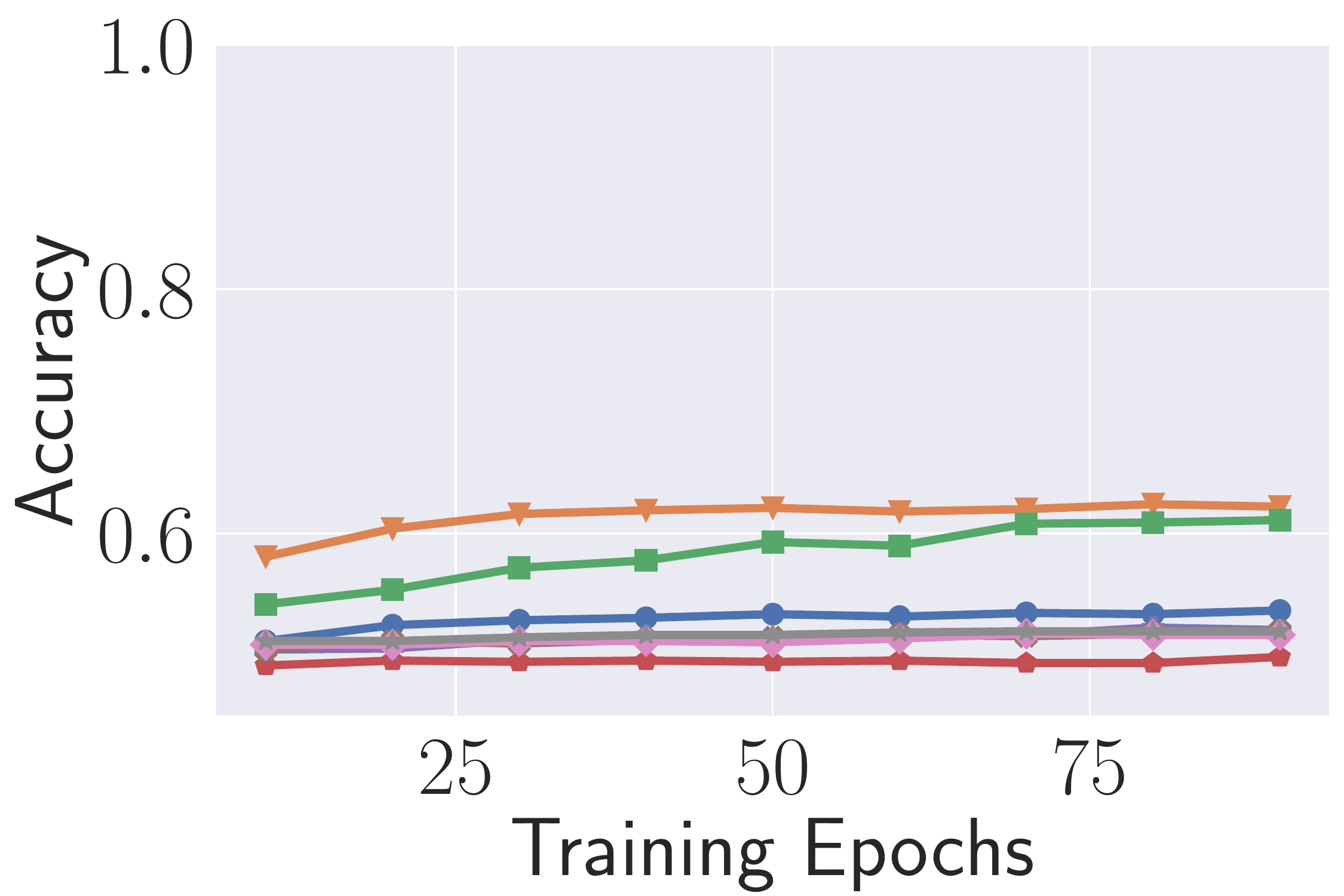}
\caption{Metric-ment}
\label{figure:mia_diff_linear_metric-ment}
\end{subfigure}
\caption{The performance of metric-based membership inference attacks against contrastive models with ResNet-50 on 8 different datasets under different numbers of epochs for classification layer training.
The x-axis represents different numbers of epochs.
The y-axis represents membership inference attacks' accuracy.
Each line corresponds to a specific dataset.}
\label{figure:mia_diff_linear_metric-based}
\end{figure*}

\begin{figure*}[!t]
\centering
\begin{subfigure}{0.5\columnwidth}
\includegraphics[width=\columnwidth]{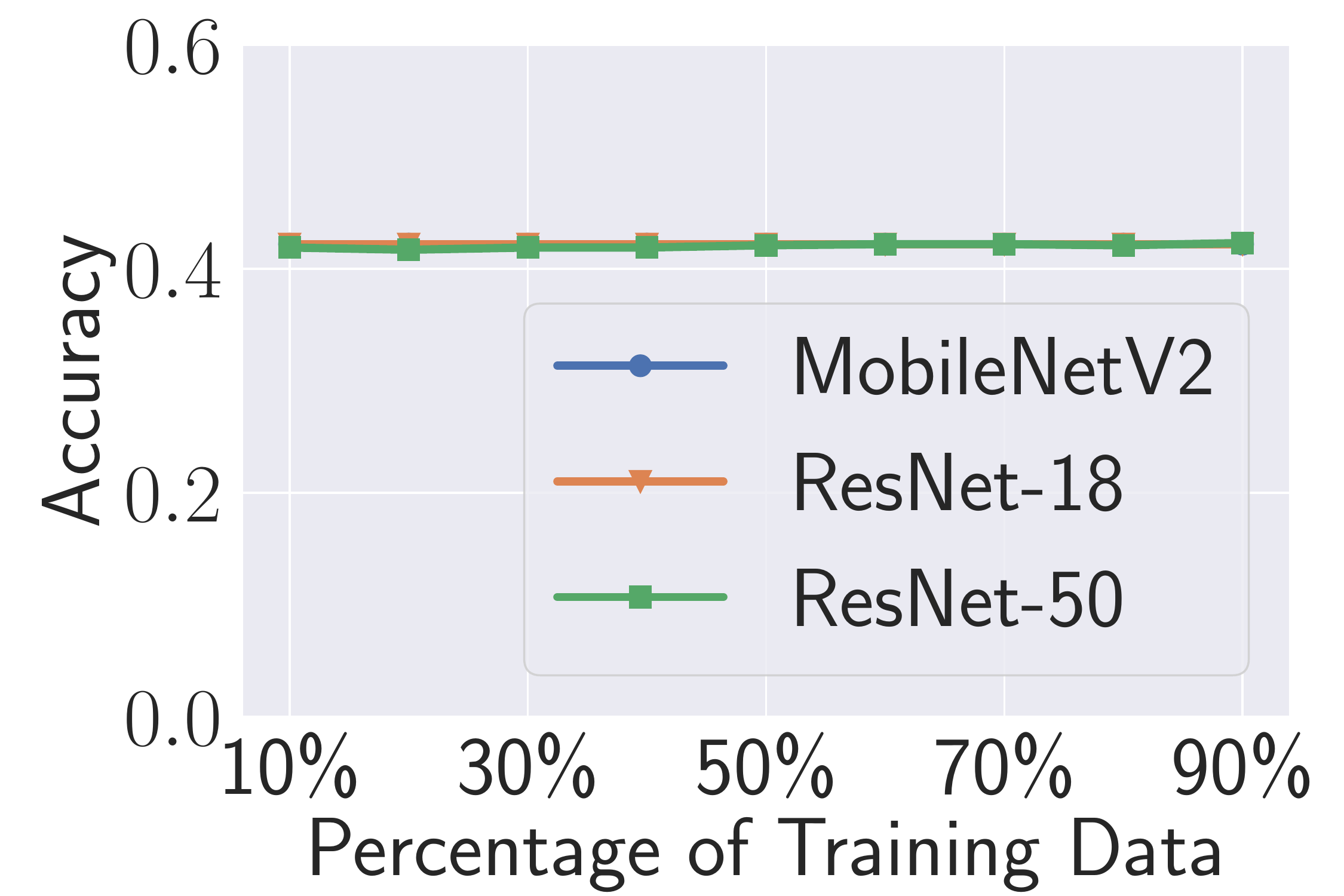}
\caption{UTKFace}
\label{figure:ai_diff_ratio_ce_utkface}
\end{subfigure}
\begin{subfigure}{0.5\columnwidth}
\includegraphics[width=\columnwidth]{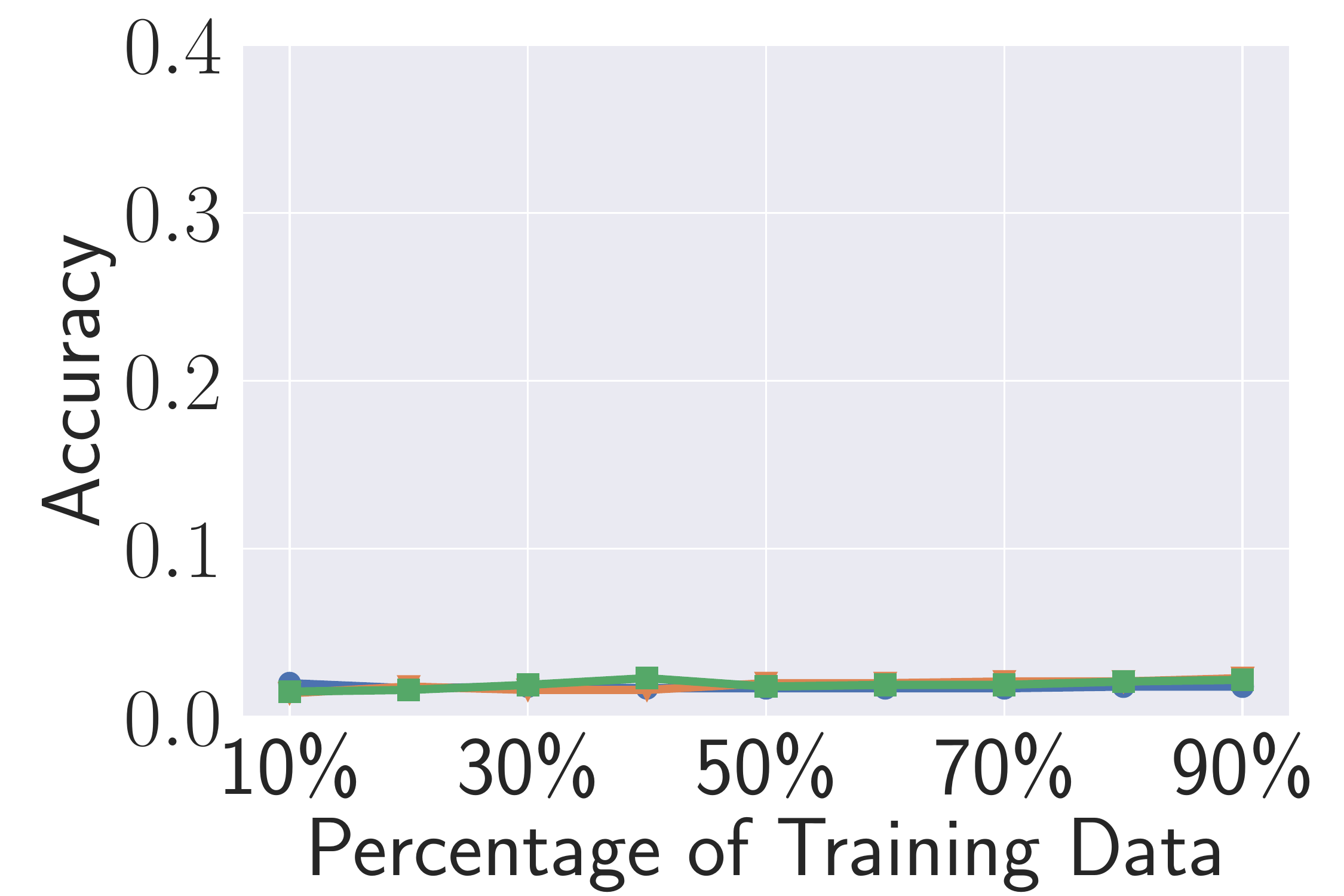}
\caption{Places100}
\label{figure:ai_diff_ratio_ce_place100}
\end{subfigure}
\begin{subfigure}{0.5\columnwidth}
\includegraphics[width=\columnwidth]{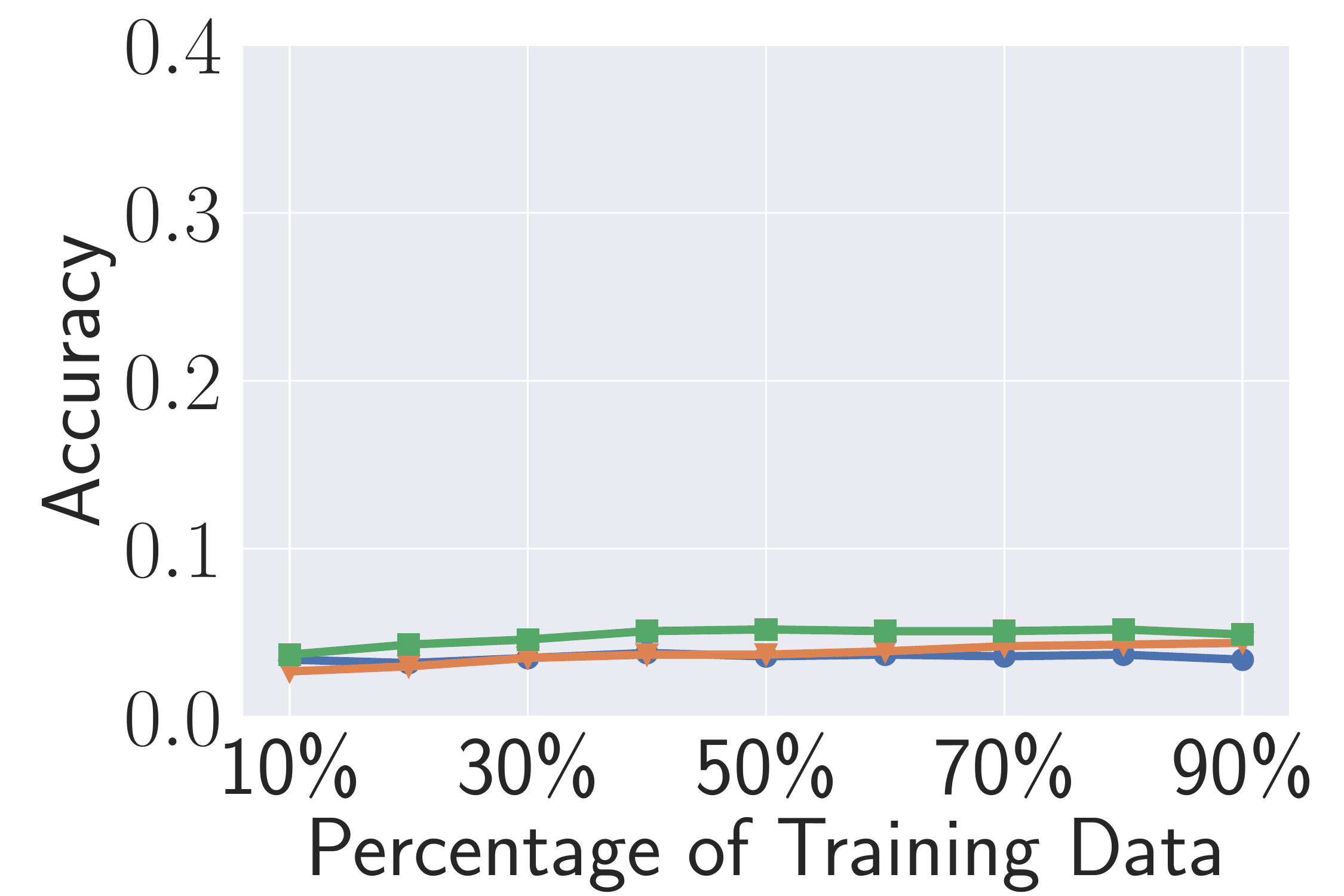}
\caption{Places50}
\label{figure:ai_diff_ratio_ce_place50}
\end{subfigure}
\begin{subfigure}{0.5\columnwidth}
\includegraphics[width=\columnwidth]{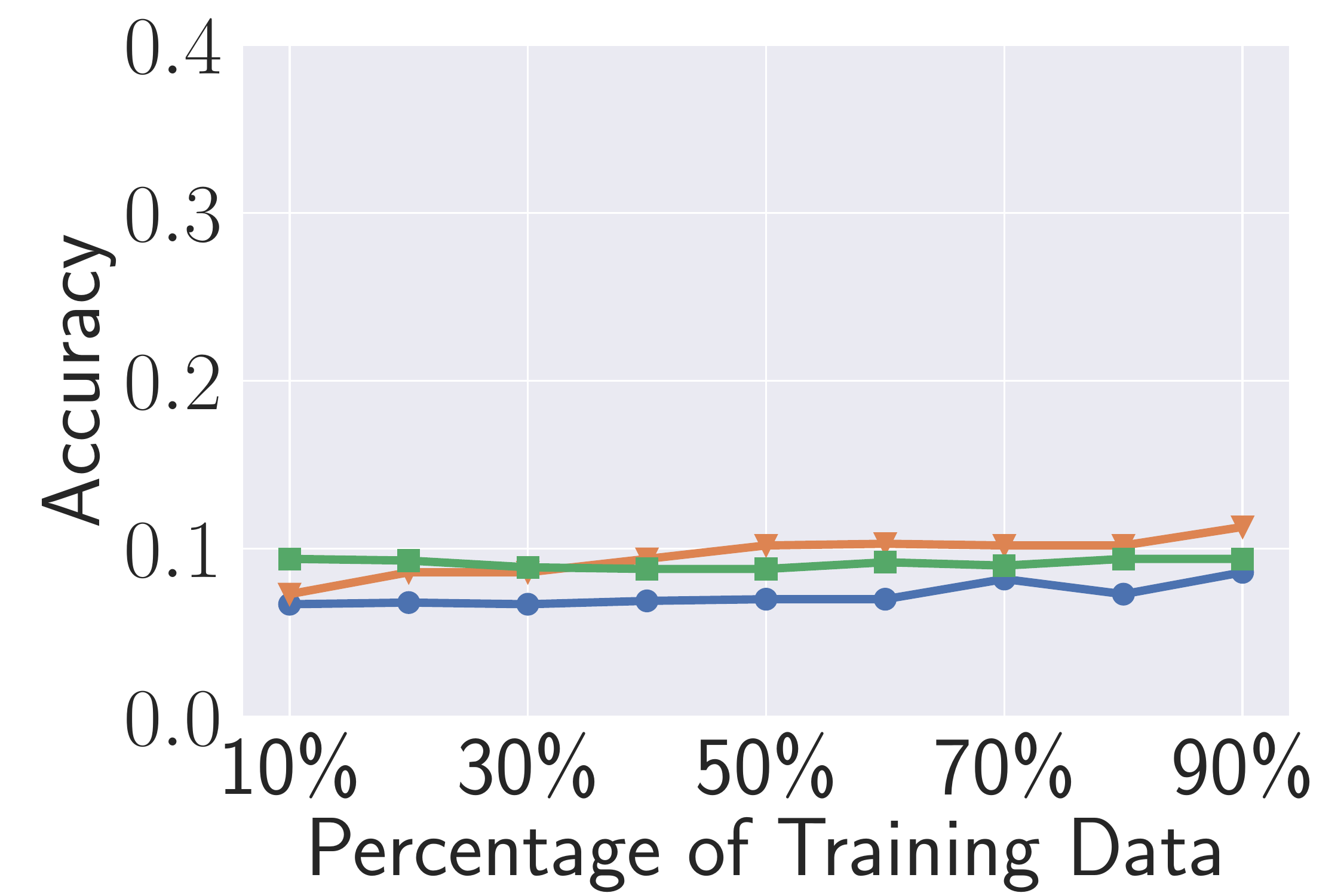}
\caption{Places20}
\label{figure:ai_diff_ratio_ce_place20}
\end{subfigure}
\caption{The performance of attribute inference attacks against supervised models on 4 different datasets under different percentages of the attack training dataset.
The x-axis represents different percentages of the attack training dataset.
The y-axis represents attribute inference attacks' accuracy.}
\label{figure:ai_diff_ratio_ce} 
\end{figure*}

\begin{figure*}[!ht]
\centering
\begin{subfigure}{0.5\columnwidth}
\includegraphics[width=\columnwidth]{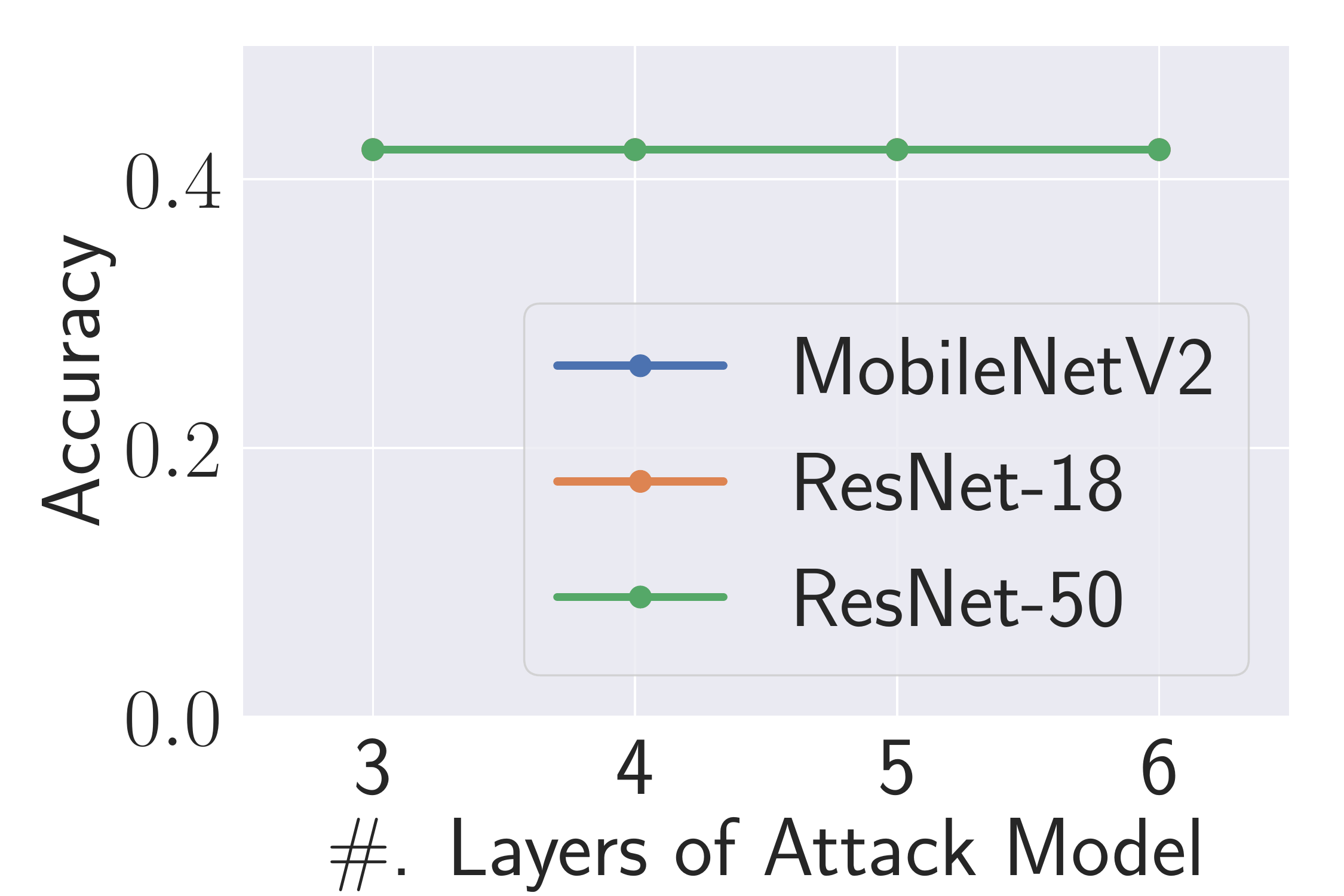}
\caption{UTKFace}
\label{figure:ai_diff_attack_layer_utkface_CE}
\end{subfigure}
\begin{subfigure}{0.5\columnwidth}
\includegraphics[width=\columnwidth]{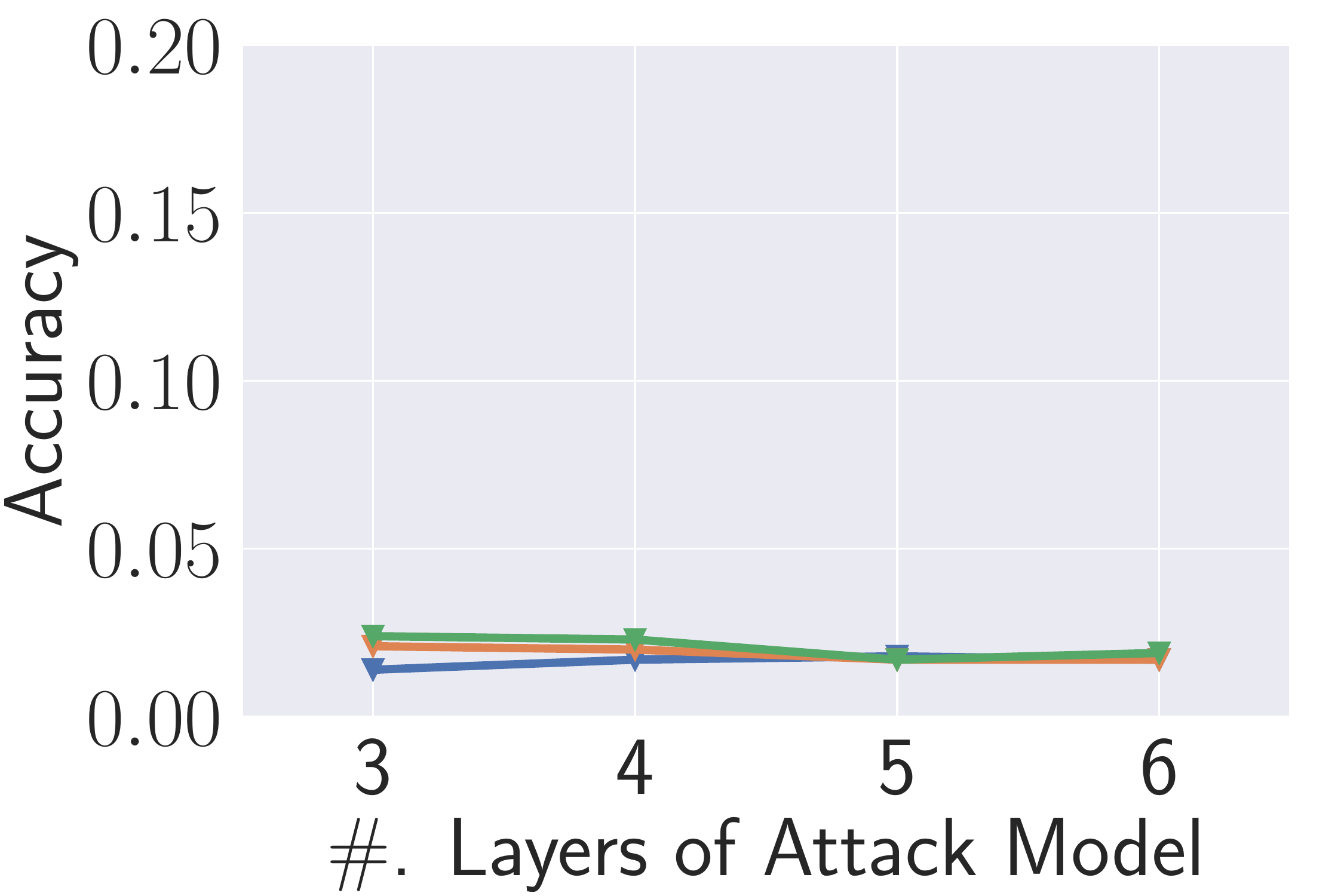}
\caption{Places100}
\label{figure:ai_diff_attack_layer_place100_CE}
\end{subfigure}
\begin{subfigure}{0.5\columnwidth}
\includegraphics[width=\columnwidth]{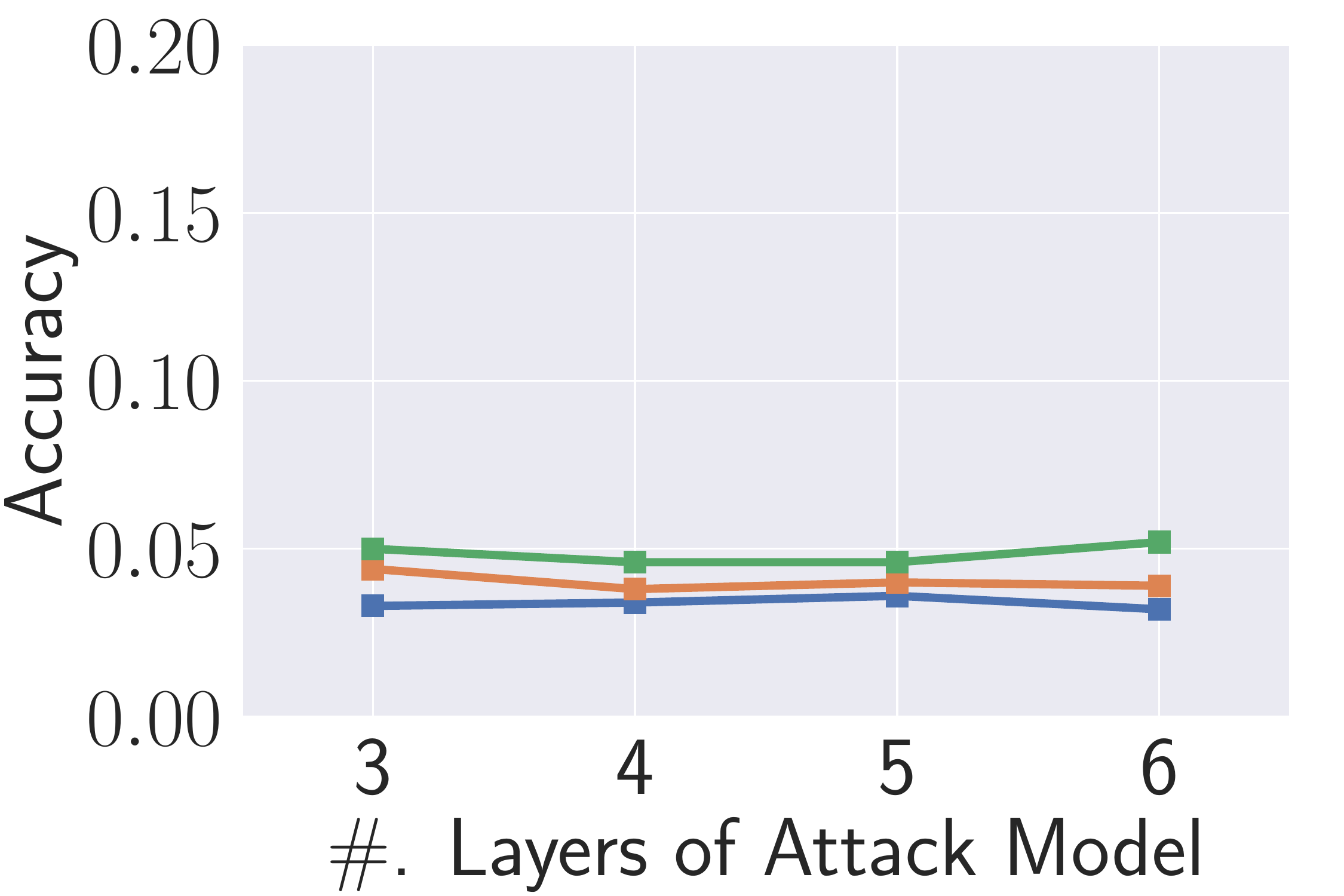}
\caption{Places50}
\label{figure:ai_diff_attack_layer_place50_CE}
\end{subfigure}
\begin{subfigure}{0.5\columnwidth}
\includegraphics[width=\columnwidth]{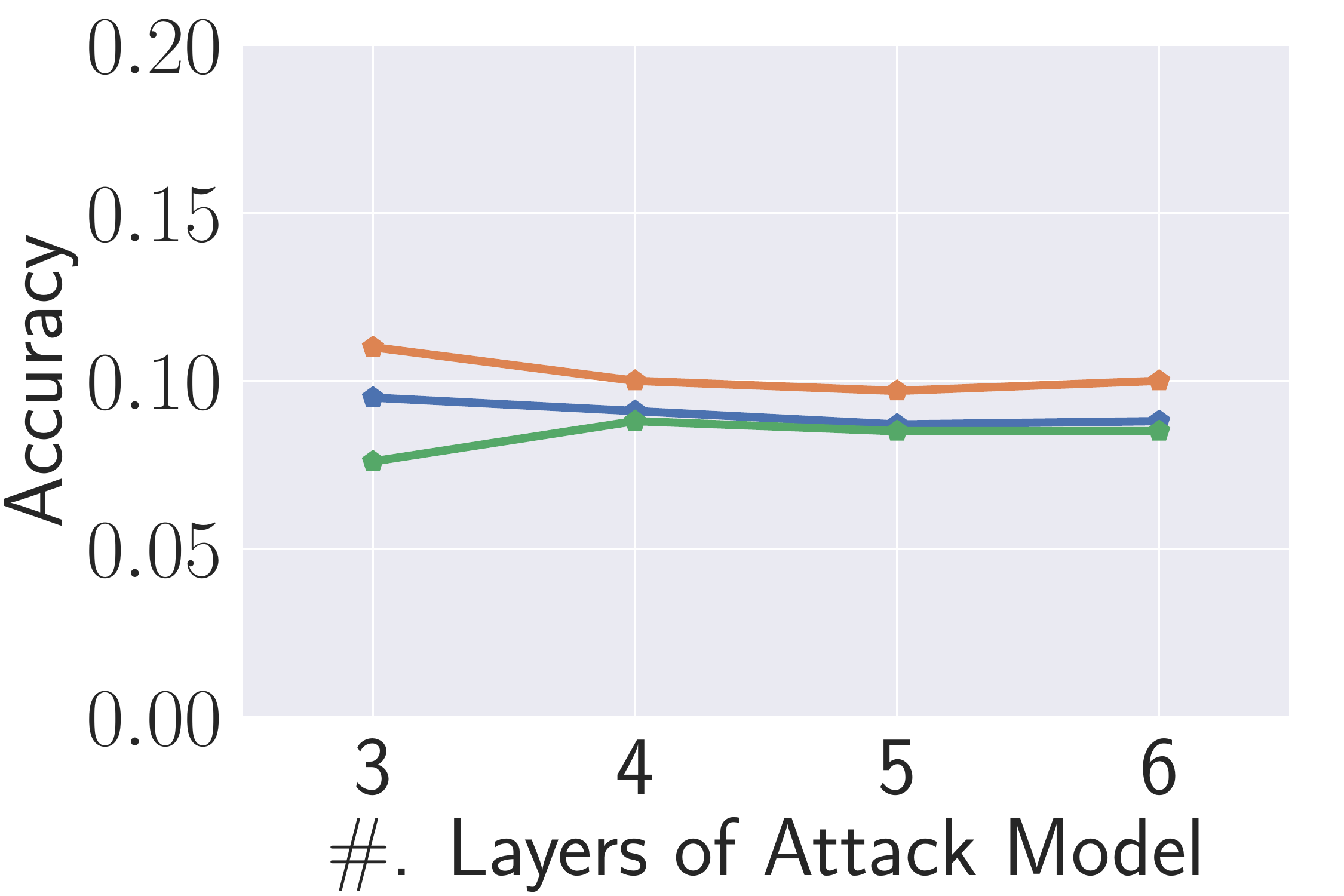}
\caption{Places20}
\label{figure:ai_diff_attack_layer_place20_CE}
\end{subfigure}
\caption{The performance of attribute inference attacks against supervised models on 4 different datasets under attack models with different layers.
The x-axis represents attack models' layers.
The y-axis represents attribute inference attacks' accuracy.}
\label{figure:ai_diff_attack_layer_CE} 
\end{figure*}

\begin{figure*}[!ht]
\centering
\begin{subfigure}{0.5\columnwidth}
\includegraphics[width=\columnwidth]{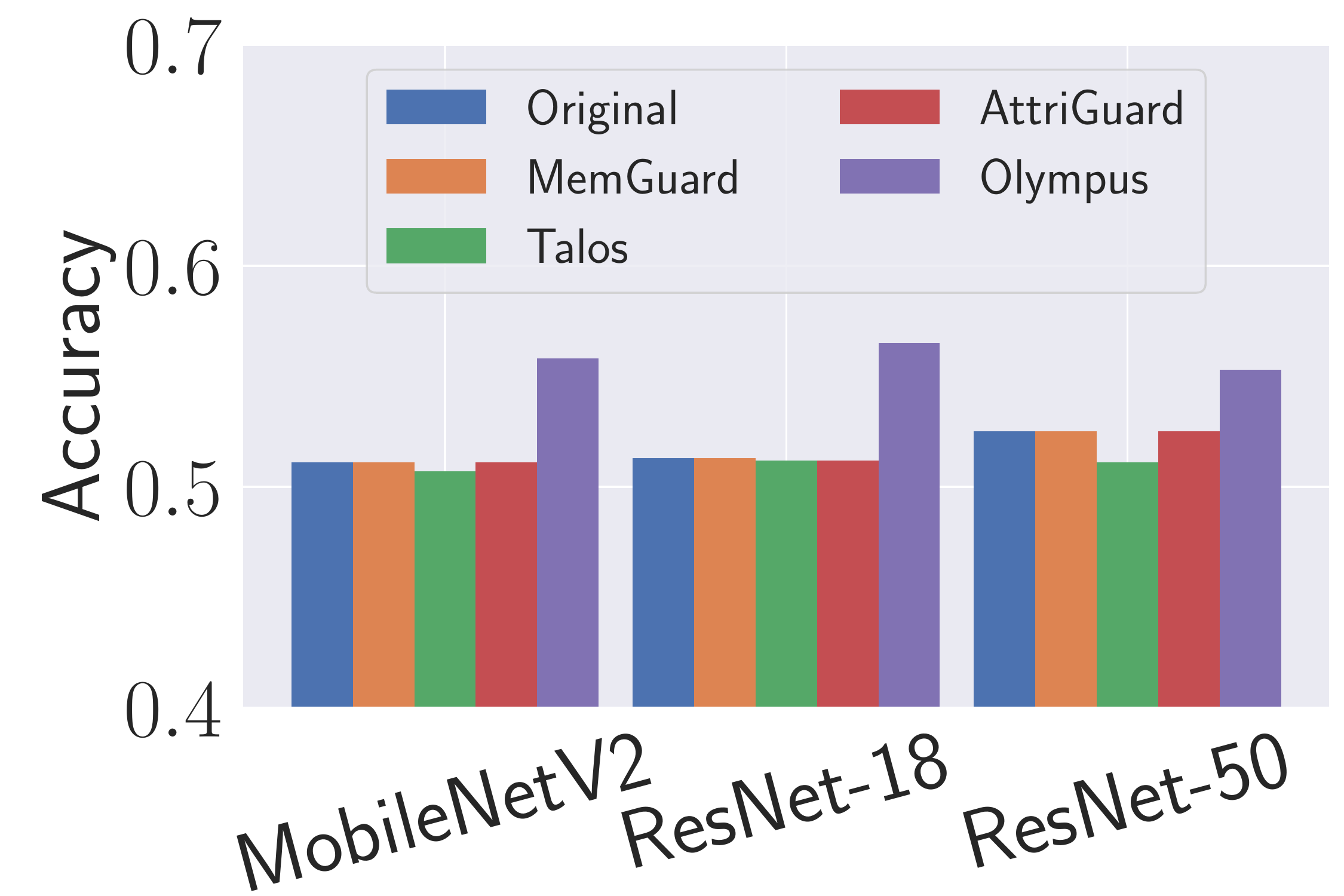}
\caption{UTKFace}
\label{figure:defense_comparison_mia_UTKFace_Metric-corr}
\end{subfigure}
\begin{subfigure}{0.5\columnwidth}
\includegraphics[width=\columnwidth]{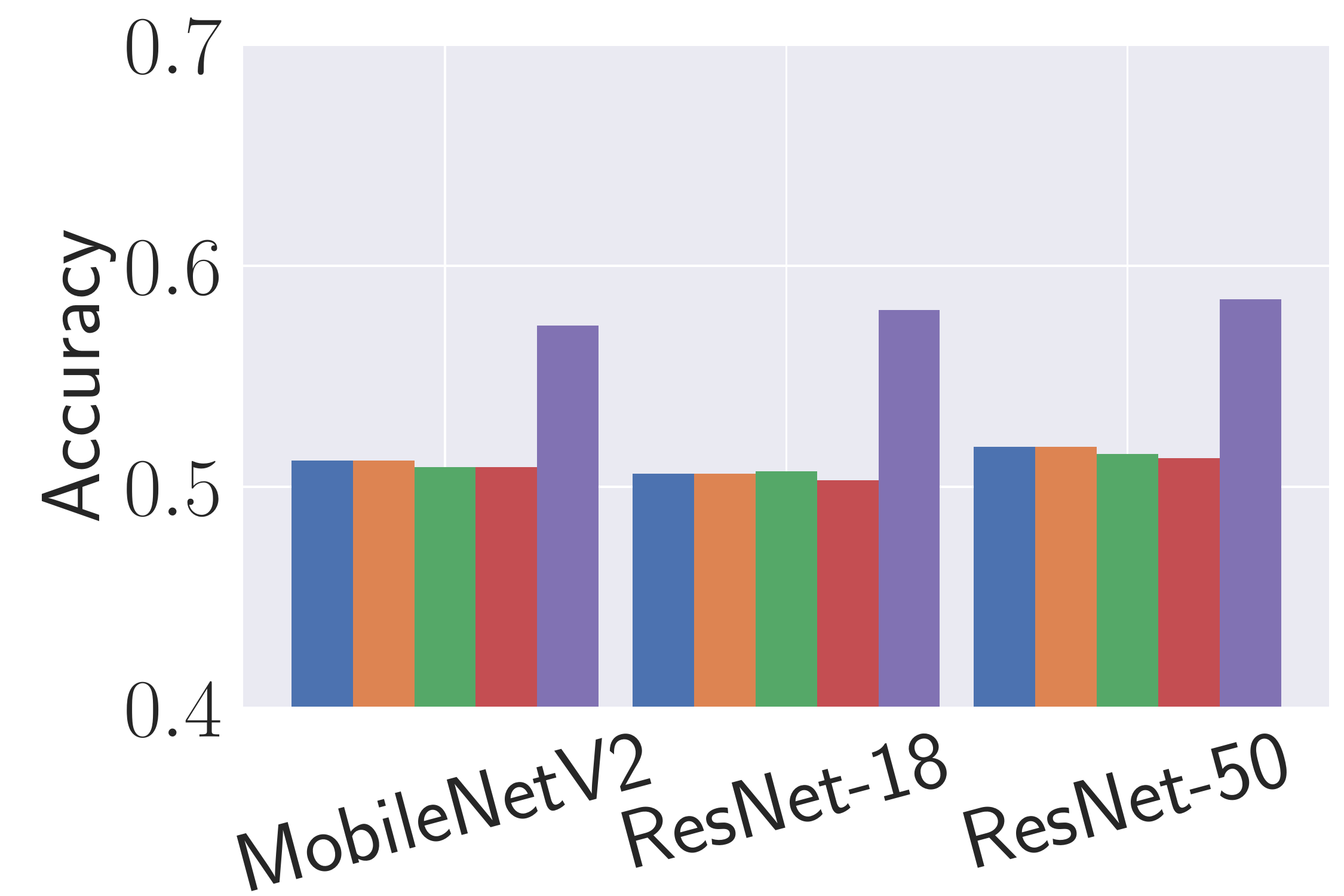}
\caption{Places100}
\label{figure:defense_comparison_mia_Places100_Metric-corr}
\end{subfigure}
\begin{subfigure}{0.5\columnwidth}
\includegraphics[width=\columnwidth]{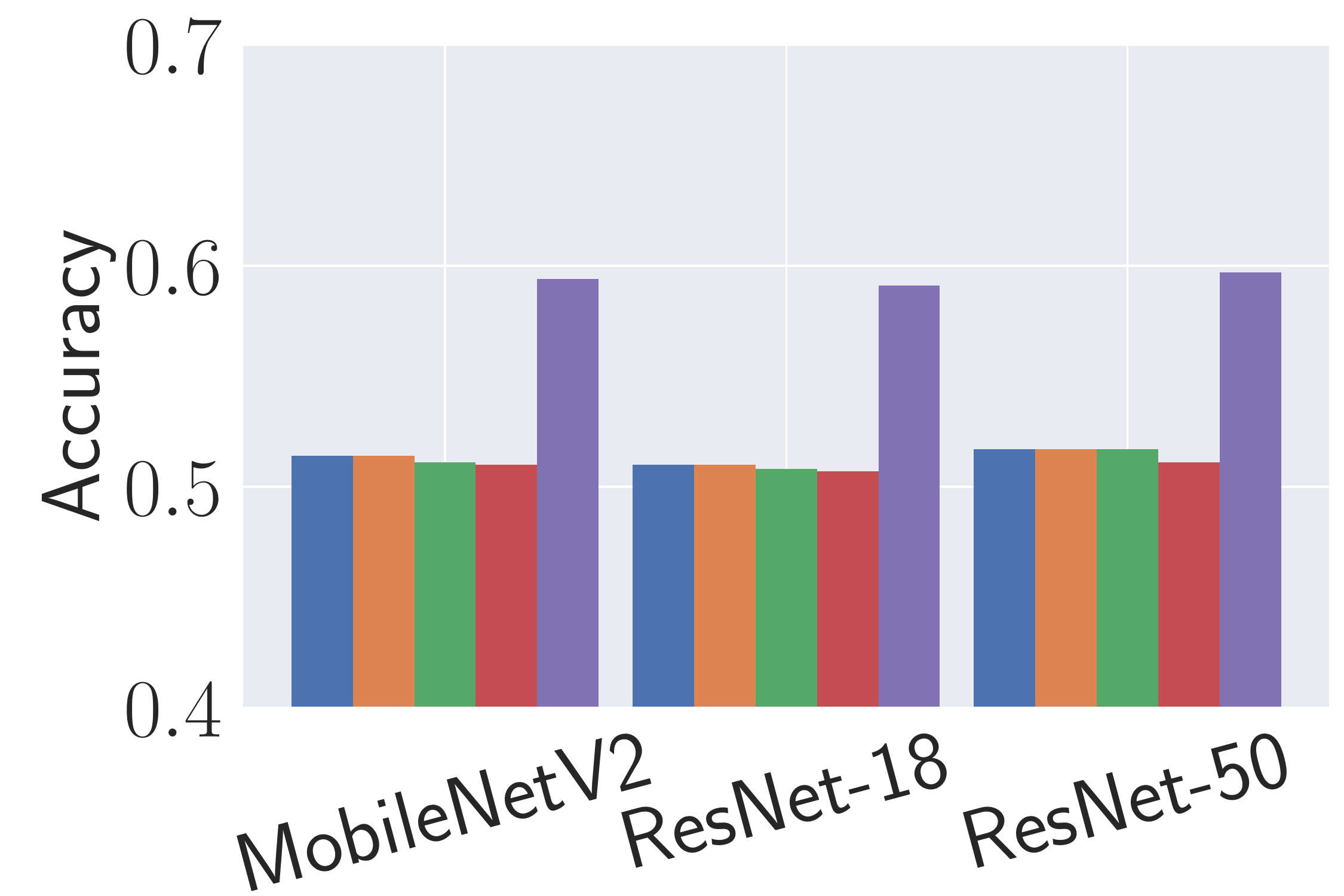}
\caption{Places50}
\label{figure:defense_comparison_mia_Places50_Metric-corr}
\end{subfigure}
\begin{subfigure}{0.5\columnwidth}
\includegraphics[width=\columnwidth]{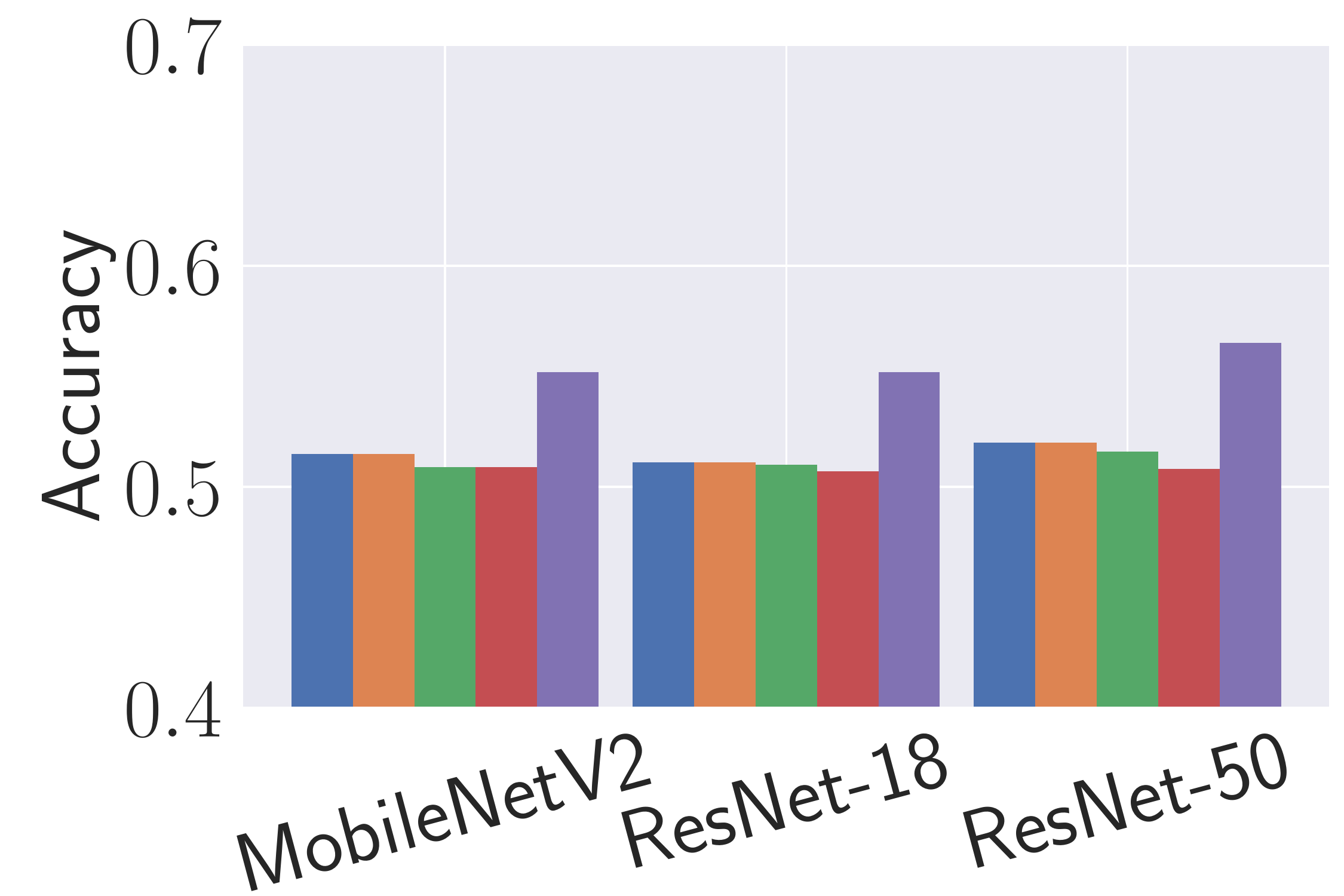}
\caption{Places20}
\label{figure:defense_comparison_mia_Places20_Metric-corr}
\end{subfigure}
\caption{
The performance of metric-corr membership inference attacks against original contrastive models,\Talos, \MemGuard, \Olympus, and \AttriGuard with MobileNetV2, ResNet-18, and ResNet-50 on 4 different datasets.
The x-axis represents different models.
The y-axis represents the accuracy of metric-corr membership inference attacks.}
\label{figure:defense_comparison_mia_Metric-corr}
\end{figure*}

\begin{figure*}[!ht]
\centering
\begin{subfigure}{0.5\columnwidth}
\includegraphics[width=\columnwidth]{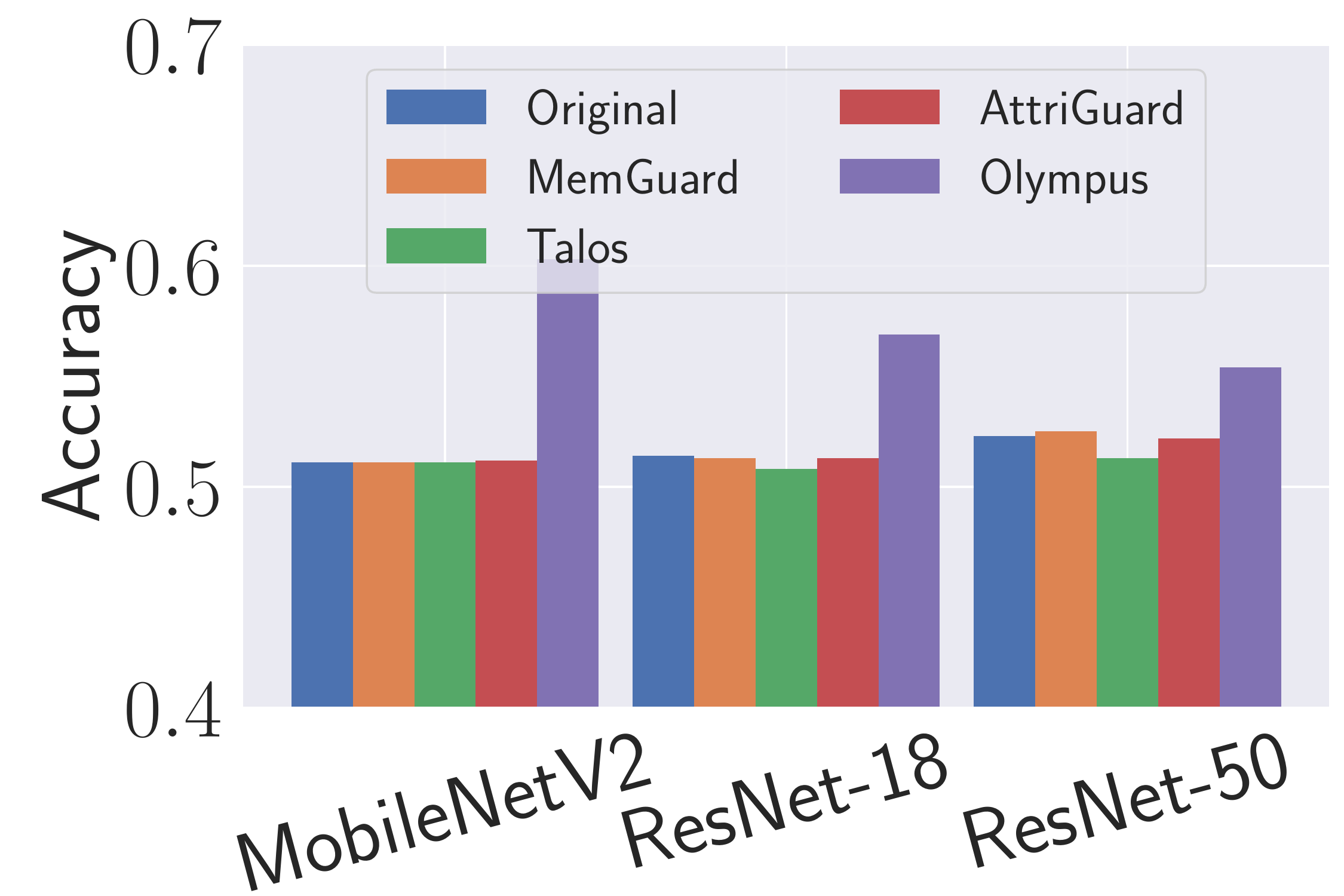}
\caption{UTKFace}
\label{figure:defense_comparison_mia_UTKFace_Metric-conf}
\end{subfigure}
\begin{subfigure}{0.5\columnwidth}
\includegraphics[width=\columnwidth]{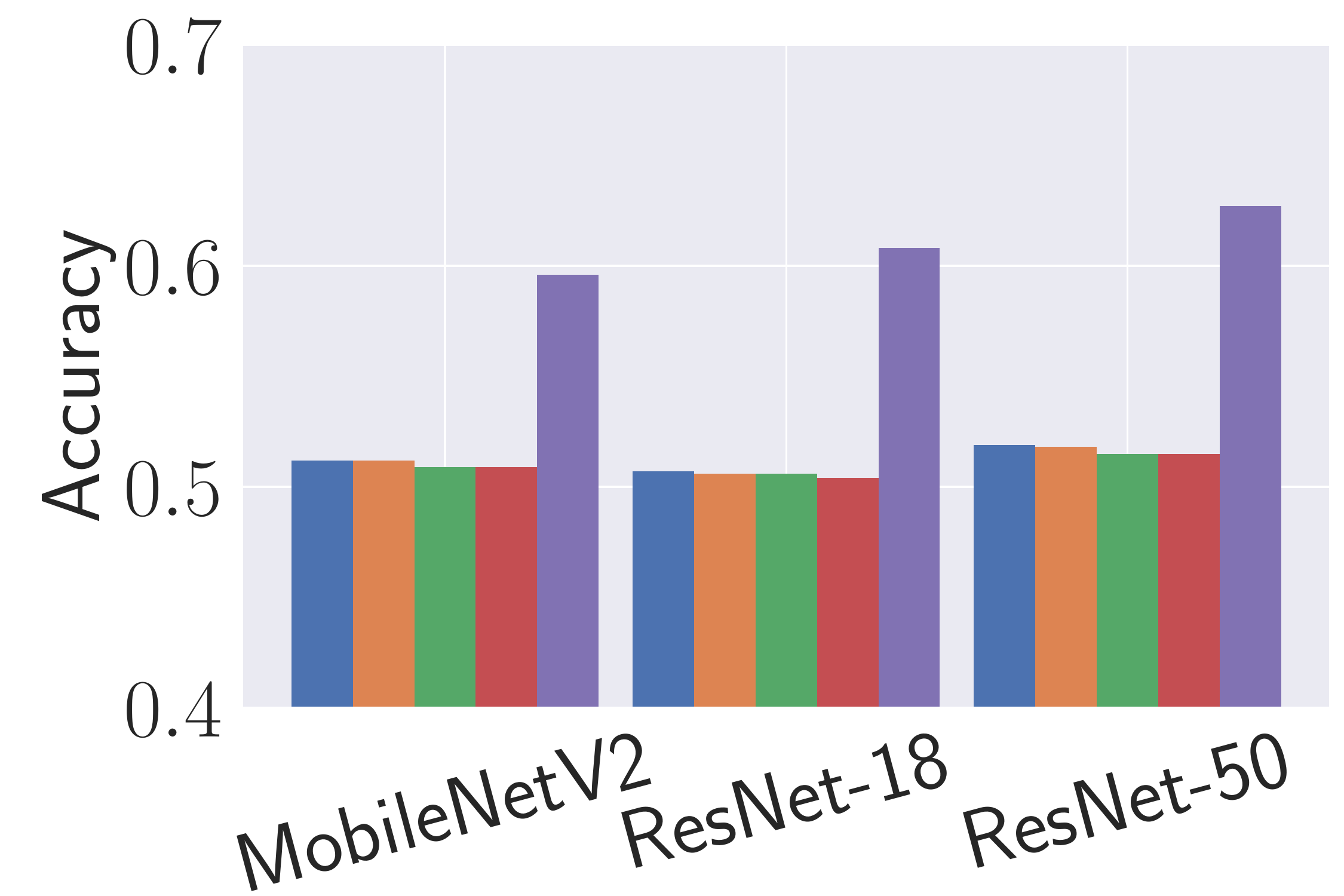}
\caption{Places100}
\label{figure:defense_comparison_mia_Places100_Metric-conf}
\end{subfigure}
\begin{subfigure}{0.5\columnwidth}
\includegraphics[width=\columnwidth]{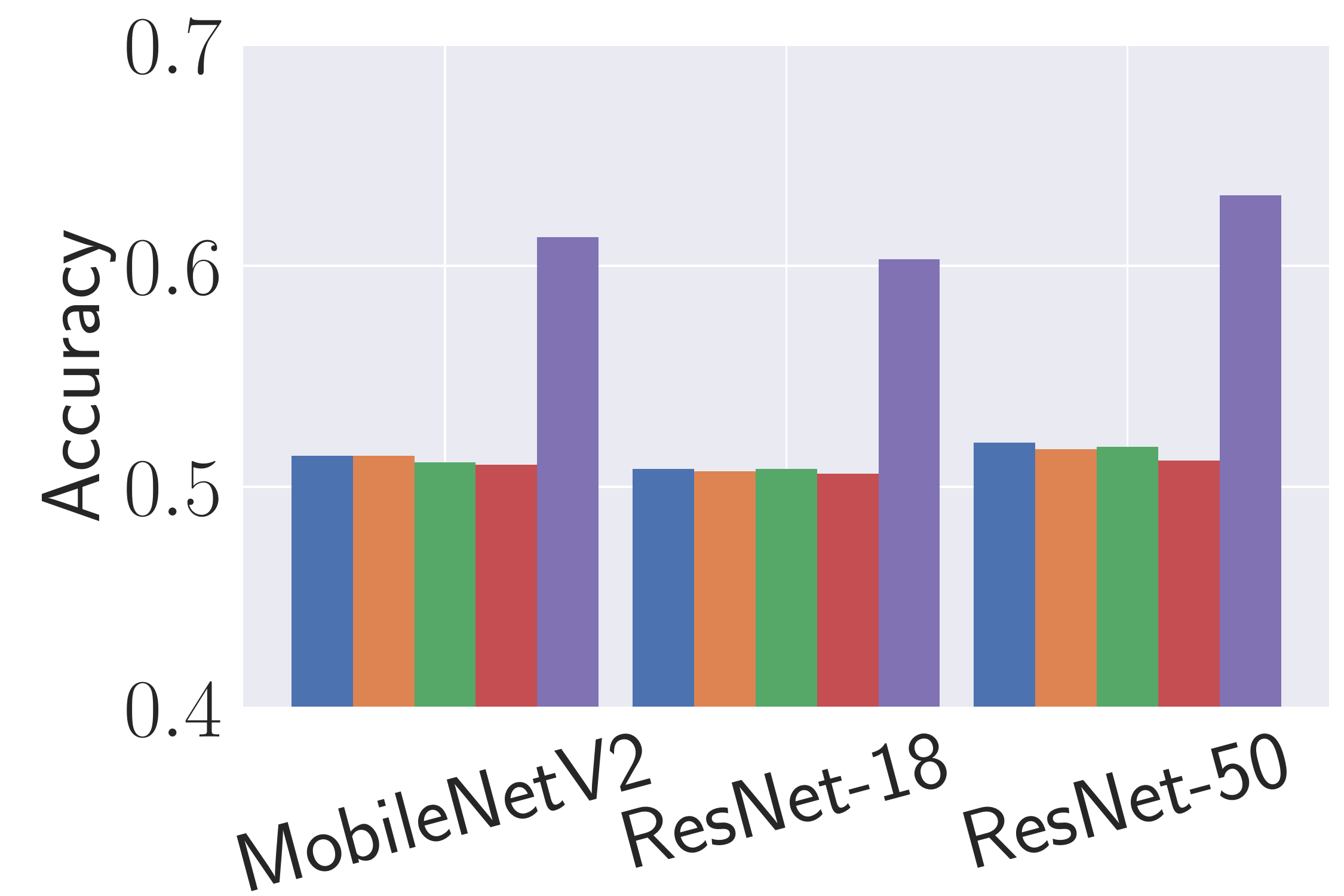}
\caption{Places50}
\label{figure:defense_comparison_mia_Places50_Metric-conf}
\end{subfigure}
\begin{subfigure}{0.5\columnwidth}
\includegraphics[width=\columnwidth]{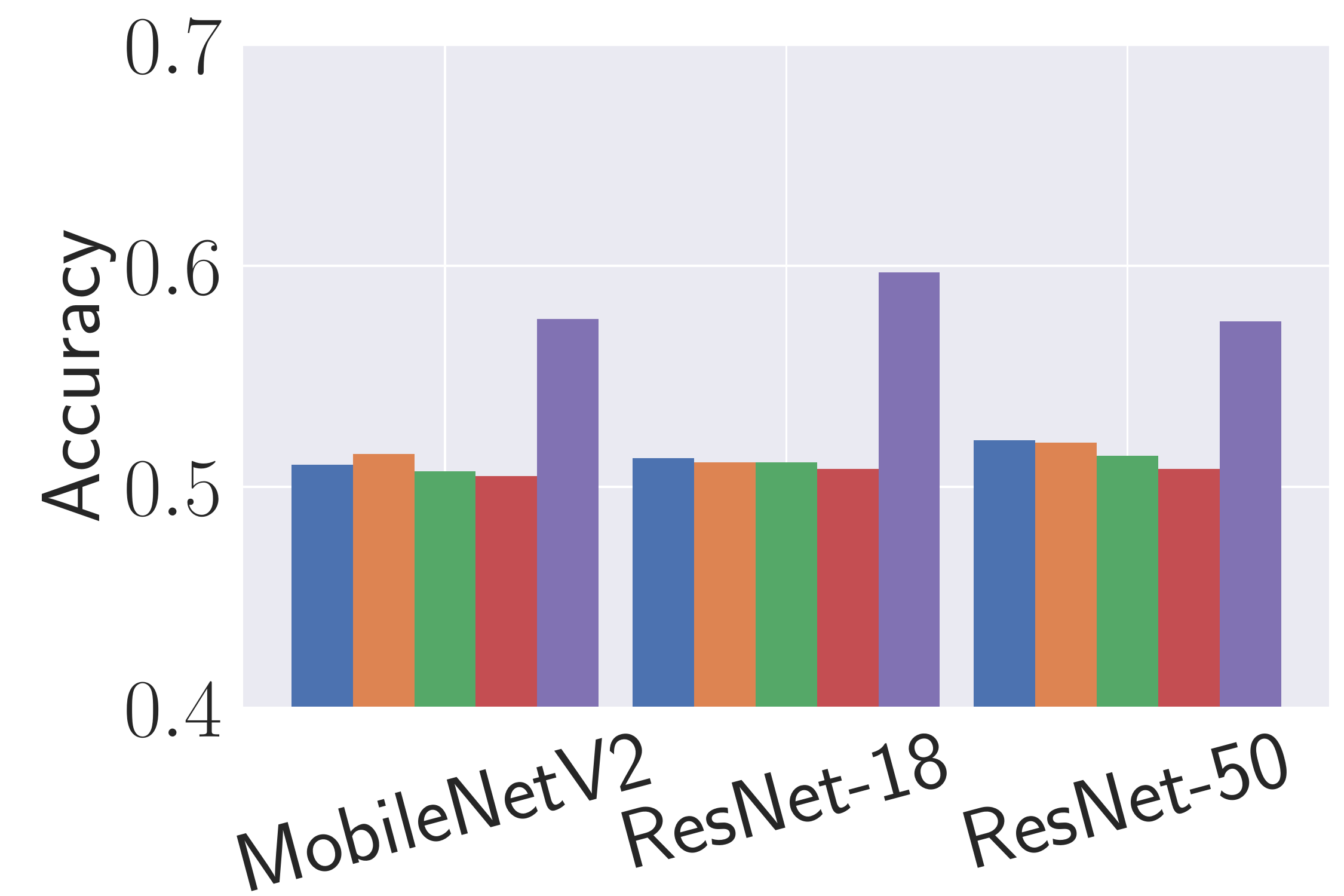}
\caption{Places20}
\label{figure:defense_comparison_mia_Places20_Metric-conf}
\end{subfigure}
\caption{
The performance of metric-conf membership inference attacks against original contrastive models, \Talos, \MemGuard, \Olympus, and \AttriGuard with MobileNetV2, ResNet-18, and ResNet-50 on 4 different datasets.
The x-axis represents different methods.
The y-axis represents the accuracy of metric-conf membership inference attacks.}
\label{figure:defense_comparison_mia_Metric-conf}
\end{figure*}

\begin{figure*}[!ht]
\centering
\begin{subfigure}{0.5\columnwidth}
\includegraphics[width=\columnwidth]{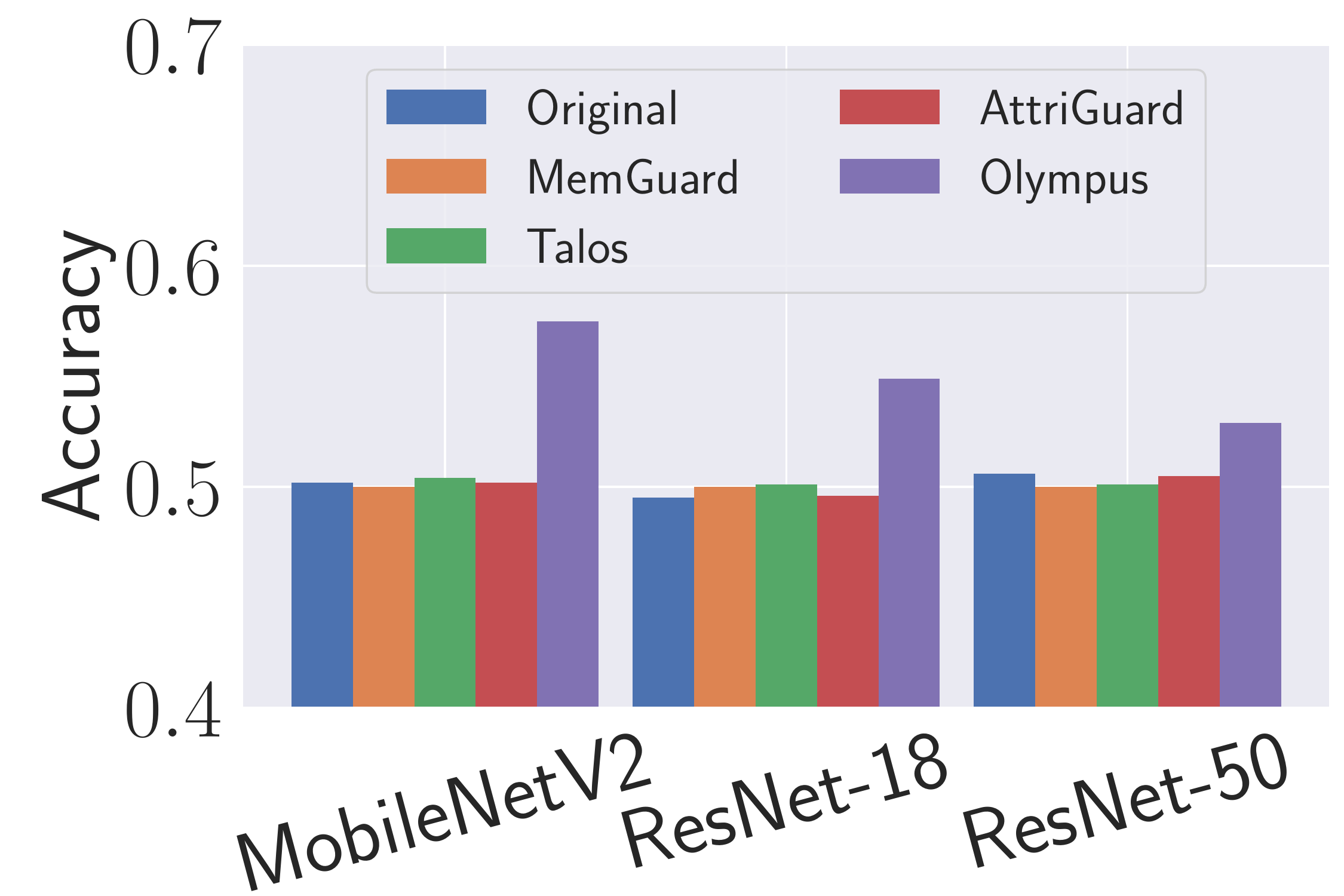}
\caption{UTKFace}
\label{figure:defense_comparison_mia_UTKFace_Metric-ent}
\end{subfigure}
\begin{subfigure}{0.5\columnwidth}
\includegraphics[width=\columnwidth]{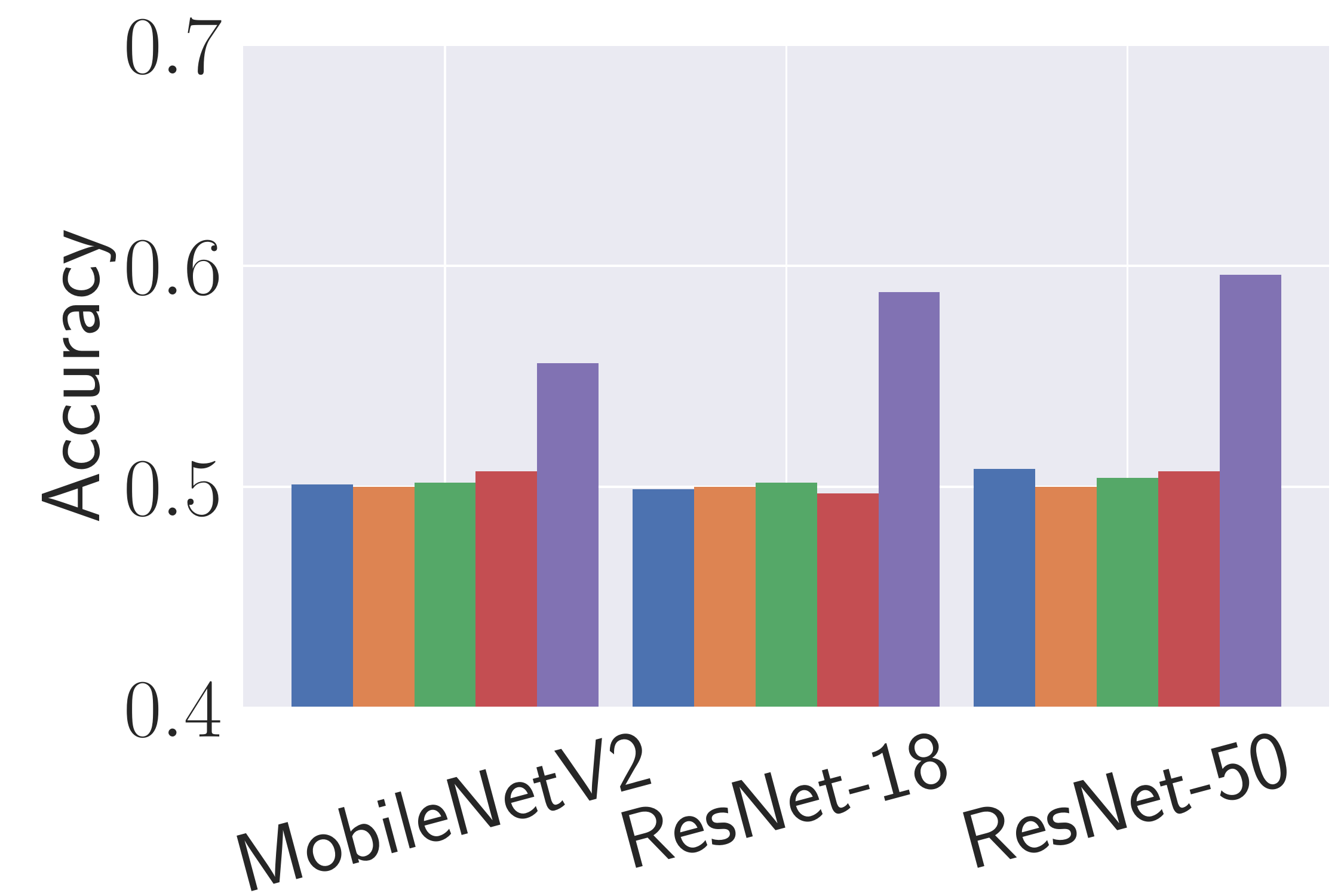}
\caption{Places100}
\label{figure:defense_comparison_mia_Places100_Metric-ent}
\end{subfigure}
\begin{subfigure}{0.5\columnwidth}
\includegraphics[width=\columnwidth]{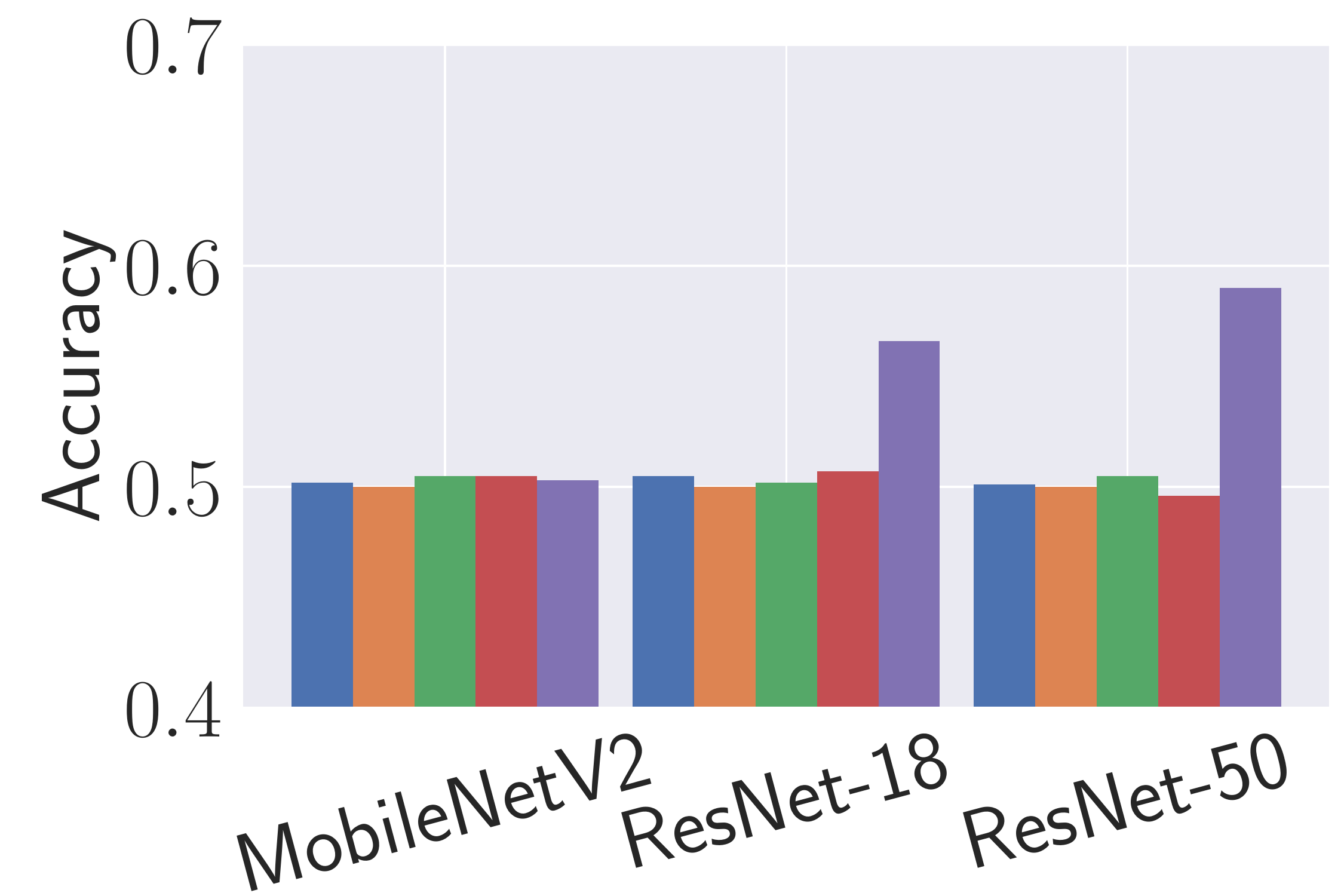}
\caption{Places50}
\label{figure:defense_comparison_mia_Places50_Metric-ent}
\end{subfigure}
\begin{subfigure}{0.5\columnwidth}
\includegraphics[width=\columnwidth]{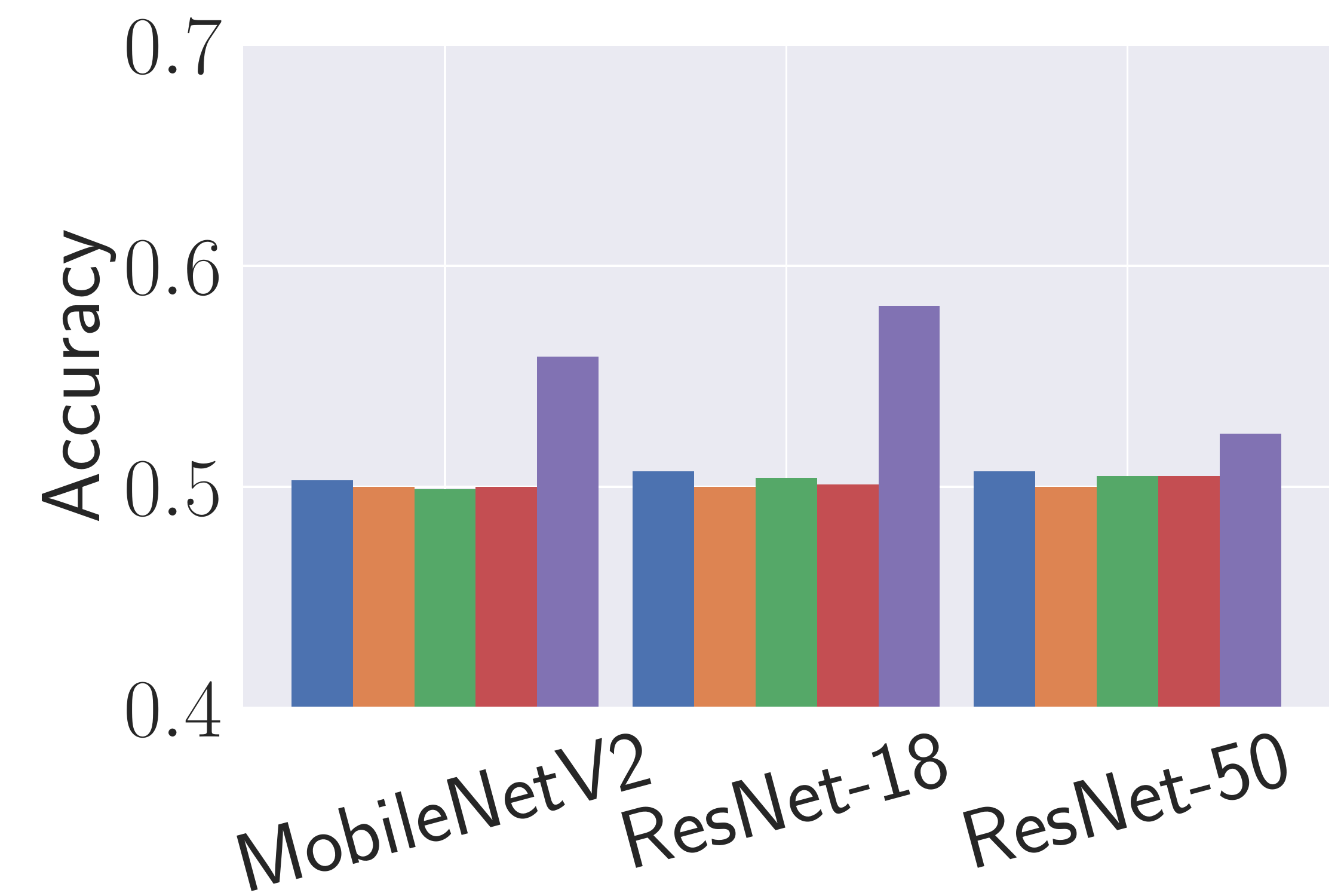}
\caption{Places20}
\label{figure:defense_comparison_mia_Places20_Metric-ent}
\end{subfigure}
\caption{
The performance of metric-ent membership inference attacks against original contrastive models, \Talos, \MemGuard, \Olympus, and \AttriGuard with MobileNetV2, ResNet-18, and ResNet-50 on 4 different datasets.
The x-axis represents different models.
The y-axis represents the accuracy of metric-ent membership inference attacks.}
\label{figure:defense_comparison_mia_Metric-ent}
\end{figure*}

\begin{figure*}[!ht]
\centering
\begin{subfigure}{0.5\columnwidth}
\includegraphics[width=\columnwidth]{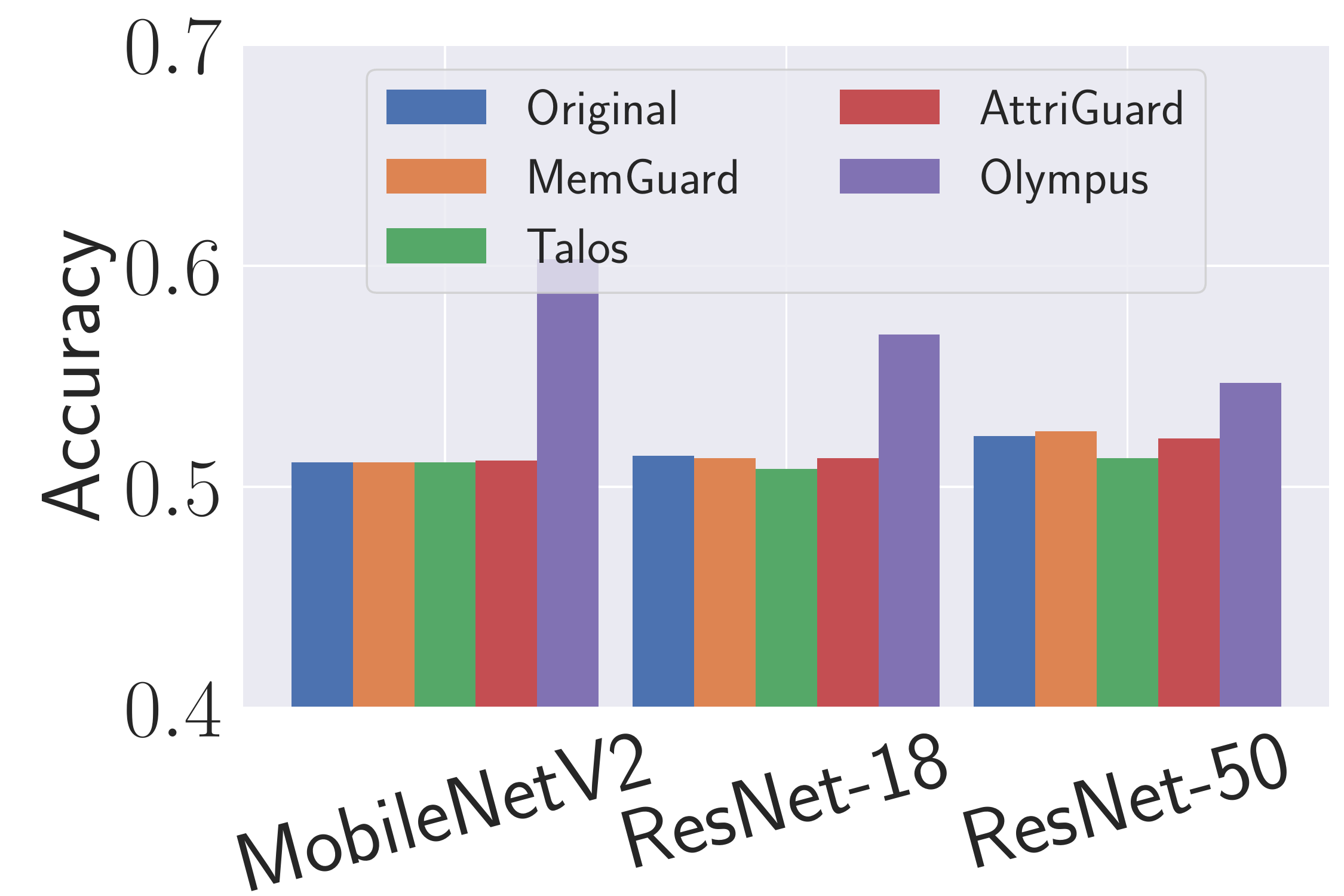}
\caption{UTKFace}
\label{figure:defense_comparison_mia_UTKFace_Metric-ment}
\end{subfigure}
\begin{subfigure}{0.5\columnwidth}
\includegraphics[width=\columnwidth]{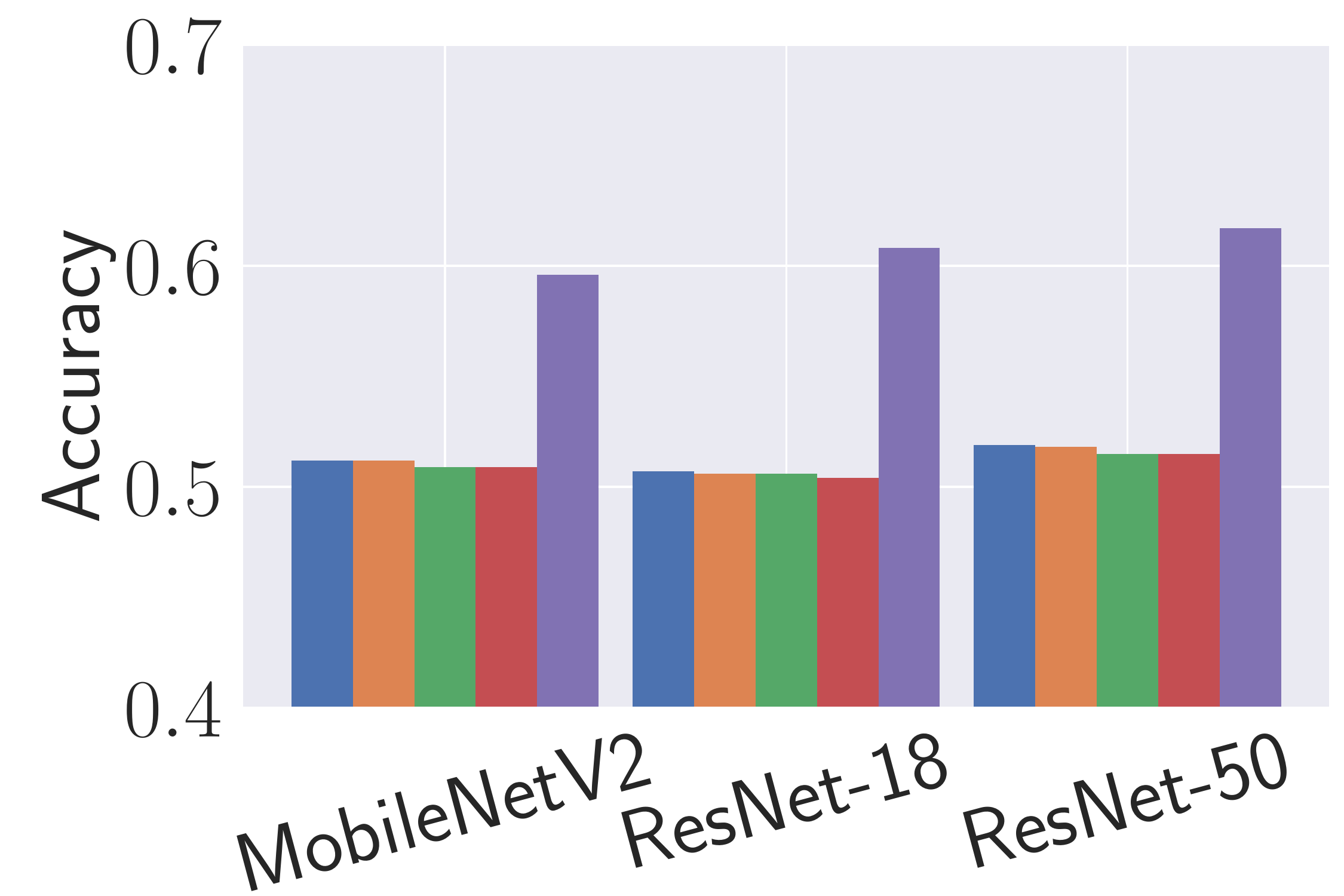}
\caption{Places100}
\label{figure:defense_comparison_mia_Places100_Metric-ment}
\end{subfigure}
\begin{subfigure}{0.5\columnwidth}
\includegraphics[width=\columnwidth]{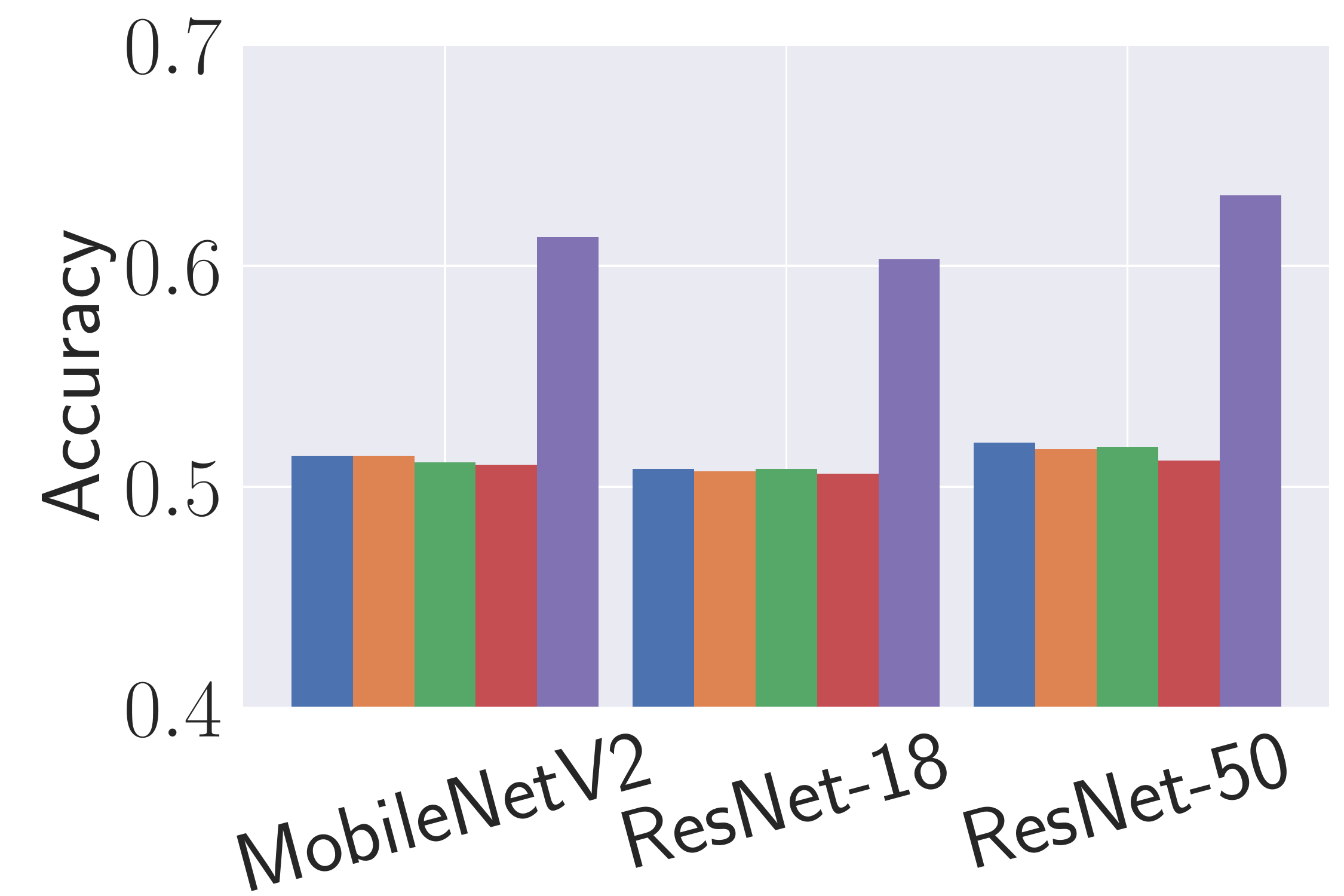}
\caption{Places50}
\label{figure:defense_comparison_mia_Places50_Metric-ment}
\end{subfigure}
\begin{subfigure}{0.5\columnwidth}
\includegraphics[width=\columnwidth]{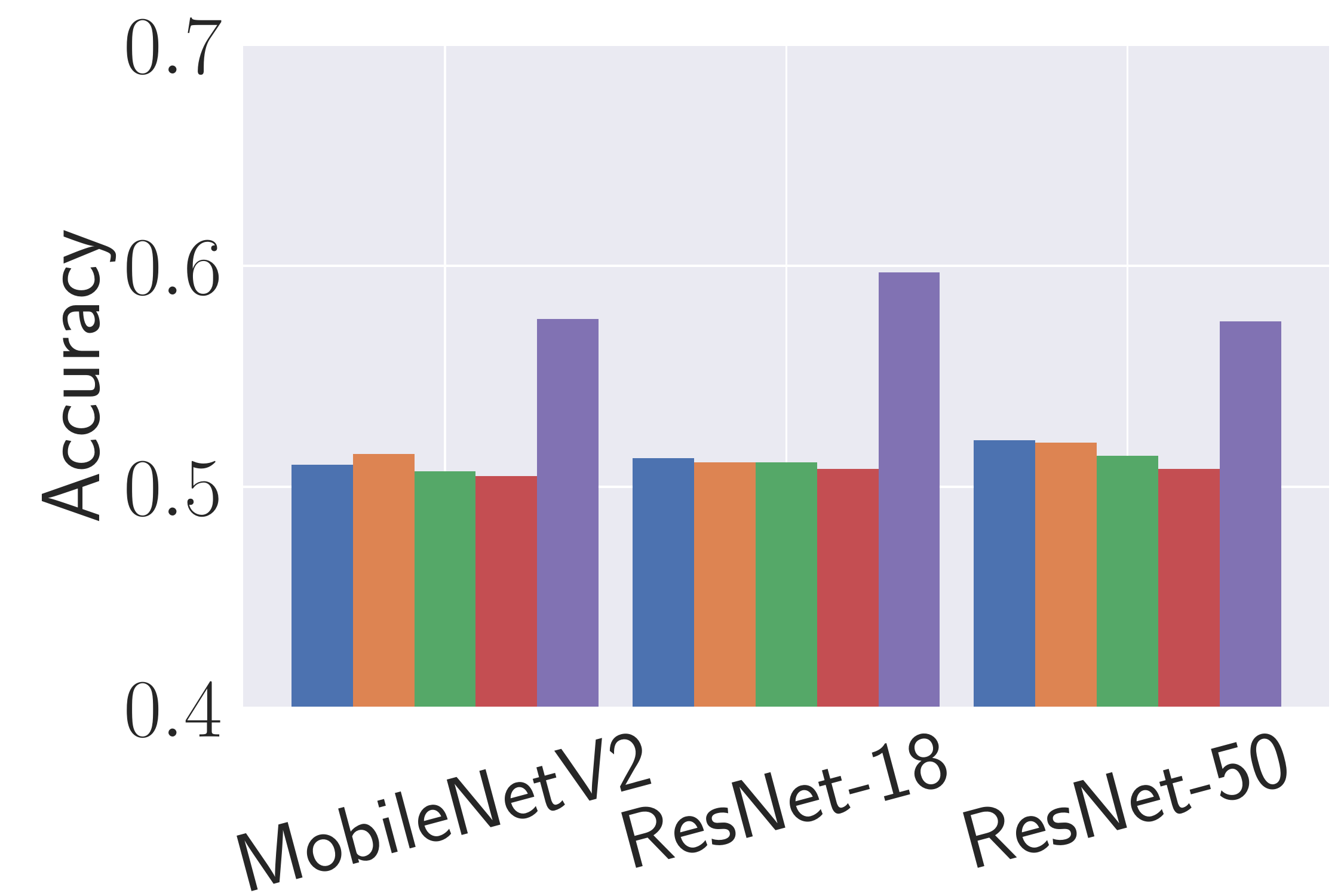}
\caption{Places20}
\label{figure:defense_comparison_mia_Places20_Metric-ment}
\end{subfigure}
\caption{
The performance of metric-ment membership inference attacks against original contrastive models, \Talos, \MemGuard, \Olympus, and \AttriGuard with MobileNetV2, ResNet-18, and ResNet-50 on 4 different datasets.
The x-axis represents different models.
The y-axis represents the accuracy of metric-ment membership inference attacks.}
\label{figure:defense_comparison_mia_Metric-ment}
\end{figure*}

\begin{figure*}[!ht]
\centering
\begin{subfigure}{0.5\columnwidth}
\includegraphics[width=\columnwidth]{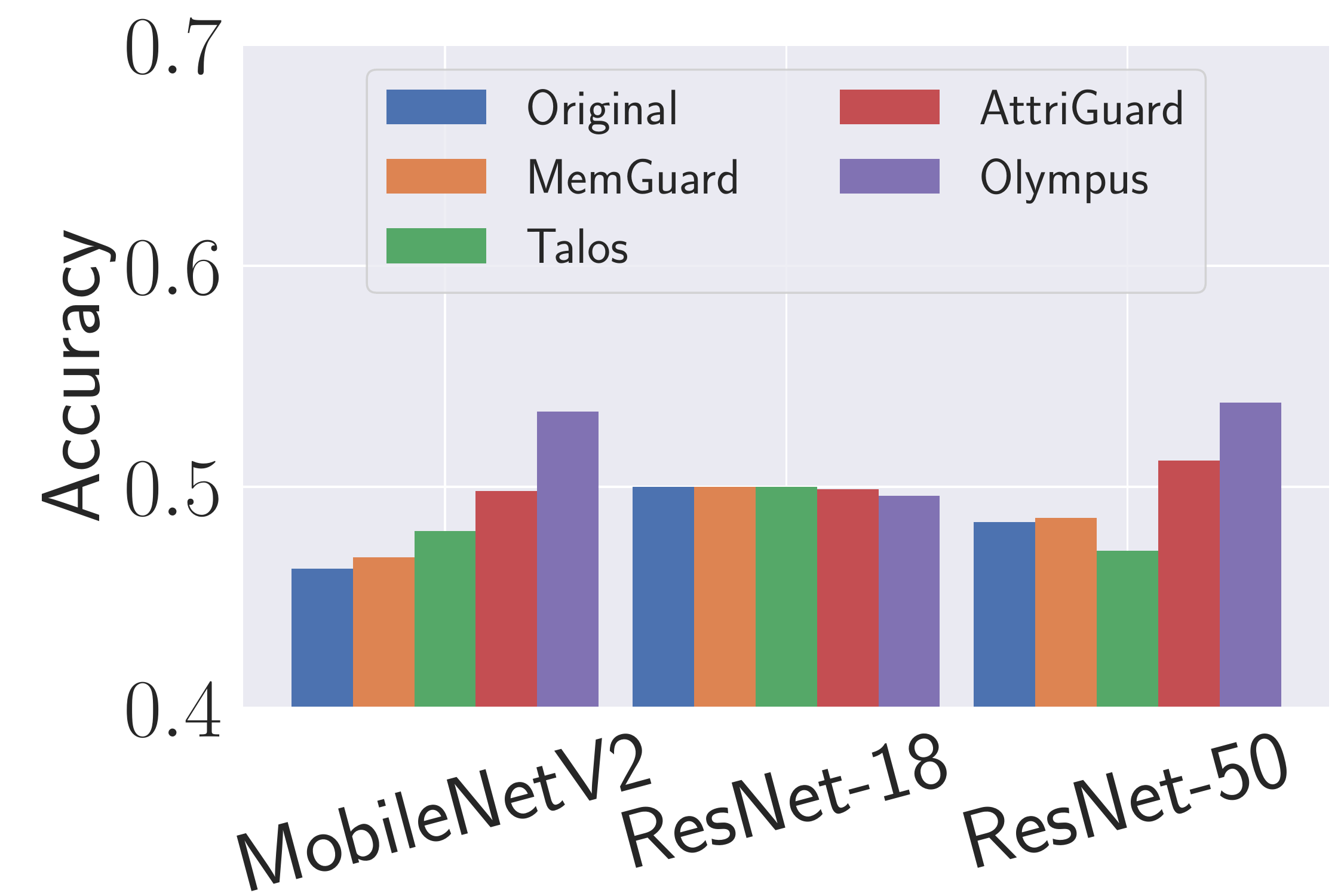}
\caption{UTKFace}
\label{figure:defense_comparison_mia_UTKFace_Label-only}
\end{subfigure}
\begin{subfigure}{0.5\columnwidth}
\includegraphics[width=\columnwidth]{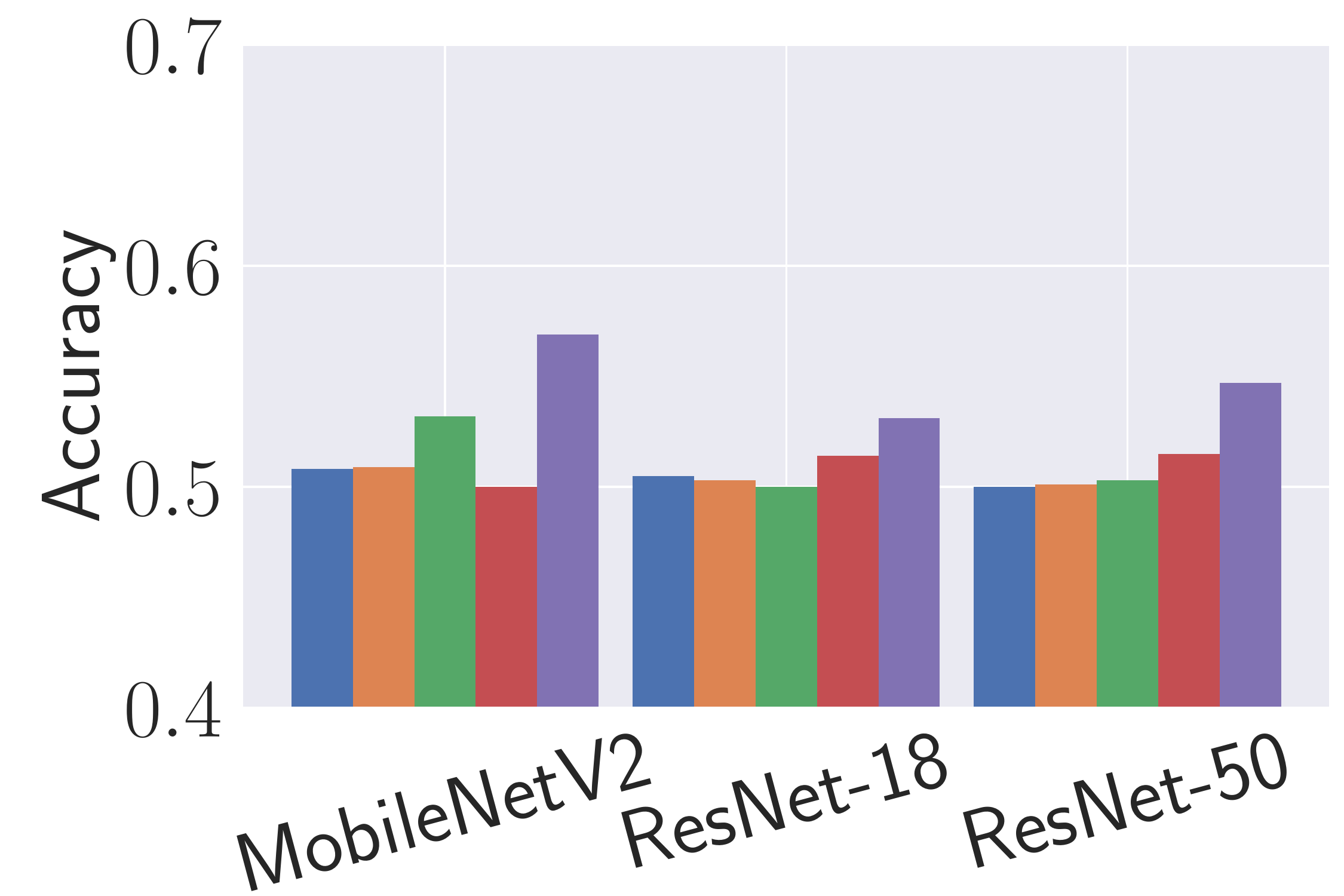}
\caption{Places100}
\label{figure:defense_comparison_mia_Places100_Label-only}
\end{subfigure}
\begin{subfigure}{0.5\columnwidth}
\includegraphics[width=\columnwidth]{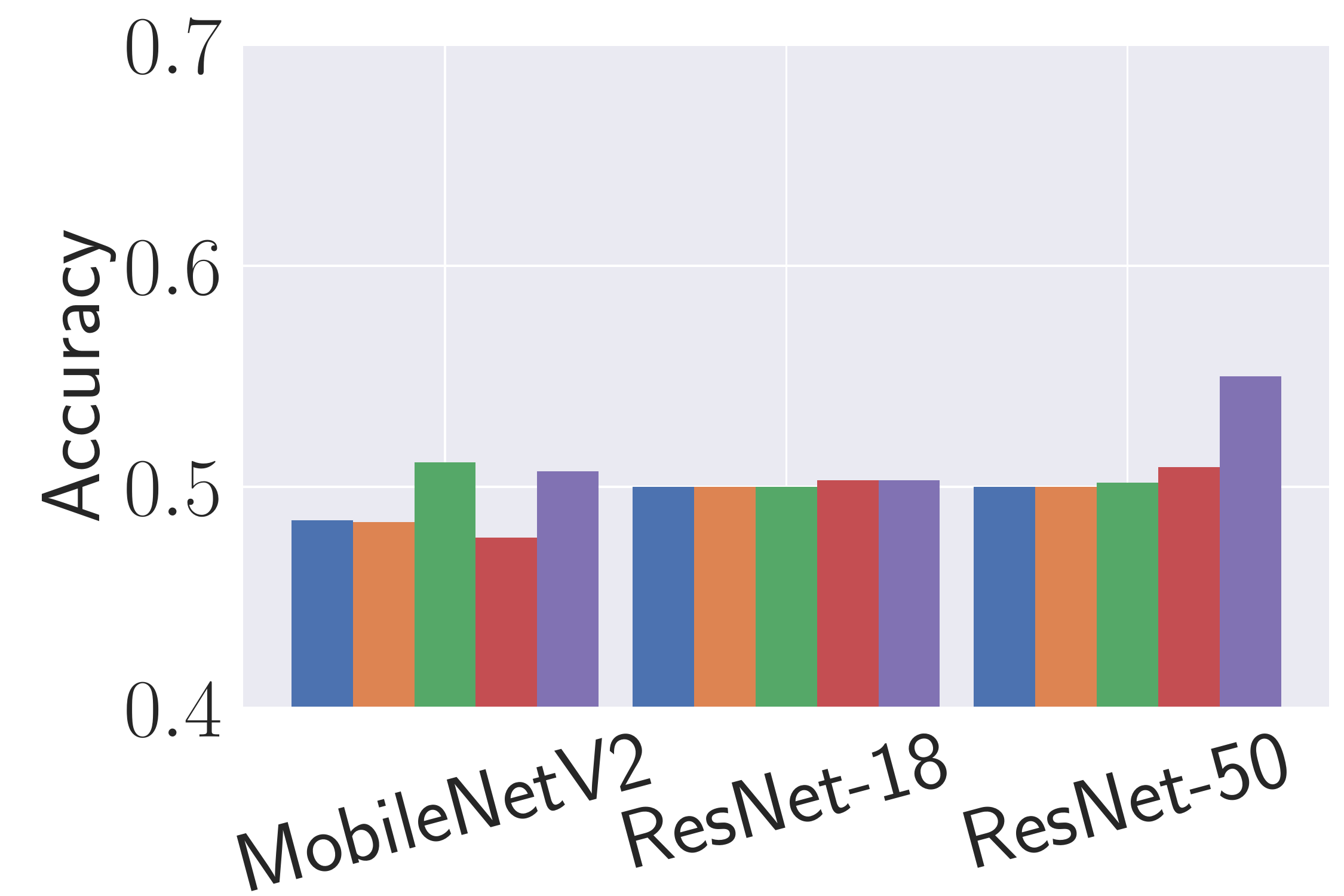}
\caption{Places50}
\label{figure:defense_comparison_mia_Places50_Label-only}
\end{subfigure}
\begin{subfigure}{0.5\columnwidth}
\includegraphics[width=\columnwidth]{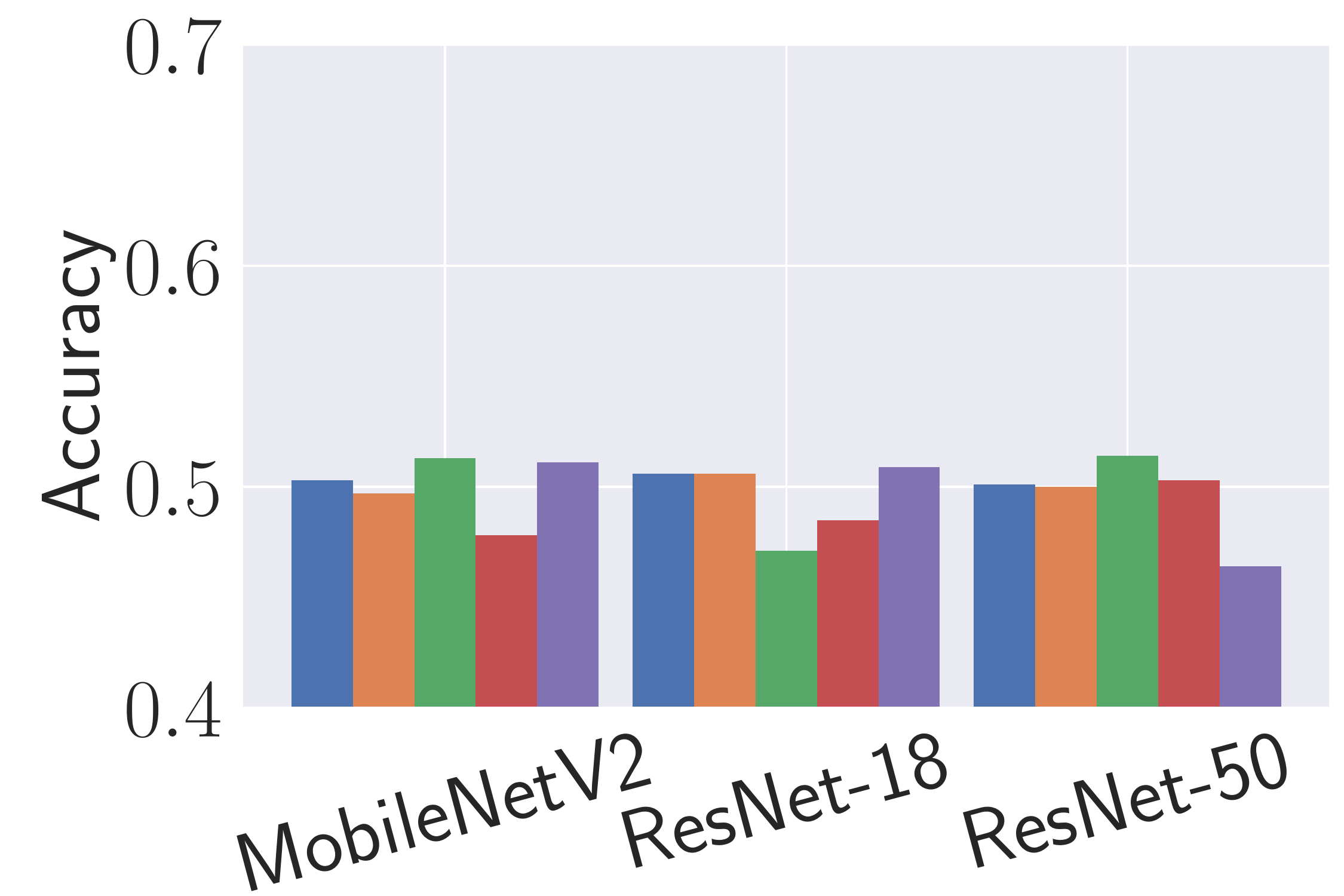}
\caption{Places20}
\label{figure:defense_comparison_mia_Places20_Label-only}
\end{subfigure}
\caption{
The performance of label-only membership inference attacks against original contrastive models, \Talos, \MemGuard, \Olympus, and \AttriGuard with MobileNetV2, ResNet-18, and ResNet-50 on 4 different datasets.
The x-axis represents different models.
The y-axis represents the accuracy of label-only membership inference attacks.}
\label{figure:defense_comparison_mia_Label-only}
\end{figure*}

\begin{figure*}[!ht]
\centering
\begin{subfigure}{0.5\columnwidth}
\includegraphics[width=\columnwidth]{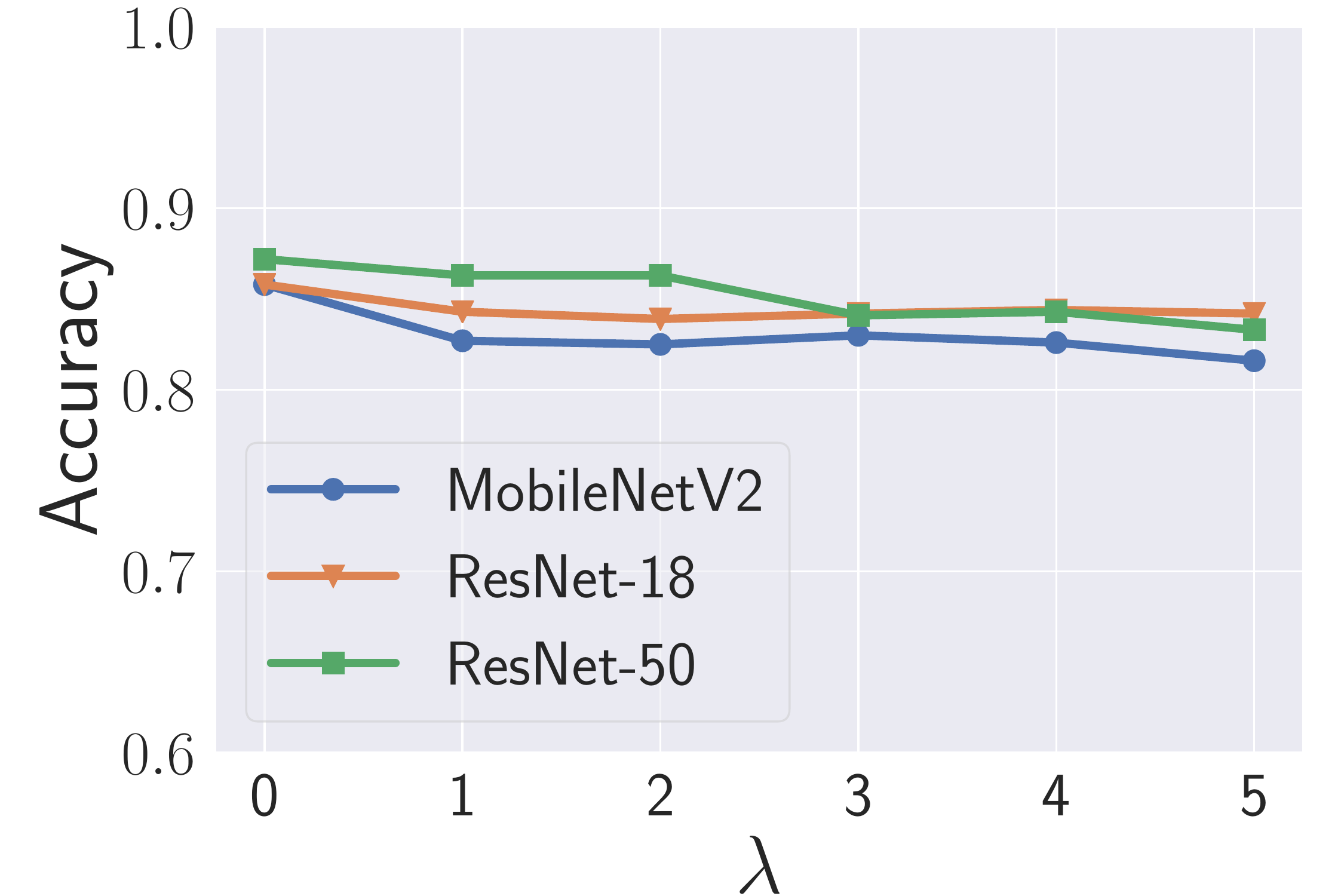}
\caption{UTKFace}
\label{figure:adv_simclr_target_utkface}
\end{subfigure}
\begin{subfigure}{0.5\columnwidth}
\includegraphics[width=\columnwidth]{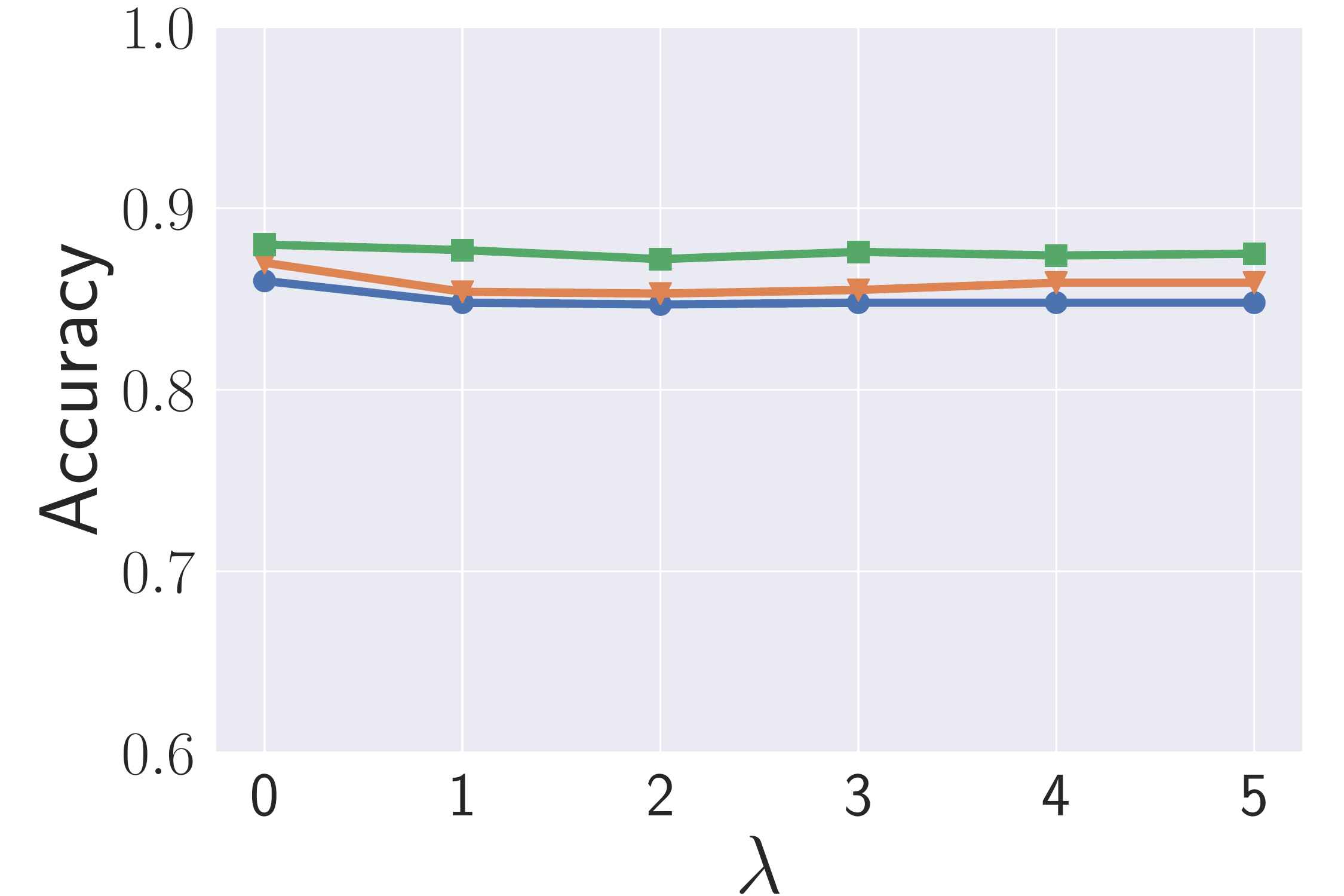}
\caption{Places100}
\label{figure:adv_simclr_target_place100}
\end{subfigure}
\begin{subfigure}{0.5\columnwidth}
\includegraphics[width=\columnwidth]{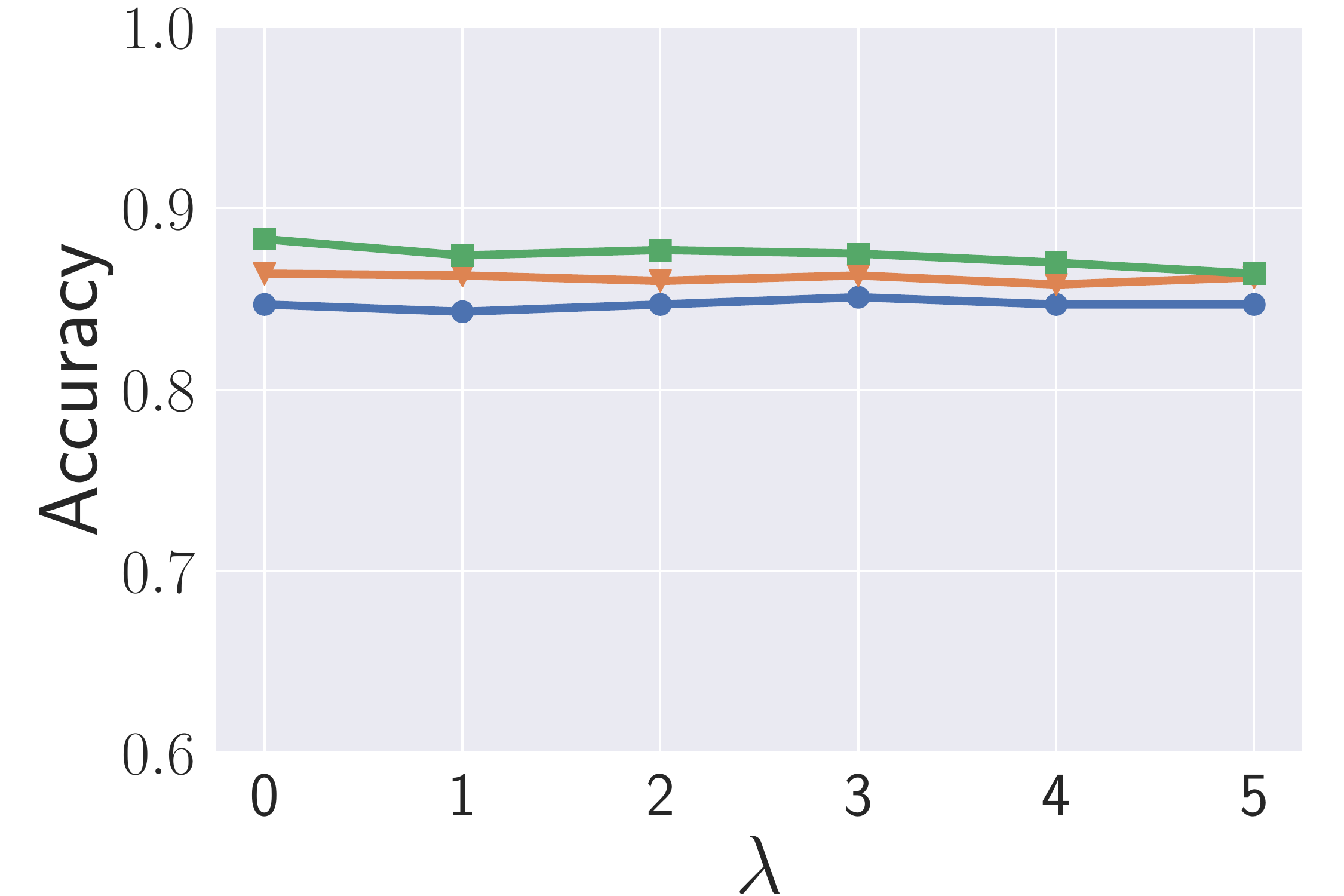}
\caption{Places50}
\label{figure:adv_simclr_target_place50}
\end{subfigure}
\begin{subfigure}{0.5\columnwidth}
\includegraphics[width=\columnwidth]{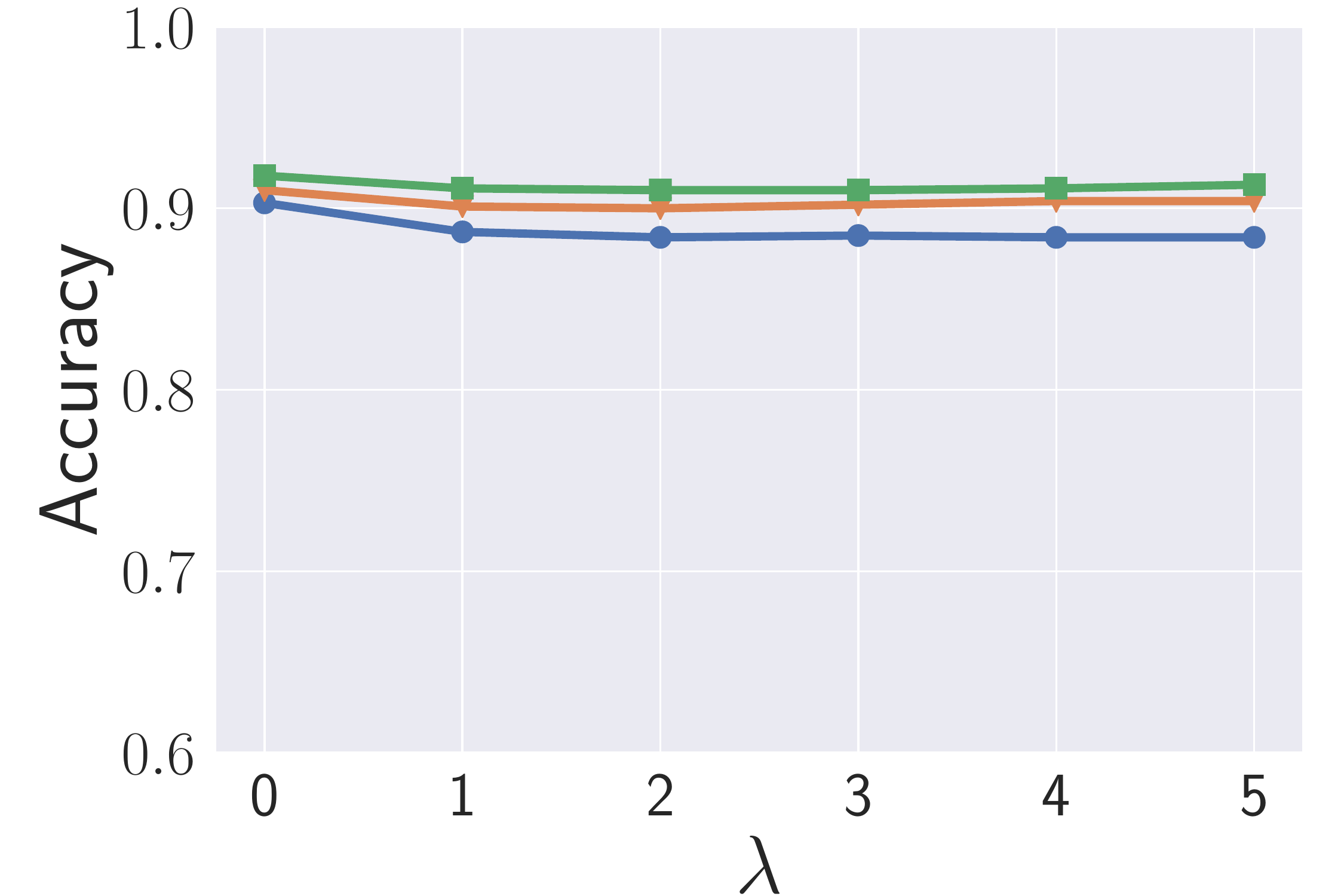}
\caption{Places20}
\label{figure:adv_simclr_target_place20}
\end{subfigure}
\caption{The performance of original classification tasks for the \Talos models with MobileNetV2, ResNet-18, and ResNet-50 on 4 different datasets under different adversarial factor $\lambda$.
The x-axis represents different $\lambda$.
The y-axis represents the corresponding performance.
}
\label{figure:adv_training_target_performance}
\end{figure*}

\begin{figure*}[!ht]
\centering
\begin{subfigure}{0.5\columnwidth}
\includegraphics[width=\columnwidth]{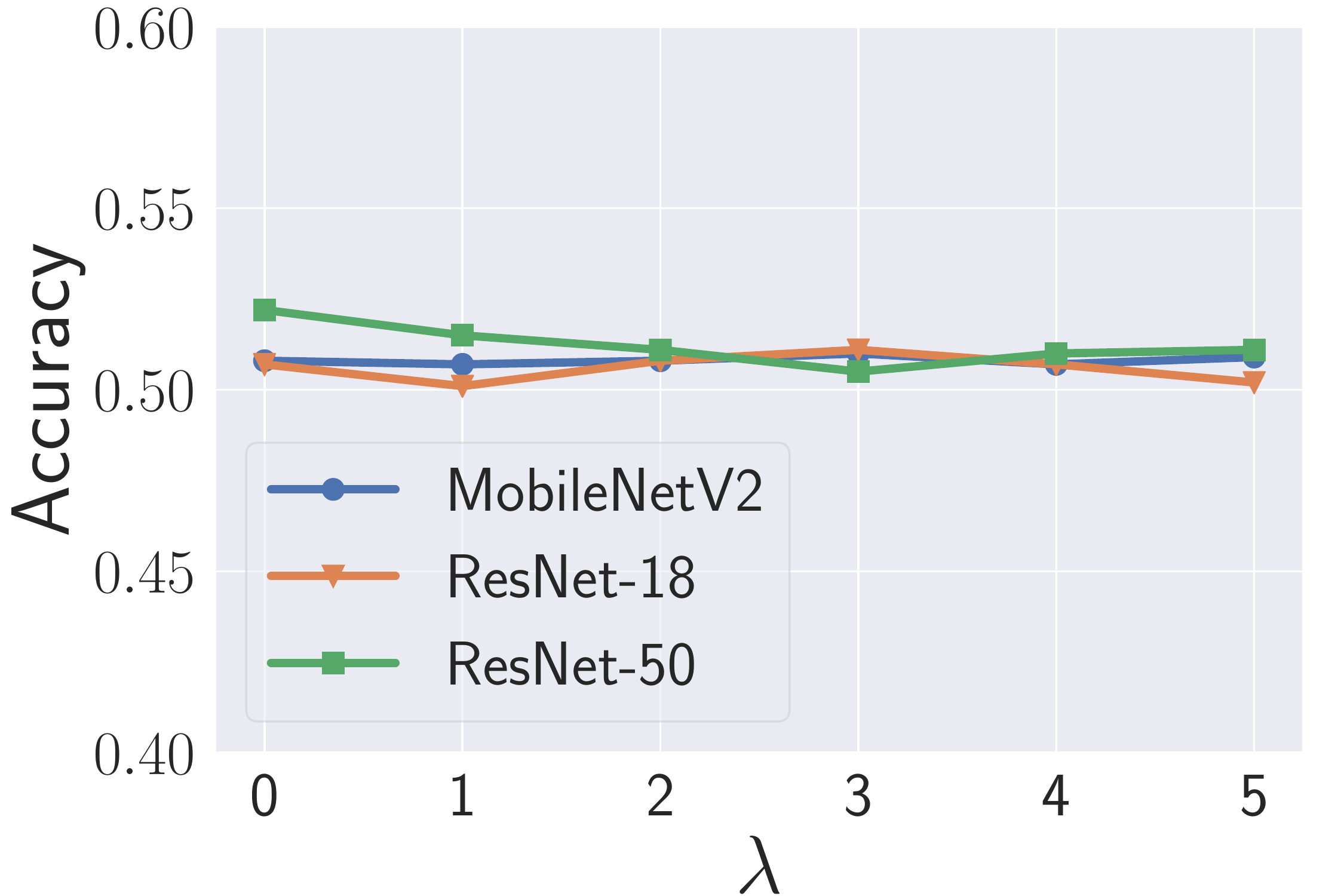}
\caption{UTKFace}
\label{figure:adv_simclr_mia_utkface}
\end{subfigure}
\begin{subfigure}{0.5\columnwidth}
\includegraphics[width=\columnwidth]{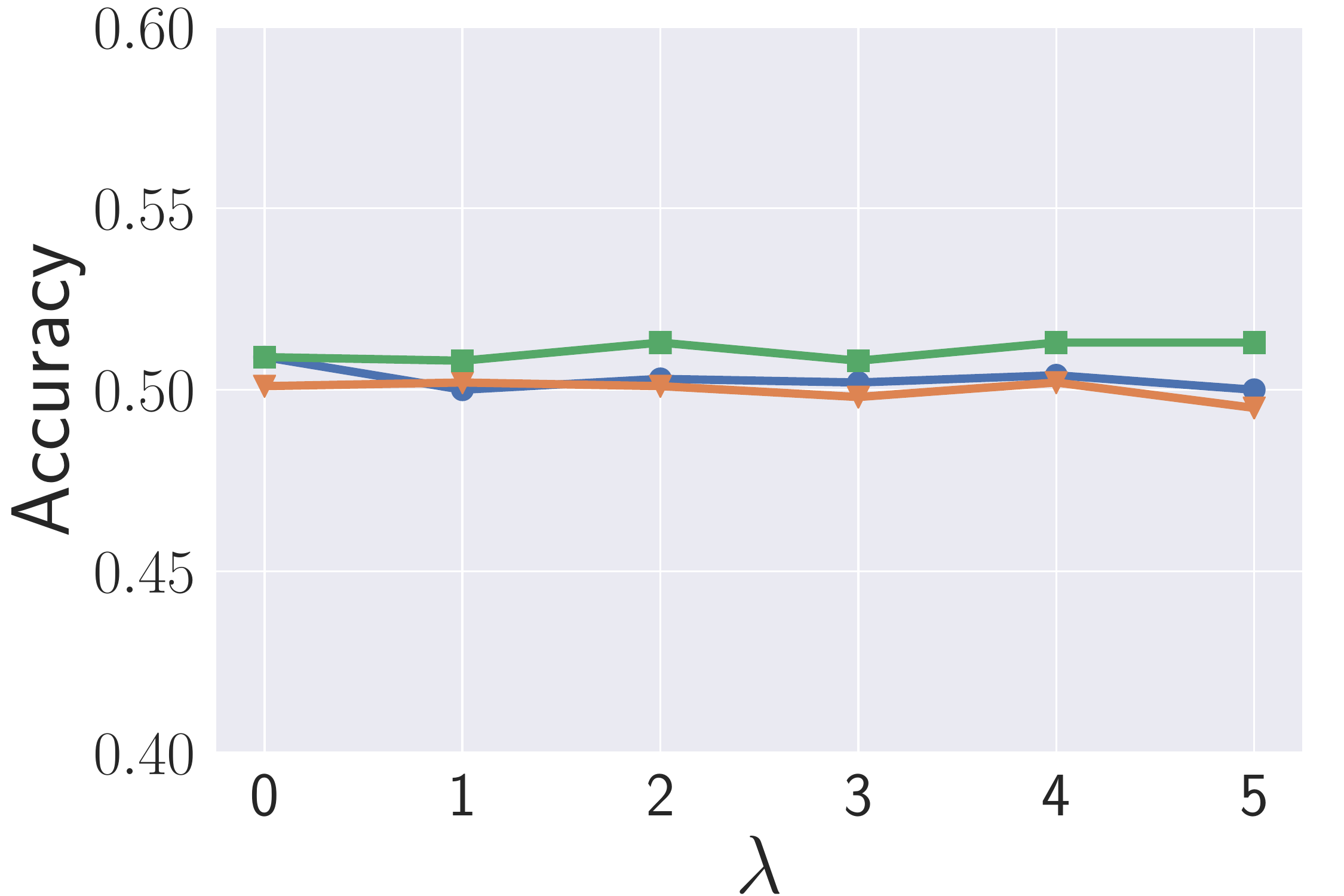}
\caption{Places100}
\label{figure:adv_simclr_mia_place100}
\end{subfigure}
\begin{subfigure}{0.5\columnwidth}
\includegraphics[width=\columnwidth]{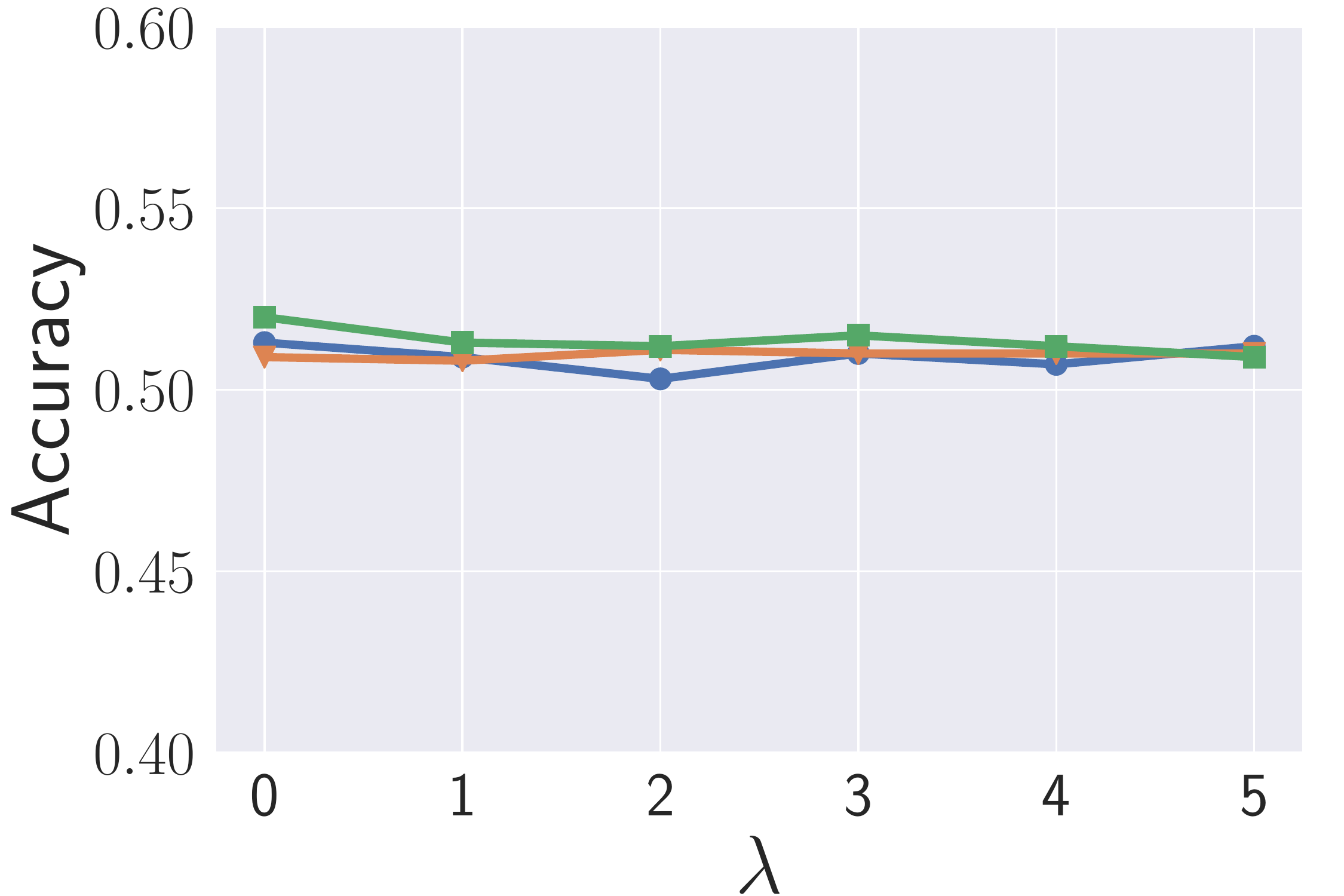}
\caption{Places50}
\label{figure:adv_simclr_mia_place50}
\end{subfigure}
\begin{subfigure}{0.5\columnwidth}
\includegraphics[width=\columnwidth]{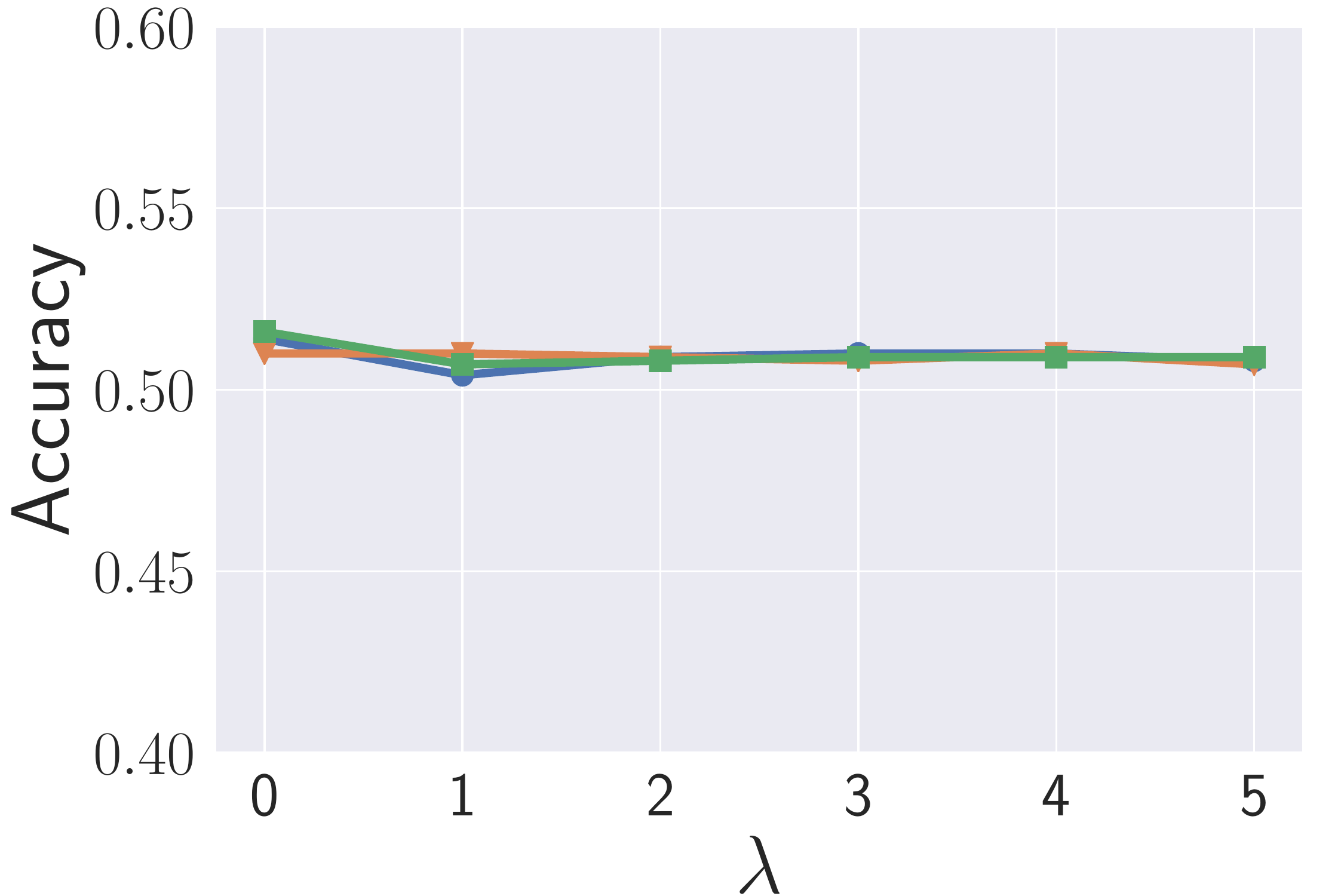}
\caption{Places20}
\label{figure:adv_simclr_mia_place20}
\end{subfigure}
\caption{The performance of membership inference attacks for the \Talos models with MobileNetV2, ResNet-18, and ResNet-50 on 4 different datasets under different adversarial factor $\lambda$.
The x-axis represents different $\lambda$.
The y-axis represents the corresponding performance.
}
\label{figure:adv_training_mia_performance}
\end{figure*}

\begin{figure*}[!ht]
\centering
\begin{subfigure}{0.5\columnwidth}
\includegraphics[width=\columnwidth]{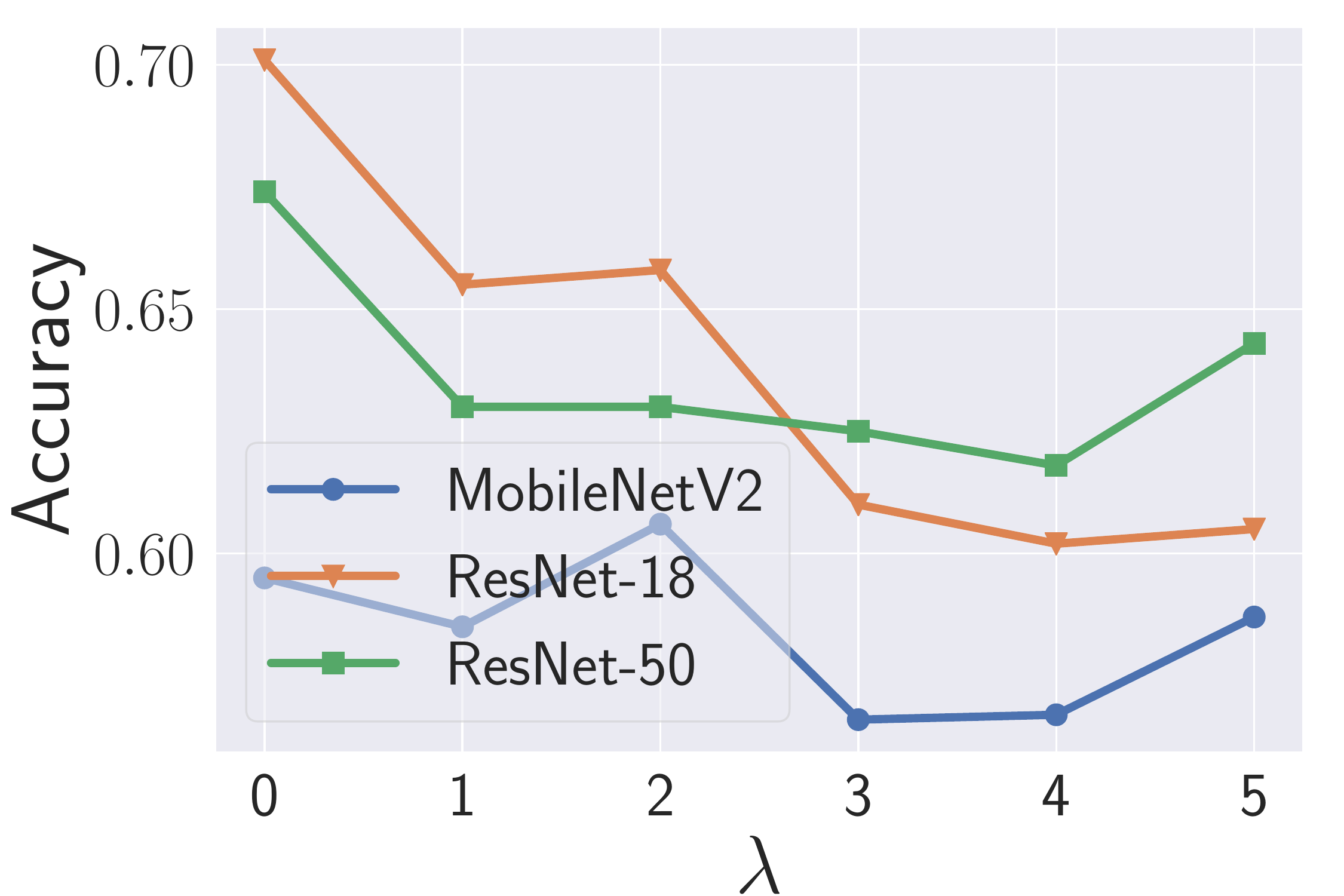}
\caption{UTKFace}
\label{figure:adv_simclr_ai_utkface}
\end{subfigure}
\begin{subfigure}{0.5\columnwidth}
\includegraphics[width=\columnwidth]{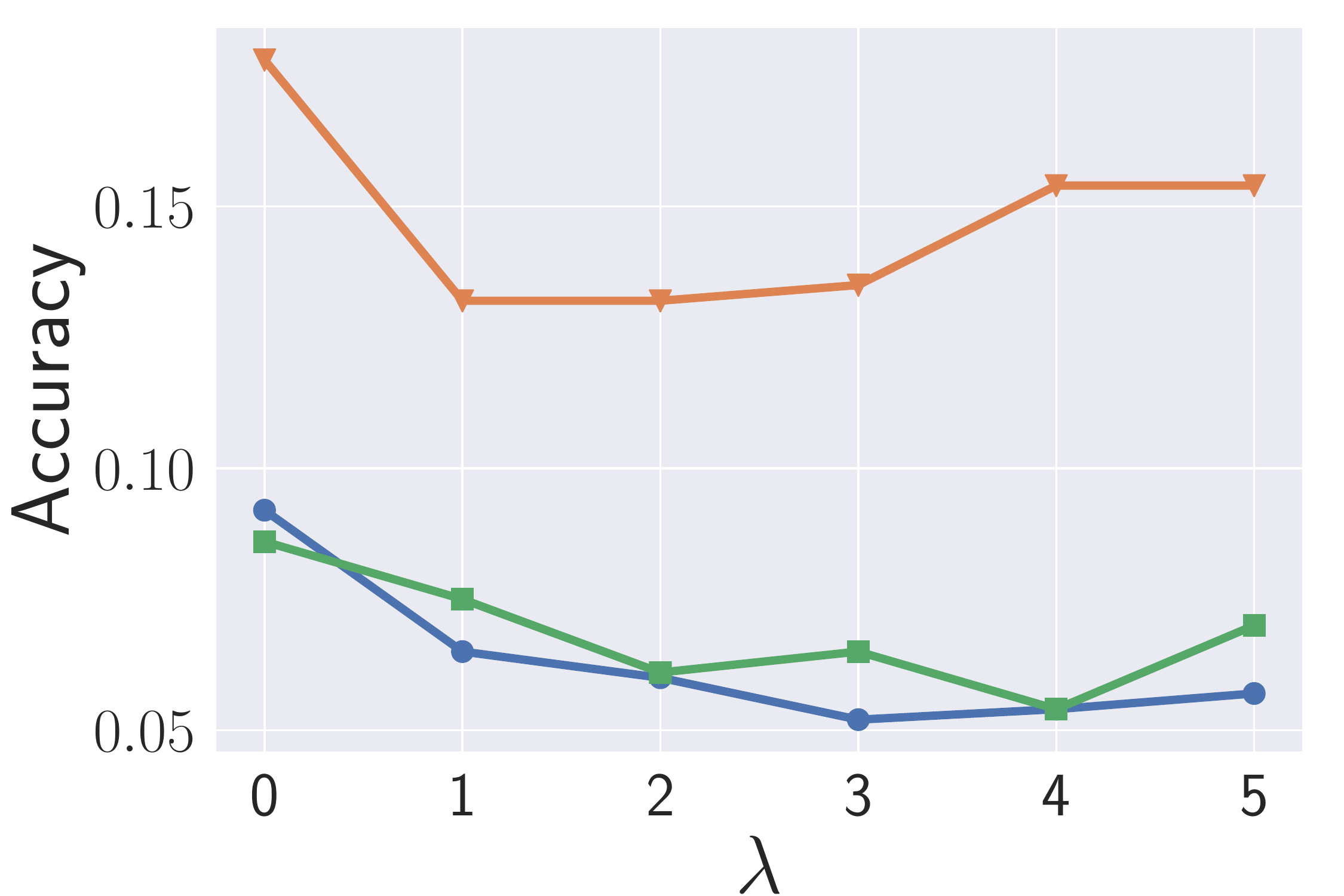}
\caption{Places100}
\label{figure:adv_simclr_ai_place100}
\end{subfigure}
\begin{subfigure}{0.5\columnwidth}
\includegraphics[width=\columnwidth]{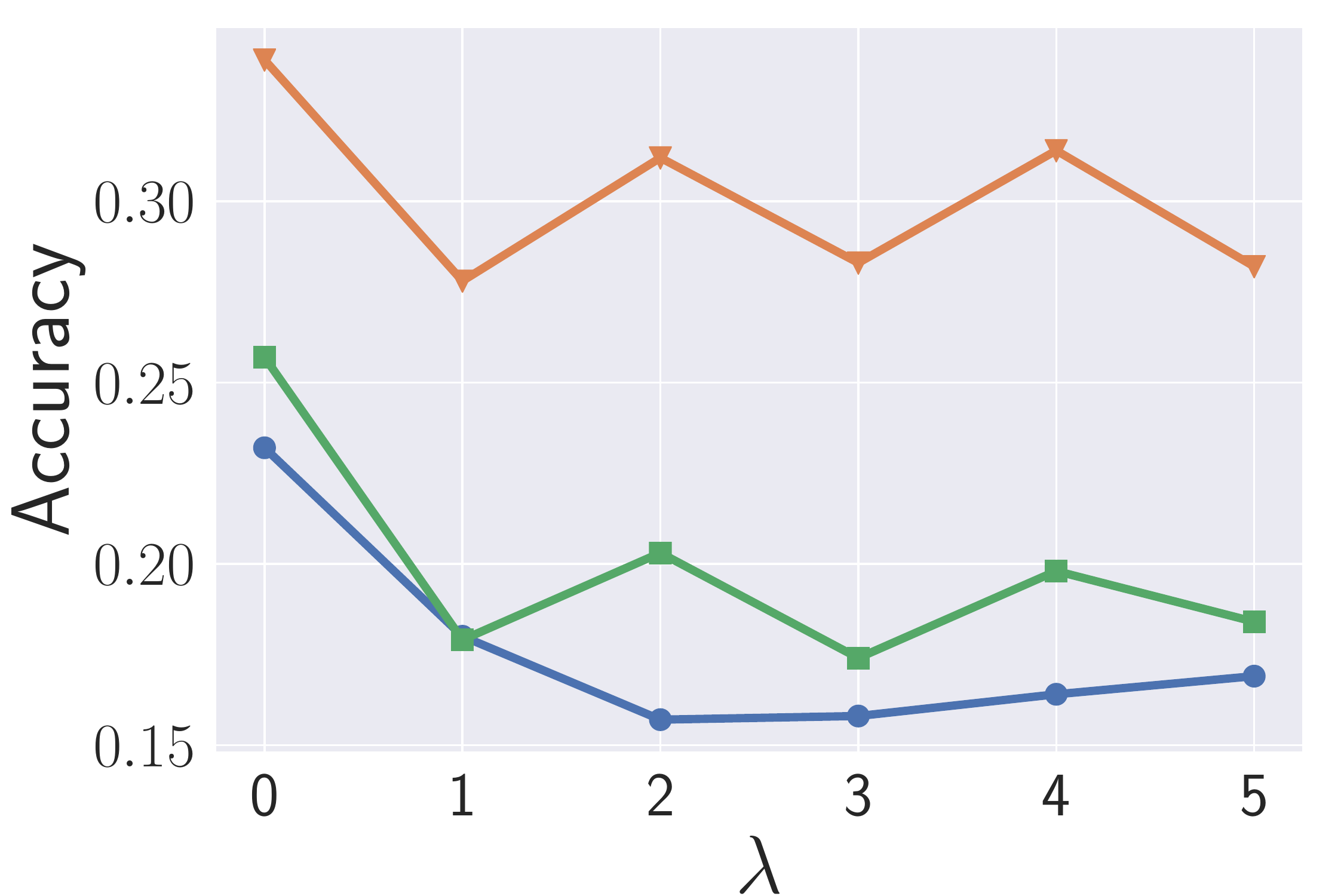}
\caption{Places50}
\label{figure:adv_simclr_ai_place50}
\end{subfigure}
\begin{subfigure}{0.5\columnwidth}
\includegraphics[width=\columnwidth]{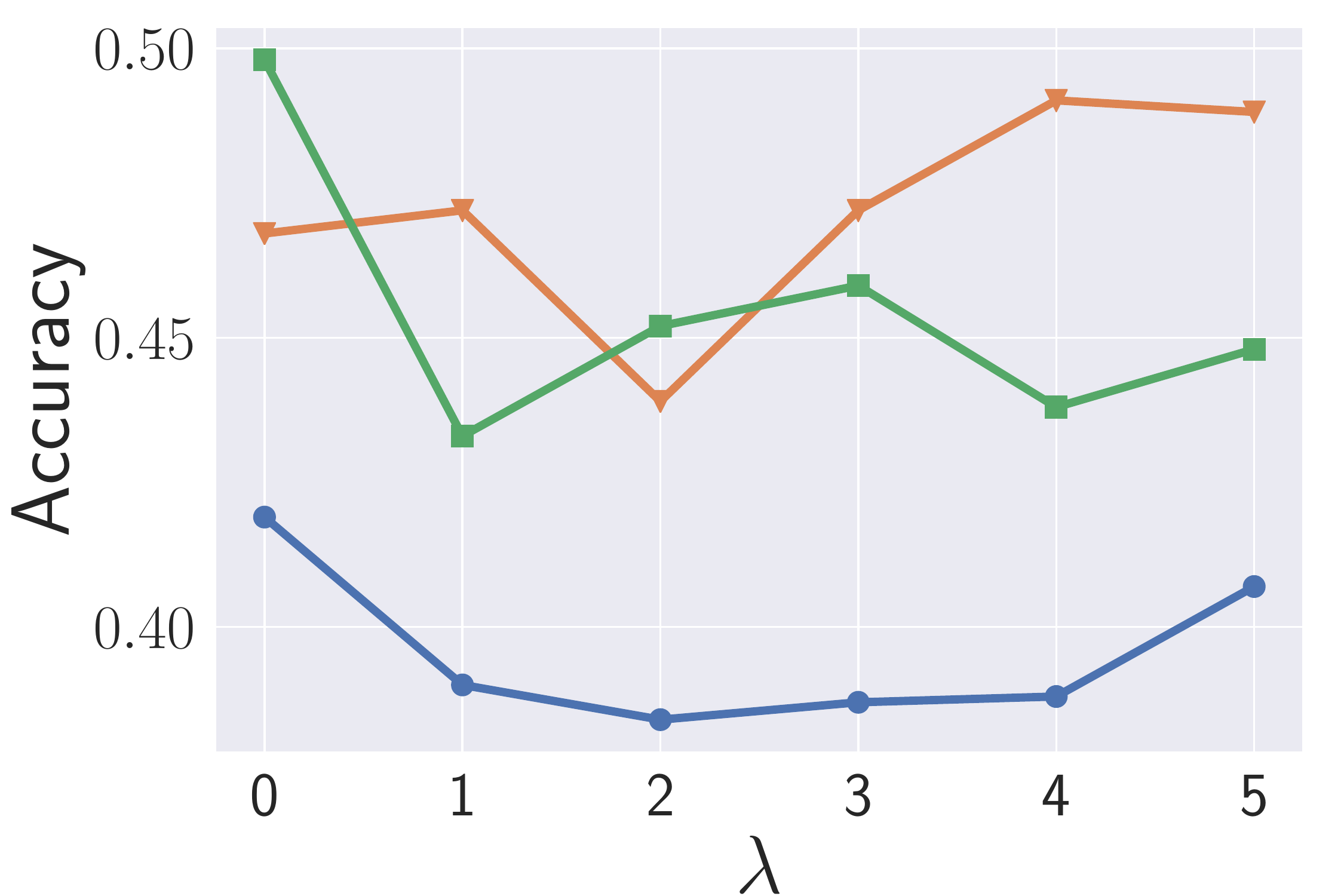}
\caption{Places20}
\label{figure:adv_simclr_ai_place20}
\end{subfigure}
\caption{The performance of attribute inference attacks for the \Talos models with MobileNetV2, ResNet-18, and ResNet-50 on 4 different datasets under different adversarial factor $\lambda$.
The x-axis represents different $\lambda$.
The y-axis represents the corresponding performance.
}
\label{figure:adv_training_ai_performance}
\end{figure*}

% ----------------------------------------------------
\end{document}